\documentclass[preprint,authoryear,10pt,a4paper]{elsarticle}
\usepackage[top=0.75in, bottom=0.75in, left=0.75in, right=0.75in]{geometry}
\usepackage{palatino}
\usepackage{amsmath,amsthm}
\usepackage{amssymb}
\usepackage{bbm}
\usepackage{amsfonts}
\usepackage{graphicx}
\usepackage{subfigure}
\usepackage{grffile}
\graphicspath{{graphics/}}
\usepackage[ruled,vlined,linesnumbered]{algorithm2e}
\usepackage{tabularx,booktabs}
\usepackage{inputenc} 
\usepackage[T1]{fontenc}   
\usepackage{url}            % simple URL 
\usepackage{booktabs}       % 
\usepackage{amsmath,amsthm}
\usepackage{amssymb}
\usepackage{bbm}
\usepackage{amsfonts}       % blackboard math 
\usepackage{nicefrac}       % compact symbols 
\usepackage{microtype}      % microtypography
\usepackage{tikz}
\usepackage{rotfloat}
\usepackage{subfigure}
\usepackage{graphicx}
\usepackage{caption}
\usepackage{appendix}
\usepackage{tablefootnote}
\usepackage{bm}
\usepackage{mathrsfs}
\usepackage{xcolor}
\usepackage{colortbl}

\usepackage{xcolor}
\definecolor{darkblue}{rgb}{0.0,0.5,0.5}
\definecolor{blue}{rgb}{0.0,0.59,0.84}
\definecolor{myblue}{RGB}{0,0,255}

\usepackage[colorlinks]{hyperref}
\hypersetup{colorlinks,breaklinks,linkcolor=blue,urlcolor=blue,anchorcolor=blue,citecolor=blue}
\usepackage{booktabs,caption}
\usepackage{multirow,bigdelim}
\usepackage{lscape}
\usepackage{lineno}
\usepackage{microtype}
\newtheorem{definition}{Definition}

\usepackage{soul}
\usepackage{lineno}
\usepackage{fancyhdr}
\newcommand{\minew}[1]{{\color{black}{#1}}}

\journal{Transportation Research Part C: Emerging Technologies}

\begin{document}

\begin{frontmatter}



% \title{{\fontfamily{lmss}\selectfont Spatiotemporal graph embedded low-rank tensor learning for network-wide traffic speed kriging with incomplete observations}}
% \title{{\fontfamily{lmss}\selectfont  Spatiotemporal traffic speed kriging for enhancing network-wide sensor perception with low coverage: an integrated graph tensor learning method}}

% \title{{\fontfamily{lmss}\selectfont  Augmenting spatial coverage of traffic volume sensors for highway network via spatiotemporal correlation adaptive graph neural networks}}
% \title{{\fontfamily{lmss}\selectfont Towards better traffic volume estimation: Tackling both underdetermination and nonequilibrium problems via correlation-adaptive graph convolutional networks}}
\title{{\fontfamily{lmss}\selectfont Towards better traffic volume estimation: Jointly addressing the underdetermination and nonequilibrium problems with correlation-adaptive GNNs}}

\author[label]{Tong Nie}
\author[label]{Guoyang Qin}
\author[label1]{Yunpeng Wang}
\author[label]{Jian Sun\corref{cor1}}
\ead{sunjian@tongji.edu.cn}

\address[label]{Department of Traffic Engineering \& Key Laboratory of Road and Traffic Engineering, Ministry of Education, Tongji University, Shanghai 201804, China}
\address[label1]{
Beijing Key Laboratory for Cooperative Vehicle Infrastructure Systems and Safety Control, School of Transportation Science and Engineering, Beihang University, Beijing 100191, China}

\cortext[cor1]{Corresponding author. Address: Cao’an Road 
4800, Shanghai 201804, China}

\begin{abstract}
Traffic volume is an indispensable ingredient to provide fine-grained information for traffic management and control. However, due to the limited deployment of traffic sensors, obtaining full-scale volume information is far from easy. Existing work on this topic focuses primarily on improving the overall estimation accuracy of a particular method and ignores the underlying challenges of volume estimation, thereby having inferior performance in some critical tasks. This paper studies two key problems with regard to traffic volume estimation: (1) underdetermined traffic flows caused by fully undetected paths that can allow arbitrary volume values without violating the conservation law, where using local side information is insufficient to tackle, and (2) nonequilibrium traffic flows arise when traffic flows vary in density over space and time due to congestion propagation delay, which produce time-shifted volume readings to varying degrees. Here we demonstrate a graph-based deep learning method that offers a data-driven, equation-free, and correlation-adaptive approach to address the above issues and perform accurate network-wide traffic volume estimation. Particularly, in order to quantify the dynamic and nonlinear speed-volume relationships for the estimation of underdetermined flows, a speed pattern-adaptive adjacency matrix based on graph attention is developed and integrated into the graph convolution process, to capture nonlocal correlations between volumes.
To measure the impacts of nonequilibrium flows, a temporal masked and clipped attention combined with a gated temporal convolution layer is customized to capture time-asynchronous correlations between upstream and downstream sensors under varying impacts of congestion delay. We then evaluate our model on a real-world highway traffic volume dataset and compare it with several benchmark models. It is demonstrated that the proposed model achieves high estimation accuracy even under $20\%$ sensor coverage rate and outperforms other baselines significantly, especially on underdetermined and
nonequilibrium flow locations. Furthermore, comprehensive quantitative model analysis is also carried out to justify the model designs. The source code and datasets are publicly available at: \url{https://github.com/tongnie/GNN4Flow}.

\end{abstract}


\end{frontmatter}

% \linenumbers
\section{Introduction}\label{Introduction}
Traffic volumes have been widely measured due to the rapid development of traffic sensing technology. As one of the three macroscopic traffic characteristics, volume directly depicts the operation of traffic flows. Real-time traffic volume information at network-wide is one of the main ingredients required to provide fundamental information for both traffic agencies and individuals to make timely decisions. All the while, ever-increasing problems in transportation systems has prompted traffic managers in dire need of obtaining full-scale of volume data for better control and management of traffic. For instance, estimating time-dependent origin-destination flows \citep{ashok2002estimation}, formulating proactive traffic control policies \citep{sirmatel2017economic,yildirimoglu2018hierarchical}, measuring urban traffic emissions \citep{liu2019spatial}, and calibrating microscopic traffic simulator \citep{saeedmanesh2021extended} entail detailed traffic volume data, and a majority of modern traffic forecasting models are built on complete volume observations.

\begin{figure}[!htb]
\centering
\includegraphics[scale=0.5]{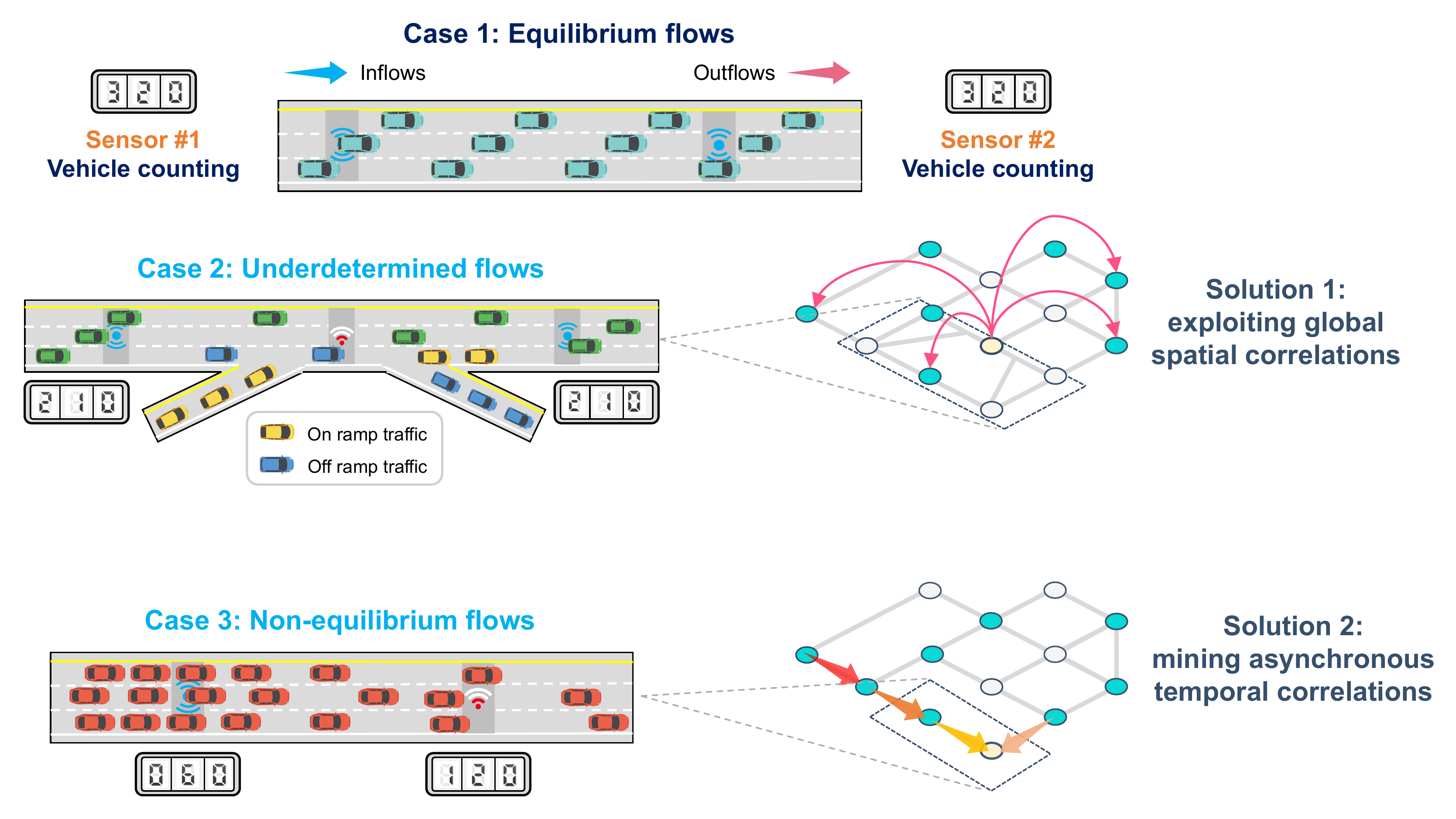}
\caption{Illustration of highway traffic volume estimation problem. 
Three typical road segments of the highway network are shown to illustrate the three scenarios studied in this paper. (a) Case of equilibrium and determined traffic flows: the density between sensor \#1 and \#2 is constant so volumes of sensor \#2 are identical to sensor \#1.
(b) Case of underdetermined traffic flows: due to lack of measurements at on/off ramps, arbitrary volume values may exist between the two sensors, even if the two detectors have the same or close reading. In this case, volumes in the middle position cannot be inferred directly from conservation equation.
(c) Case of nonequilibrium traffic flows: influenced by congestion delay and different reactions of vehicles, the outflows are distinct from inflows, resulting in time-shifted readings of sensor \#1 and \#2.
 In general, cases (b) and (c) are more intractable because when sensor \#1 or \#2 is unavailable, their readings deviate from neighborhood values, which the focus of this paper is on. To tackle the two key problems, global spatial correlations and asynchronous temporal correlations are exploited by customizing a GNN model.
}% In this example, sensor \#1,2 and 3 are observed locations, virtue sensor \#1, 2, and 3 are unobserved locations need to estimate.  However, the situations of virtue sensor \#1 and \#3 are more challenging: (1) influenced by nonequilibrium flows, the readings of virtue sensor \#1 and sensor \#2 are time-shifted; (2) as the number of missing flow values are greater than the number of conservation equation for virtue sensor \#1, \#3 and sensor \#3, their values are underdetermined.
\label{intro}
\end{figure}

Obtaining network-wide traffic volume over a period of time is nontrivial due to the limited deployment of expensive traffic volume sensors. Current sensor coverage rate hardly meets the demands of high spatial resolution of volume data. As traffic control strategy is generally performed at the network level \citep{aboudolas2013perimeter}, efficient and accurate traffic volume estimation approaches are therefore much-needed for analyzing multi-lane traffic network conditions. Distinct from traffic forecasting, in this paper we study network-wide multi-lane urban highway traffic volume estimation, which aims to predict traffic volume data at current time slot for the whole sensor network based on partial measurements of link traffic. This problem can be viewed as installing virtue volume sensors on road segments without sensors to enhance spatial resolution of link traffic. Fig. \ref{intro} gives an example to illustrate the studied problem.

In light of the preceding analysis, a collection of pioneering works have been dedicated to tackle this challenge and achieved valuable results \citep{aslam2012city,zhan2016citywide,meng2017city,tang2019joint,zhang2020network}. However, a majority of existing studies on this topic has focused on primarily validating the capability of a particular model to accurately estimate or forecast real-world traffic volume. The issue discussed much less than the overall accuracy is that of a model's ability of dealing with core challenges of volume estimation. We claim that some of the shortcomings of previous studies are not primarily their lack of overall accuracy, but more so their inferior performances on some critical and challenging scenarios, e.g., estimation of underdetermined and nonequilibrium traffic flows.

To elaborate the above view and emphasize the motivations of this paper, we discuss several inherent issues of network-wide traffic volume estimation that have not been fully solved in the literature:

\textbf{Modeling semantic relationships between sensors.} Estimating sensors' readings for unmeasured locations are typically cast into spatiotemporal traffic kriging in prior works \citep{appleby2020kriging,wu2021inductive,nie2023correlating}.
Most recent approaches that target performing kriging on traffic speeds adopt a simple distance metric or function (e.g., graph Laplacian) to model the pair-wise correlations between sensors \citep{takeuchi2017autoregressive,lei2022bayesian,nie2023correlating}. Nevertheless, the interplay and semantic similarities between traffic volume sensors are much more complex \citep{saeedmanesh2016clustering} and volume loaded on a network is strongly associated with the underlying environment such as road geometric factors, driving directions and POIs. 
Besides, the stochastic and non-stationary nature of traffic flow pattern further complicates this problem. 
Although graph neural network (GNN) has been widely leveraged to perform spatial modeling in existing works \citep{appleby2020kriging,wu2021inductive}, they are designed for general issues that are essentially different from the traffic estimation problem. Direct usage of an existing kriging model on this task might be less than satisfactory.
As a consequence, more advanced spatial relations modeling techniques and a tailored model architecture that can handle complex spatial semantic correlations as well as physical constraints simultaneously in the context of spatiotemporal traffic volume data are much-needed.

\textbf{Non-local spatial dependence for underdetermined traffic flows.} 
Due to the limited observations and existence of routine choice, not all link flows can be directly inferred from conservation equations (see Fig. \ref{intro} case 2). In terms of systems of linear equations, this road network is \textit{underdetermined} \citep{ng2012synergistic}. 
Underdetermined flows can arise on open paths which went fully undetected and could allow arbitrary volume values without violating the local flow conservation law if no further information is provided.
For instance, traffic participants usually travel to/from POIs settled in the middle of two adjacent detectors, and these undetected behaviors cause volume records of adjacent sensors large inconsistency, where neighboring information is inadequate for accurate recovery. Fortunately, it is demonstrated that traffic flow interdependence among road segments can exist and evolve in a wider range of network \citep{wu2019graph,wang2020forecast}. In this case, how to identify globally-related sensors to can provide referable information for the estimation of underdetermined sensors. 
 
 Current GNN based kriging models typically consider locally proximal detectors and aggregates available information only from $k$-th order neighborhoods \citep{appleby2020kriging,wu2021inductive}. Although a few techniques are developed for capturing spatial correlations globally on the graph for traffic forecasting models \citep{wu2019graph,yu20193d,bai2020adaptive}, the requirement of complete historical data for recognizing hidden patterns renders them unfeasible for traffic volume estimation problem. As traffic speed and traffic volume are linked by traffic density, they usually co-evolve under certain rules. Two road segments with similar road network context and traffic speeds are prone to exhibit similar volume patterns, despite far distance on the graph. Compared to cross-sectional volume that requires careful and laborious vehicle counting, average spot speed or average link speed are much more accessible \citep{meng2017city}. By exploiting shared speed patterns among roads, well-established speed-volume relationships can be utilized to offer hints for estimating underdetermined flows, beyond the description of spatial proximity.
   
 % Confining spatial receptive field to a local scope could amplify the impacts of extra noise brought by irrelevant neighboring features. 

 % On the one hand, stacking more GNN layers can enlarging the receptive field in space dimension, but the over-smoothing problem \citep{chen2020measuring} undermines its effectiveness in application. 
 % On the other hand, although a few techniques are developed for capturing spatial correlations globally on the graph for traffic forecasting models \citep{wu2019graph,yu20193d,bai2020adaptive}, the requirement of complete historical data for recognizing hidden patterns renders them unfeasible for traffic volume estimation problem. Therefore, extracting spatial correlation that beyond the description of spatial proximity is of vital importance for coping with such underdetermined scenarios.

% \textbf{Exploiting dynamic, nonlinear speed-volume relations.} 

However, modeling speed-volume relationship has been studied for decades yet is far beyond solved. Most prior works assume speed-volume vary by locations linearly, and the correlations of speed patterns approximate to ones of volume patterns \citep{zhan2016citywide,meng2017city,zhang2020network}. In these works, they elaborate specific rules (e.g., Pearson correlation coefficient, or static adjacent graph) to simplify this problem. In practice, nevertheless, these correlations are found to be nonlinear \citep{zhu2022multitask}, heterogeneous \citep{ji2012spatial,ramezani2015dynamics} and dynamically changing over time \citep{yao2019revisiting}. For example, different congestion formation and dissipation patterns across the traffic network could produce heterogeneous macroscopic fundamental diagrams with multiple kinds of speed-volume curves. And different traffic demands during rush hours and off-peak hours lead to various intensities of correlations.
Consequently, open questions remain about how to understand the entangled relationships between speed and volume in a transferable, adaptive, and elegant routine without exogenous assumptions on traffic flow characteristics, such as a fundamental diagram \citep{geroliminis2008existence,daganzo2011macroscopic}.

\textbf{Time-asynchronous correlations for nonequilibrium traffic flows.} Intrinsic traffic flow phenomena are strongly associated with high-resolution volume estimation, and few studies go further on this issue. Apart from the inductive bias (or misfit) brought by machine learning models themselves, the behaviors of traffic flow could also affect the estimation accuracy to a large extent.
Specifically, if the traffic density between two sensors keeps constant over space and time, then the incoming and outgoing flux/flow of each upstream and downstream sensor pair should be identical or matched at the same time slot, which is called  \textit{equilibrium flow} (see Fig. \ref{intro} case 1). Otherwise, there exists \textit{nonequilibrium} traffic flows \citep{zhang1998theory,zhang2009conserved} caused by asynchronous response time of drivers, manifesting as the macroscopic congestion propagation of traffic flow. The affect of traffic congestion propagates from upstream segment to downstream segment with a time delay, and this propagation is determined by the road network topology and traffic speed \citep{ji2014empirical}. Such nonequilibrium flows could make the volume dependence of sensors inactivated within a short time window, or in other words, generating time-asynchronous correlations between upstream and downstream sensors (see Fig. \ref{intro} case 3). However, as presented in \citep{yang2023traffic}, recognizing a reasonable scope to consider the influence of propagation delay is not an easy task. And such delayed temporal correlations often vary by space and time, associating with the current traffic conditions. A prescribed value may be somewhat subjective.
Besides, temporal correlations are often ignored or considered as a feature extraction module independent of spatial features in existing kriging or imputation works \citep{appleby2020kriging,wu2021inductive,liang2022memory}, which are inadequate for consideration of time-asynchronous correlations.

Motivated by the above research gap, this study contributes to the state-of-the-art with novel perspective and methodologies. A GNN framework is proposed to serve as a data-driven, equation-free, and correlation-adaptive link traffic volume estimator.
In summary, our contributions are four-fold:
  \begin{itemize}
      % \item We model the traffic volume estimation as a kriging problem and tailor a task-specific diffusion convolution network (TDCN) to account for the particular characteristics of traffic volume. Multiple factors such as road network geometry and operations of traffic flow are handled, so that (the semantic relationships between nodes can be well mined. )
      \item We model the traffic volume estimation as a kriging problem and identify two critical problems that affect the volume estimation significantly, i.e., underdetermined and nonequilibrium traffic flows. Two quantitative metrics are developed to measure their impacts.
      \item To improve the performance on underdetermined flows, we quantify the dynamic and nonlinear speed-volume relationships by proposing a graph attention based speed pattern-adaptive adjacency matrix (SPAM) construction module. After incorporating SPAM into graph convolution, our model can capture both local and non-local spatial relations simultaneously. 
      \item To consider the impacts of nonequilibrium flows, we formulate a temporal masked and localized attention (Tatt) module to learn the potential impacts of previous congested traffic states. Collaborating with a gated temporal convolution network (TCN), the time-asynchronous correlations between neighborhoods are well captured.
      \item Experiments are conducted on a real-world highway traffic volume dataset, and results indicate the state-of-the-art performances of our model on both overall accuracy and critical scenarios. Besides, ablation studies, sensitivity analysis, as well as interpretations of sample rate and learned graphs, further justify the rationality of model designs.
  \end{itemize}

The rest of this article is organized as follows. Section \ref{Literature review} briefly reviews related works about traffic state estimation. Section \ref{Notations and Problem Definitions} describes basic concepts and formulates the problem. 
Section \ref{methodology} introduces our model architecture and details. In Section \ref{experiments}, we evaluate our model on a real-world dataset and ablation studies, sensitivity analysis as well as model interpretations are provided. Section \ref{conclusions} concludes this study and provides future directions.

\section{Literature review}\label{Literature review}
This section revisits the status quo about traffic data modeling and state estimation. This review covers three aspects: 1) traffic volume estimation methods based on heterogeneous data fusion;
2) spatiotemporal kriging models and graph neural networks for traffic data modeling; 3) dynamic traffic assignment and link flow observability methods.

\subsection{Traffic volume estimation by data fusion} 
Unlike like traffic state imputation problem centering on estimating missing elements of multivariate time series, traffic volume estimation for locations without volume detectors is more crucial but challenging. In general, there exists two two categories of volume estimation models in the literature. The first is a fully supervised learning treatment \citep{aslam2012city,ide2016city,yi2019citytraffic,zhang2022full}, where various kinds of statistic models or machine learning models are established for volume regression with selected covariates. Another paradigm is the semi-supervised learning that graph-based approaches are desirable choices \citep{meng2017city,tang2019joint,yu2019citywide,zhang2020network}.
These pioneering works prove that fusing different sources of traffic data is a viable option for volume estimation \citep{xing2022traffic}. However, most of them either rely on strong assumption (e.g., specific locations of volume sensors \citep{rostami2020state}, availability of historical average data \citep{zhang2022full}), or extra data that is laborious to collect (e.g., probe vehicle counts \citep{aslam2012city}, taxi trajectories \citep{meng2017city,yu2019citywide}), undermining their utility and effectiveness in real-world applications.

% \subsection{Traffic volume estimation considering speed-volume correlations}
Particularly, there are a few works achieve this task by considering speed-volume relationships.
It is widely acknowledged that there is a potential link between traffic speed and volume. Therefore, it is rational to attempt to infer volume from speed. 
To this end, some studies model spatial dependencies of traffic volumes with speed information. \cite{meng2017city} and \cite{zhang2020network} assume that traffic speed and volume co-evolve in a linear relationship and locally adjacent road segments/sensors with similar speed profiles share similar volume patterns. \cite{zhan2016citywide} describe this relationship by a calibrated fundamental diagram. \cite{mahajan2022predicting} predicts volumes from speeds directly with a deep learning model. In addition, \cite{zhu2022multitask} design a speed-volume joint imputation framework by setting a shared feature extraction and learning module, achieving volume reconstruction from speed. However, their method depends on clear correlations between volume and speed that a Greenshields model can be well-fitted. As shown in \citep{zhu2022multitask, mahajan2022predicting}, the estimation performances degrade dramatically when the speed-volume relationships change. To fully exploit the speed-volume relationship and filter the most useful speed information, a data-driven approach is preferable.

\subsection{Graph neural networks and low-rank kriging models for network-wide traffic flow modeling}
GNNs have driven giant advances in prior works for temporal and spatial analysis. In the literature, GNNs have been broadly adopted for traffic state forecasting \citep{yu2017spatio,li2017diffusion, yu2017spatio,wu2019graph,wang2020forecast} and imputation \citep{ye2021spatial,liang2022spatial,liang2022memory,chen2022novel}. Among these works, \cite{li2017diffusion} developed a diffusion convolution networks that model the diffusion process of traffic flow as random walks and enable ones to build GNNs for traffic flow on directed graph. \cite{wu2019graph} further generalized the diffusion convolution model and proposed an adaptive graph learning module. These pioneering studies have a wide spectrum of applications and shed light on the versatility of GNNs to model network-wide traffic flows. Whereas, different from forecasting task with sufficient historical data for model training, volume estimation problem considered in this paper is much more tricky because indirect information is treated as predictor variables for unmeasured locations.

In the sphere of geostatistics, the problem of estimating values of unobserved locations is usually cast as spatial kriging (spatial regression). Most recently, kriging models have aroused great interests in transportation domain \citep{takeuchi2017autoregressive,wu2021inductive,wu2021spatial,liang2022spatial,lei2022bayesian,nie2023correlating}. To demonstrate the inductive power of GNNs, \cite{wu2021inductive} first proposed a kriging convolution network for spatiotemporal traffic speed. \cite{wu2019graph} and \cite{liang2022spatial} further improve the model performances with more advanced architectures. The key rationale of traffic kriging methods is to design a kernel function to capture spatial relationships between traffic states at different locations.
However, almost all of them focus on traffic speeds that can be fully characterized by simple and static graphs. Unfortunately, traffic volume behaves much more complicated than speed that simplified and static rules are not applicable to model the correlations. As traffic volume is strongly associated with the current traffic condition and many other factors, direct usage of a kriging model on traffic volume estimation could produce undesirable results.

\subsection{Dynamic traffic assignment and network flow conservation methods for link flow inference}

In the area of transportation research, a typical link traffic estimation method falls into the category of traffic assignment. In general, dynamic traffic assignment (DTA) solves the travel times for each time interval of each road segment with user equilibrium assumptions and all-day link flows can also be obtained \citep{janson1991dynamic}.

Despite more factors can be considered and more detailed information can be produced, DTA has stringent requirements for inputs. On the one hand, a complete picture of traffic demand is needed, but collecting all-day traffic demand is quite difficult (e.g., origin-destination survey). On the other hand, as the computation of DTA contains a path enumeration process \citep{wang2020estimating,zhang2020path}, a large-scale vehicle path dataset is sometimes necessary. Therefore, origin-destination matrix estimation \citep{krishnakumari2020data}, or path reconstruction \citep{yang2015vehicle} procedure which are both intractable, could be a prerequisite. In DTA, the relationship of travel speed (time) and link volume is established by calibrating a Bureau of Public Roads (BPR) power function \citep{merchant1978model}. This simplified assumption could deviate from the reality. In the meanwhile, as DTA method usually bases on path assignment at the lower level of a bi-level optimization model, these analytical or simulation solutions require high computational burden \citep{ros2022practical}. 

% \subsection{Link flow inference considering network flow conservation and sensor location}
There is also another branch of link flow inference research that is of wide interest, i.e., the full link traffic flow observability problem \citep{ng2012synergistic}. The goal of this problem is to identify the minimum number and combination of link sensors, to infer the link traffic for the whole network, using conservation equation only \citep{viti2014assessing,xu2016robust}. 
We would like to clarify that our focus is different from these works. Link flow observability problems place emphasis on the flow distribution of a static network, and the time dimension is ignored. They can select the locations of sensors and usually assume intact observations without noises. The volume estimation is implemented by solving conservation equations and macroscopic simulation is widely used for validation \citep{salari2019optimization,shao2021optimization}. 

While we consider more realistic scenarios and data collected in real-world traffic measurement systems, where the conservation equations may be inadequate.
We estimate the all-day network-wide link flow series for a long time period in the perspective of kriging and data reconstruction, given the detector already installed. Besides, our work targets at higher-resolution sensor-level measurements and the sensor network, which is a basic unit for modern traffic control and management strategies.

\section{Preliminaries and problem descriptions}
\label{Notations and Problem Definitions}

In this section, we first introduce basic concepts used herein and formulate the traffic volume estimation problem explicitly. Then we present the rationale for how GNN can treat traffic volume estimation as a kriging problem.

\subsection{Graph representation of traffic network}
In this work, we formulate the network-wide traffic volume estimation task as a graph modeling and reconstruction problem. As traffic flows load on a road network that connects related detectors rather than a 2D grid, we can abstract their relationships using a directed graph. Please note that we use the terms "sensor", "detector", and "node" interchangeably throughout the paper. A sensor network can be modeled as a weighted directed graph $\mathscr{G}=\{\mathscr{V},\mathscr{E},\mathbf{A}\}$, where vertices $\mathscr{V}$ denote $N$ sensors or detectors of the network, edges $\mathscr{E}$ indicate the connectivity between nodes and their directions are assigned according to the moving directions of traffic flow. $\mathbf{A}\in\mathbb{R}^{N\times N}$ is the weighted adjacency matrix (weights of edges) describing the correlations of nodes, and it is usually treated as a function of network topology or travel distance.

In the literature, a commonly used distance-based adjacency matrix $\mathbf{A}$ is computed as:
\begin{equation}\label{Gaussian}
    a_{ij}=\left\{\begin{array}{l}\operatorname{exp}\left(-\left(\operatorname{dist}(v_i,v_j)/\delta\right)^2\right),~\text{if } \operatorname{dist}(v_i,v_j)\leq \epsilon, \\
    0,~\text{otherwise}, \\    
    \end{array}\right. \\
\end{equation}
where $\operatorname{dist}(\cdot)$ is the travel distance between two sensors, $\delta$ is a scaling parameter which can be selected as the standard deviation of distances, and $\epsilon$ is the threshold.
As we can seen, $\mathbf{A}$ is asymmetric because the sensor graph is directed.

Traffic flow operating on the graph can be 
recorded as graph signals. We use $\mathbf{X}_v \in\mathbb{R}^{N\times T}$ and $\mathbf{X}_s\in\mathbb{R}^{N\times T}$ to represent traffic volumes and speeds produced at $N$ locations during $T$ time intervals. Accordingly, each element like $x^t_{v_n}$ is the volume of location $n$ at time window $t$. To obtain a fine-grained spatiotemporal traffic volume estimation, we set the time window size as 5 minute in this work.

\subsection{Network-wide traffic volume estimation with speed information}
This paper centers on performing spatial estimation of traffic volume at locations without volume detectors and we can discuss this problem in the context of spatial kriging \citep{nie2023correlating}. Let $\mathscr{V}^o$ be the locations with volume sensors, $\mathscr{V}^u$ be the unobserved locations, and $n_o=\vert\mathscr{V}^o\vert$, $n_u=\vert\mathscr{V}^u\vert$ are number of observed/unobserved locations, the goal of traffic volume estimation/kriging is:
\begin{equation}
\begin{aligned}
    &\widehat{\mathbf{X}}_v=\mathscr{F}(\mathbf{X}_v^i,\mathbf{X}_s^j,\mathscr{G}), ~i\in\mathscr{V}^o,j\in\mathscr{V}^o\cup\mathscr{V}^u,\\
    &\text {s.t.}~ \widehat{\mathbf{X}}_v^i=\mathbf{X}_v^i,\\
    \end{aligned}
\label{model_form}
\end{equation}

$\widehat{\mathbf{X}}_v\in\mathbb{R}^{N\times T}$ is the estimated volume of entire graph and $\mathscr{F}$ is the targeted neural network model. In this paper, we keep the same assumption as in \citep{meng2017city} and \citep{zhang2020network}, that traffic speed records are supposed to be complete and available for all road segments (sensors). Actually, this assumption is easy to satisfy in reality. On the one hand, a full picture of traffic speeds can be readily measured and recorded by float cars or crowdsourcing vehicles \citep{liu2019spatial,zhang2020network,yu2020urban}. Such Lagrangian measurements can account for a large proportion of urban road network, with a high data updating frequency. By conducting a map-matching procedure, the speed data is applicable in practice. 
% there is a higher possibility that traffic volume records are unavailable than traffic speeds when equipment failures occur in traffic measurement system. 
% Considering the fact that volume values are derived from the total number of passing vehicles detected by sensors during a specific time interval, while speed records are obtained by averaging the spot speeds, when detector is corrupt, e.g., missing detection, the reported speed values can still be available but volumes may be far from truth values.
On the other hand, recent studies aiming at estimating network-wide traffic speed have demonstrated that full-scale of speeds can be recovered with high accuracy, having limited amount of observations \citep{lei2022bayesian,nie2023correlating,wu2021spatial,liang2022spatial}. Given partially observed traffic volume and speed, high-quality and complete speed data can be estimated at first, using an already established spatiotemporal speed kriging model. 

% \subsection{Spatial smoothness and temporal alignment metrics}\label{sec_SSI_TAI}

\subsection{Treating volume estimation as kriging}\label{estimation_kriging}

\begin{figure}[!htb]
\centering
\centering
\includegraphics[scale=0.52]{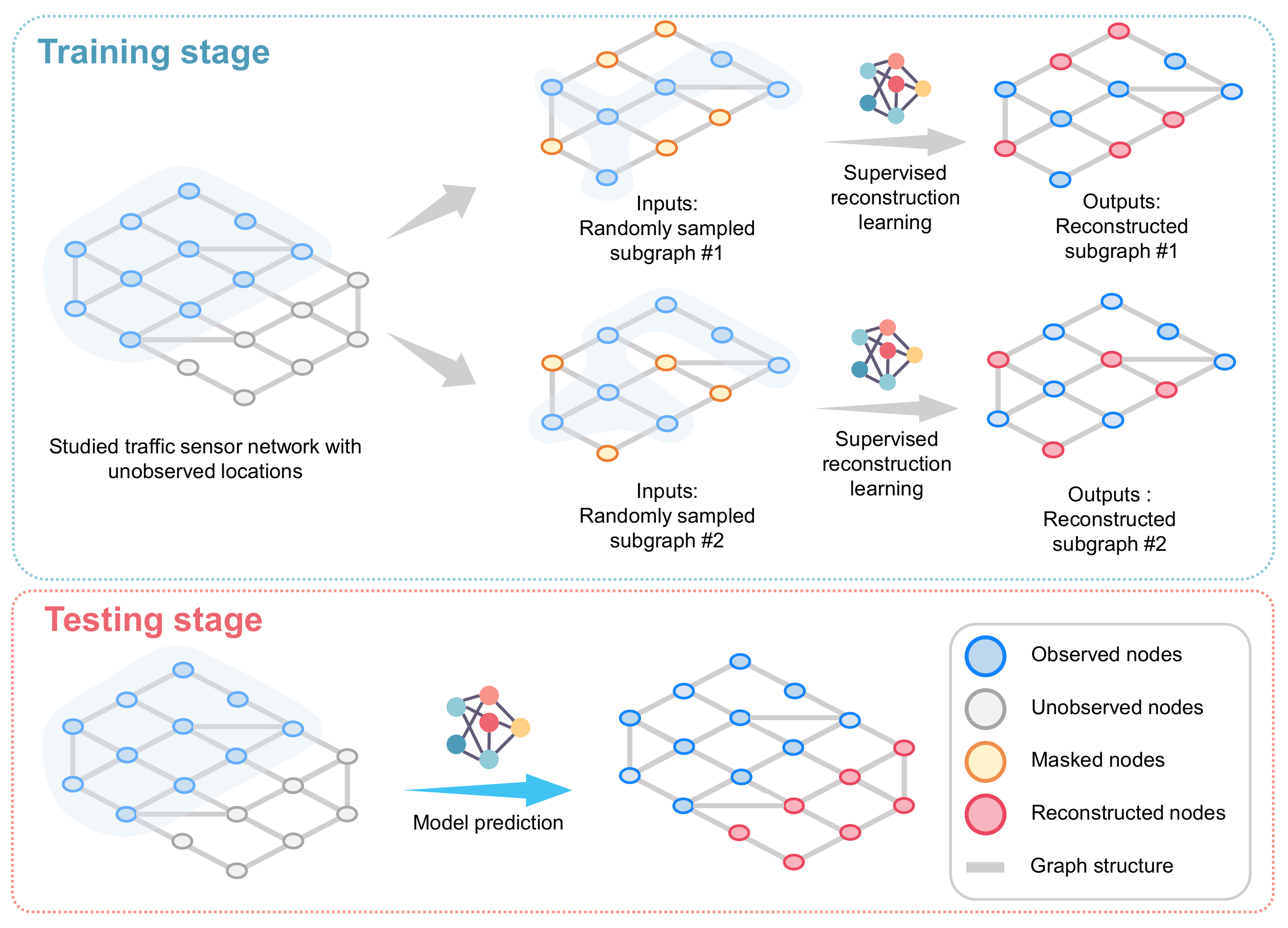}
\caption{Explanation of the working process of volume kriging model. Only observed graph information (data and location) is provided during training stage. By randomly sampling different sub-graphs, the model can inductively learn to reconstruct the whole graph and generalize to unseen nodes. A trained model can make predictions at newly added locations, even if they do not appear in the training graph.}
\label{graph_recon}
\end{figure}

To solve Eq. \eqref{model_form}, in this work we develop a GNN kriging architecture to handle highway traffic volume estimation problem, and our model belongs to the category of kriging convolution networks (KCN) \citep{appleby2020kriging,wu2021inductive}. Typically, KCN is used for spatial modeling of non-cumulative data, such as temperature and traffic speed. To explain the rationality of kriging on traffic volume, in what follows we will first demonstrate how our model can be used for volume estimation under kriging settings.

Generally, volume estimation problem can be solved by a regression method in which some selected independent variables such as historical traffic measurements, traffic demand, network features, as well as time information, are used to fit the target volume value \citep{aslam2012city,ide2016city,zhang2022full}.

\begin{equation}\label{reg}
\begin{aligned}
    \mathbf{X}^i &= \operatorname{AGG}(\{\mathbf{x}_j:j\in\mathcal{N}_i\}\cup\{\mathbf{z}_j:j\in\mathcal{F}_i\}) ,\\
    \mathbf{Y}^i &= \operatorname{REG}(\mathbf{X}^i,\mathcal{G}),\\
    \end{aligned}
\end{equation}
where $\operatorname{AGG}(\cdot)$ denotes a parameterized function that aggregates all available information from the neighborhood $\mathcal{N}_i$ and relevant features $\mathcal{F}_i$ of node $i$. respectively. Then combining with the graph information $\mathcal{G}$, $\operatorname{REG}(\cdot)$ produces the target volume values $\mathbf{Y}^i$.

To use as few external features as possible, which are usually laborious to collect, e.g., full-scale traffic demand, the input volume value can become as the only feature to consider, and the label is also the true volume value, problem \eqref{reg} turns into the following:
\begin{equation}
\begin{aligned}
    \min_{\Theta}&\sum_{i\in\mathscr{V}}\operatorname{LOSSFUNC}(\mathbf{X}^i,\hat{\mathbf{X}}^i),\\
    \hat{\mathbf{X}}^i&=\operatorname{GNN}(\mathbf{X},\mathbf{M},\mathcal{G}\vert\Theta),
\end{aligned}
\label{gnn_reg}
\end{equation}
where $\Theta$ is the model parameter needs to learn, $\operatorname{LOSSFUNC}(\cdot)$ measures the discrepancy of estimated and true values.
Here, GNN serves as the predictive function to produce estimation. By this supervised learning setting, GNN learns the message aggregation mechanism on the graph and can generalize to new node if its location is given. We can see that problem \eqref{gnn_reg} has equivalent form to problem \eqref{reg}, with specific input features. In this work, we only choose the historical traffic speed data and network geometries as external features, which are readily available for most urban road networks.

One can find that in this volume kriging problem, the labels are true volume data for all locations, which are also the inputs to the GNN model. However, directly feed the labels into model training will produce a trivial model without learning anything. Therefore, a random sampling matrix $\mathbf{M}$ is introduced to generate diverse training samples, see Fig. \ref{graph_recon}. In this sense, this problem resembles a graph reconstruction learning process.

A trained or calibrated volume estimator should have the ability to predict the label of new test data point at any location using the observed training labels directly. Fortunately, the inductive power of GNN makes it possible \citep{wu2021inductive}. Besides, the training of GNN in Eq. \eqref{gnn_reg} allows missing values of the observations, and real-time implementation is also feasible.

Although treating volume estimation as kriging is reasonable, however, three key characteristics of highway traffic volume estimation problems hinder the direct usage of a KCN on this topic. In short, traffic flow characteristics and underdetermined problem will affect $\operatorname{AGG}(\cdot)$, e.g., aggregating locally proximal information is insufficient when neighbor sensors vary greatly in space,
and nonequilibrium phenomenon will impact $\operatorname{REG}(\cdot)$, e.g., the inputs and labels are time-shifted on temporal dimension. 
In the rest of this section, we will describe how these problems affect the estimation performance and how our model designs tackle such challenges.

\section{Methodology}\label{methodology}
In this section, we first develop two metrics that are used to evaluate the undetermined and nonequilibrium problems of highway traffic volume estimation. In what follows, we elaborate the proposed \textbf{S}patio\textbf{T}emporal \textbf{C}orrelation-\textbf{A}daptive \textbf{G}raph \textbf{C}onvolution \textbf{N}etworks (STCAGCN) model. 
The core concept of STCAGCN is to make use of the redundancy of sensor networks and rich correlations of traffic volumes, i.e., unobserved volumes can be estimated by exploiting spatial relationships from both globally or locally correlated sensors, and time-asynchronous correlations from previous traffic states.
%We first introduce the overall model architecture and then give descriptions about each component in details.

\begin{figure}[!htb]
\centering
\centering
\includegraphics[scale=0.52]{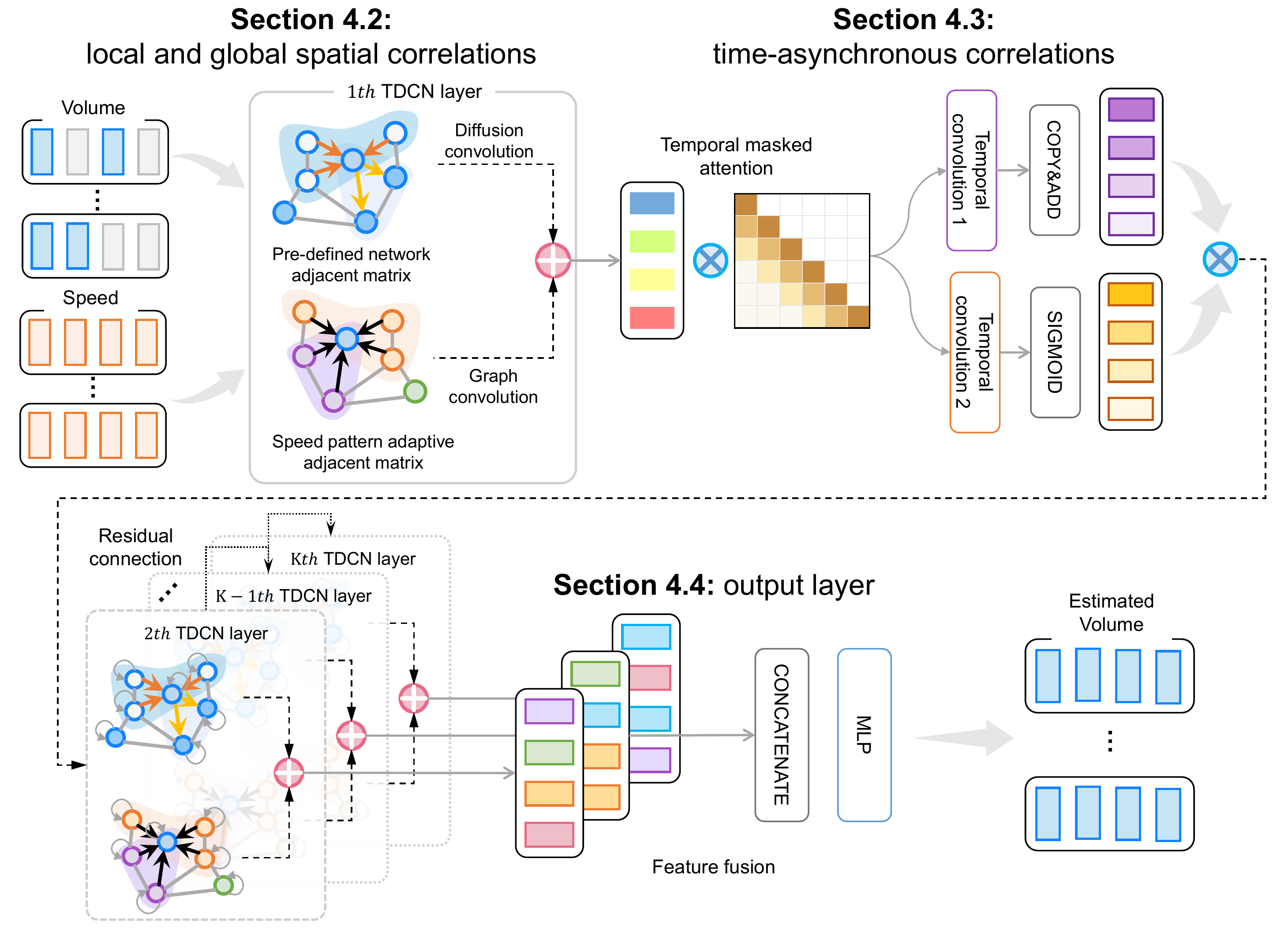}
\caption{Overall architecture of proposed STCAGCN model. One initial TDCN layer, one Tatt-TCN layer and several stacked TDCN layers are applied. The inputs are masked traffic volume and complete traffic speed data, and STCAGCN reconstructs the complete volume for whole network.}
\label{STCAGCN_arc}
\end{figure}

Fig. \ref{STCAGCN_arc} shows the holistic architecture of STCAGCN. It consists of two basic components: 
% (1) spatial information aggregation block for filtering valid features applied to determined and equilibrium flows,
(1) spatial information aggregation block for alleviating underdetermined problem with learned speed-volume relationships.
(2) time-asynchronous correlations extraction block for coping with nonequilibrium flows.

\subsection{Measuring undetermined and nonequilibrium phenomena}\label{sec_SSI_TAI}

As introduced in section \ref{Introduction}, we attribute the core challenges of volume estimation to underdetermined and nonequilibrium traffic flows. To examine the existence and evaluate the degree of these phenomena, we first develop the spatial smoothness and temporal alignment metrics.

A common belief is that adjacent sensors report similar traffic volumes because flows are supposed to be conserved if vehicles only traverse between them. This can be recognized as a spatial stationary property of traffic network \citep{zhang2020network}. However, traffic participants often travel to/from middle POIs of urban networks or on-off ramps of highway networks. If there are only a few observations, the sensor network can be viewed as a underdetermined linear system and volume values can not be obtained from conservation laws directly. The spatial smooth index is thereby adopted to measure this nonstationary phenomenon.

Following \citep{zhang2020network}, a weighted and directed spatial smoothness index is modified as follows:

\begin{definition}(Weighted and directed spatial smooth index): The weighted and directed spatial smooth index (WDSSI) quantifies the spatial local stationary (or conservation) property of the traffic volume loaded on a road network based on the ground truth volume values and network geometry, which can be formulated as:
\begin{equation}\label{SSI}
    \operatorname{WDSSI}(i) = \frac{1}{T}\sum_{t=1}^T{\left\vert \frac{\sum_{j\in\mathscr{N}_i}a_{ij}x^t_{v_j}}{\sum_{j\in\mathscr{N}_i}a_{ij}}-x^t_{v_i}\right\vert}/x^t_{v_i}, ~\forall i \in \mathscr{V},
\end{equation}
where $\operatorname{WDSSI}(i)$ is the value for the $i$-th location of the sensor network as well as the $i$-th node on the affinity graph, $\mathscr{N}_i$ is the set of neighbors of node $i$ either on inflow or outflow direction, \minew{and $T$ is the total observation period.}
\end{definition}

WDSSI considers the \minew{high-order} proximity and observability of a sensor network. \minew{For highway corridors and trunks,}
if the road segment between a sensor and its adjacent sensors is fully conserved, the WDSSI value should close to zero. In highway or freeway network, there exists a large amount of on/off ramps that are not equipped with sensors. Such positions could cause large discrepancy of the measurements between adjacent sensors, and we use WDSSI to identify the exist of this underdetermined issue. \minew{More interpretations on WDSSI can be found in \ref{Appendix_C}.}

Dynamic time warping (DTW) is a widely adopted distance metric for time series data that can be used to compare time series of different lengths and rhythms \citep{rakthanmanon2012searching}. Generally speaking, if two time series are not time-aligned, timestep based distance metrics like Euclidean distance are inapplicable. In this case, DTW will automatically warp the time series (i.e., local scaling on the time axis) so that the morphology of the two series is as consistent as possible to obtain the maximum possible similarity. Based on dynamic programming algorithm, the DTW calculation is given as follows.
\begin{definition}(Dynamic time warping distance):
Given two time series $\mathbf{x}=(x_1,x_2,\dots,x_n)$ and $\mathbf{y}=(y_1,y_2,\dots,y_m)$ with length $n$ and $m$, let $d_{ij}=(x_i-y_j)^2$ be the Euclidean distance of two series points, then the accumulated distance $\gamma(i,j)$ can be defined as:
\minew{    
\begin{equation}\label{dtw}
\gamma(i,j)=d_{ij}+\min(\gamma(i-1,j-1),\gamma(i-1,j),\gamma(i,j-1)),~1\leq i\leq n,~1\leq j\leq m,
\end{equation}}
i.e., the sum of the Euclidean distance of points i and j and the cumulative distance of the smallest neighboring elements that can reach this point. After the increase in $i,j$, the final accumulated distance $\gamma(n,m)$ has the best alignment, and the accumulated distance along the warping path is minimized. \minew{Then the final accumulated distance $\gamma(n,m)$ is termed as the dynamic time warping distance (DTW) between the two series.}
\end{definition}

Under this definition, if congestion delay occurs between two sensors, their recorded signals are supposed to be time-asynchronous and should have small DTW distance but large Euclidean distance.
With regard to this, we propose a time alignment indicator (TAI) to examine if two time series from upstream and downstream are time-asynchronous (i.e., nonequilibrium flows occur in the middle road segment). The smaller the TAI is, the higher possibility the target sensor is influenced by traffic propagation delay.

\begin{definition}(Time alignment indicator): The time alignment indicator quantifies the temporal local alignment (or asynchronous correlation) of volume between upper- and lower-stream detectors:
\minew{
\begin{equation}\label{TAI}
\operatorname{TAI}(I)=\frac{\operatorname{DTW}(I,J)}{\operatorname{EUC}(I,J)},~J\in\mathscr{N}_I^u,
\end{equation}}
% where $\mathscr{N}_i^u$ is the nearest upstream neighbor of node $i$, $\operatorname{DTW}(\cdot),\operatorname{EUC}(\cdot)$ are DTW and Euclidean distance matrix respectively. 
\minew{where $\mathscr{N}_I^u$ is the nearest upstream neighbor of node $I$, $\operatorname{DTW}(I,J)$ is the DTW distance between sensors $I,J$ calculated from Eq. \ref{dtw}, and $\operatorname{EUC}(I,J)=\sqrt{\sum_i(\mathbf{x}_I(i)-\mathbf{x}_J(i))^2}$ is the Euclidean distance of two volume readings.}
\end{definition}

These two metrics help us identify and understand the occurrence of underdetermined and nonequilibrium problems, elaborate corresponding solutions, and evaluate model performances on each task.
% \subsection{Overall architecture of STCAGCN}

\subsection{Modeling both local and global spatial relations for estimation with underdetermined flows}
\subsubsection{Task-specific diffusion convolution for local spatial dependency}
Graph convolution networks (GCNs) have been widely applied in recent spatial modeling works \citep{li2017diffusion,yu2017spatio,wu2019graph,xue2022quantifying}. By aggregating features from neighbor nodes on the graph, GCNs transform graph signals in feature space and model local spatial dependencies in a data-driven routine. For any node on a sensor graph, both inflows and outflows decide its instant traffic state. Therefore, directed nature of real-world traffic networks encourages a model that can deal with asymmetric adjacency matrix \citep{nie2023correlating}.

Diffusion convolution network (DCN) proposed by \citep{li2017diffusion} provides an effective tool for this problem. DCN models the propagation of traffic flow as a finite step diffusion process to facilitate convolution on directional graphs. According to \citep{wu2019graph}, for an input graph signal matrix $\mathbf{X}\in\mathbb{R}^{N\times C_{in}}$, the diffusion convolution operator can be generalized into the following form:
\begin{equation}
\mathbf{Y}=\sum_{k=0}^K\left(\mathbf{A}_f^k\mathbf{XW}_{kf}+\mathbf{A}_b^k\mathbf{XW}_{kb}\right),
\end{equation}
where $\mathbf{A}_f^k,\mathbf{A}_b^k$ are the forward and backward transition matrix at $k$-th diffusion step, $K$ is the total steps of diffusion, $\mathbf{W}_{kf},\mathbf{W}_{bf}\in\mathbb{R}^{C_{in}\times C_{out}}$ are learnable parameters of DCN layer. The bidirectional transition matrices can be obtained from the adjacency matrix of sensor graphs:
\begin{equation}
    \mathbf{A}_f=\frac{\mathbf{A}}{\operatorname{rowsum}(\mathbf{A})},
    ~\mathbf{A}_b=\frac{\mathbf{A}^\mathsf{T}}{\operatorname{rowsum}(\mathbf{A}^\mathsf{T})}.
\end{equation}

$\mathbf{A}_f$ and $\mathbf{A}_b$ are in fact the row-wise normalization of $\mathbf{A}$ and $\mathbf{A}^\mathsf{T}$, in order to sum the weights to one. This diffusion convolution operator can take into account the influence of inflows and outflows simultaneously.

Recall that the propagation of traffic flow is directed and along each travel direction, the basic DCN only considers inflows and outflows of nodes, but ignores the fact that traffic states from opposite moving direction could be quite different. Considering this, we modify the basic DCN with a Task-specific Diffusion Convolution Network (TDCN) in three aspects: diffusion direction, setting of self-loop, and lane-wise normalization. We will demonstrate the effectiveness of these simple but useful modifications in section \ref{experiments}.

% As discussed in section \ref{Literature review}, DCN has driven tremendous achievements in traffic speed forecasting and kriging applications. Almost all these prior works construct the adjacency matrix $\mathbf{A}$ in Eq. \eqref{Gaussian} by computing the Euclidean distance or travel distance between sensors in a direction-agnostic way. 
\minew{While DCN has proven effective at capturing spatial correlations between upstream and downstream sensors, there are some limitations in how existing studies have applied it in practical settings. First, previous works construct the adjacency matrix $\mathbf{A}$ by computing the Euclidean distance or travel distance between sensors in a direction-agnostic way, where sensors from different driving directions and upstream-downstream segments are considered indiscriminately. This treatment can be inadequate for volume that is strongly associated with road network properties and the analysis of \citep{wu2019graph} reveals that the correlation intensity of the sensors from opposite directions can be distinct. To avoid spurious correlations, we confine the diffusion process to the lane where the detector is located. Specifically, the proposed TDCN only aggregates neighboring features along a specific driving direction.} As a consequence, a new asymmetric adjacency matrix $\widetilde{\mathbf{A}}$ can be constructed as follows:
\begin{equation}
    \widetilde{a}_{i,j}=\left\{\begin{array}{l}a_{ij},~\text{if } i~\text{and }j ~\text{are in the same driving direction }, \\
    0,~\text{otherwise}. \\    
    \end{array}\right. \\
\end{equation}

\minew{Second}, historical records are important input features for traffic forecasting task, so after fusing available information from neighborhood detectors, the original features/inputs of each sensor itself need to be incorporated and the mixed features are feed into the next GCN layer. Given this, the adjacency matrix $\mathbf{A}$ is supposed to contain a self-loop for each node \citep{yu2017spatio}. However, with regard to our kriging problem, if we feed the ground truth values of unobserved locations to the neural network during model training stage, this model would not learn any useful information but only outputs the input itself. Hence, TDCN has no access to the inputs of unobserved or masked locations (details about random masking will appear in subsection \ref{train}) in the first layer and the adjacency matrix of first-layer should not have self-loops.

\minew{Third,} another important impact factor of traffic volume is the number of lanes. Since cross-sectional volume directly relates with the lane number, a wider road segment with more lanes will record more passing vehicles within a time interval. Rather than using lane number as an indicator to predict volume directly, we normalize the input volume data by lane numbers thus per lane volume is used for training and testing.

After considering all these special characteristics of this task, we customize the TDCN layer as following:

\begin{equation}\label{TDCN_layer}
\begin{aligned}
\mathbf{H}^{1}&=\sum_{k=0}^K\left(\widetilde{\mathbf{A}}_{f0}^k\mathbf{X}\mathbf{W}_{kf0}+\widetilde{\mathbf{A}}_{b0}^k\mathbf{X}\mathbf{W}_{kb0}\right)\oslash\mathbf{e}, ~l=0\\
\mathbf{H}^{l+1}&=\operatorname{a}\left(\sum_{k=0}^K\left(\widetilde{\mathbf{A}}_f^k\mathbf{H}^l\mathbf{W}_{kf}+\widetilde{\mathbf{A}}_b^k\mathbf{H}^l\mathbf{W}_{kb}\right)\right), ~l\geq1\\
\end{aligned}
\end{equation}
where $\mathbf{e}\in\mathbb{R}^N$ denotes a vector whose element is the lane number of each sensor, $\oslash$ denotes the element-wise division, $\widetilde{\mathbf{A}}_{f0}$,$\widetilde{\mathbf{A}}_{b0}$ are adjacent matrices without self-loop, $\mathbf{H}^{l}$ is the output spatial features of $l$-th DCN layer, and $\operatorname{a}(\cdot)$ denotes the nonlinear activation function, e.g., ReLU function. By stacking several TDCN layers, GCN can capture multi-hop spatial correlations on the graph and the features of each node contain information from neighbor nodes. 

In practice, to encourage parallel computation and mini-batch treatment, the implementation of TDCN can be extended to high-dimensional tensors. For input batch of graph signal embedding $\mathcal{H}\in\mathbb{R}^{B\times N\times T \times C_{in}}$ with batch size $B$, number of nodes $N$, sequence length $T$ and feature dimensions $C_{in}$, Eq. \eqref{TDCN_layer} can be expressed as:
\begin{equation}
    \mathcal{H}^{l+1}=\operatorname{a}\left(\sum_{k=0}^K\widetilde{\mathcal{A}}_f^k\otimes\mathcal{H}^l\times_4\mathbf{W}_{kf}+\widetilde{\mathcal{A}}_b^k\otimes\mathcal{H}^l\times_4\mathbf{W}_{kb}\right), ~l\geq1
\end{equation}
where $\otimes$ is the batch-wise matrix multiplication operation, $\times_4$ denotes the mode-4 tensor-matrix product. 

As can be seen, Eq. \eqref{TDCN_layer} only aggregates localized spatial information and suitable for dealing with determined flow locations where neighborhood sensors can provide sufficient information. When confront with the underdetermined scenario, observations in the proximity of target sensor become less informative.

\subsubsection{Quantifying speed-volume relationships by speed pattern-adaptive graph construction module}

\begin{figure}[!htb]
\centering
\centering
\includegraphics[scale=0.5]{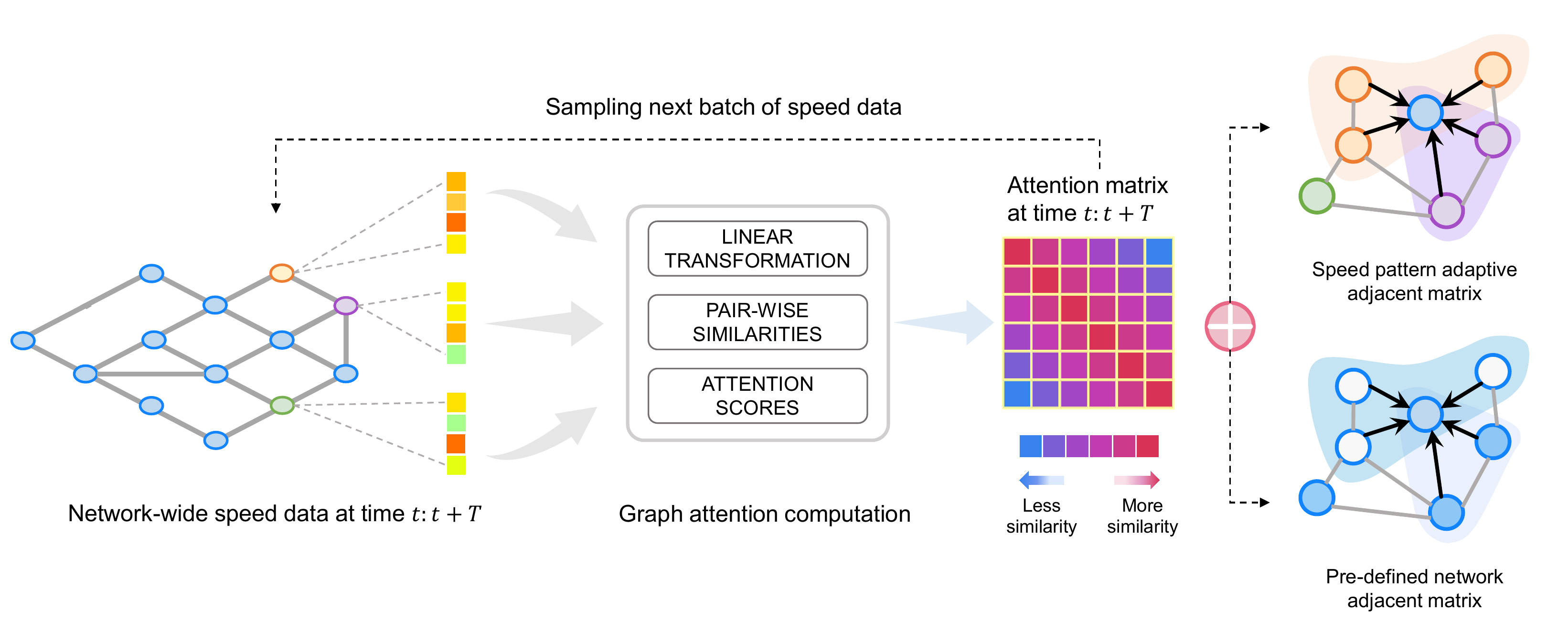}
\caption{Mechanism of speed pattern adaptive matrix construction module. SPAM is constructed for each input speed batch, thereby changing over time.}
\label{SPAM_arc}
\end{figure}

A shortcoming not to be overlooked of DCN is that it highly depends on the predefined adjacency matrix (based on distance metric or correlation analysis). The predefined graph could be subjective and cannot fully mirror the underlying semantic similarity of items. Previous works have confirmed that the specification of adjacency matrix has significant impacts on the performances of GCN \citep{wu2019graph}. To alleviate this restriction, many studies about traffic forecasting resort to the internal similarities of data and learn a semantic graph as a complementary \citep{wang2020forecast,bai2020adaptive,liang2022memory}. However, the adaptively learned graph in these works are acquired from historical input data and can not be generalized to our case that no historical data is available for unmeasured locations. Therefore, external data source like traffic speeds need to be incorporated as an auxiliary. As stated in section \ref{Notations and Problem Definitions}, in this paper we assume the historical speed records are available for all road segments. These speed information provide opportunities for revealing the underlying correlations of traffic volume.

A natural question comes up: \textit{how we use the underlying relationship between traffic speed and volume to help the estimation of underdetermined volume?} Modeling the relationship between speed and volume is a longstanding but open question in traffic studies \citep{weng2011modeling,fu2020empirical}. There are some pioneering works introduce the usage of simplified speed-volume relationship to help estimate the network-wide volume \citep{zhan2016citywide,meng2017city,zhang2020network,zhu2022multitask}, and their approaches can be divided into two categories: 1) adopting a deterministic model to describe this relationship; 2) estimating volume from speed directly. Both of these two considerations rely on a well-fitted speed-volume model (such as well-defined fundamental diagrams \citep{geroliminis2011properties}) and the estimation accuracy is significantly affected by the variance of speed-volume scatter. 
Moreover, as highway traffic states form a high-dimensional nonlinear dynamical system, with time-evolving spatial patterns \citep{wang2023anti}, these kinds of static and linear methods fail to reveal the dynamically varying, non-exclusive, nonlinear speed-volume relationships.

To circumvent these restrictions, we propose a Speed Pattern Adaptive adjacency Matrix (SPAM) learning module to quantify the underlying speed-volume correlations directly from data, which is shown in Fig. \ref{SPAM_arc}. 
Specifically, inspired by the graph attention (GAT) architecture \citep{velivckovic2017graph}, SPAM first computes the mutual importance of each node's speed features, and then mapping the similarity of speed features to the similarity of volumes via end-to-end stochastic gradient descent, discarding all structural and prior information about the sensor graph.

Given the input traffic speed for all sensors $\mathbf{X}=\{\mathbf{x}_1,\mathbf{x}_2,\dots,\mathbf{x}_N\},\mathbf{x}_i\in\mathbb{R}^D$, where $D$ is the feature dimension, $N$ is the node numbers, SPAM first computes the query vectors $\mathbf{q}$ and key vectors $\mathbf{k}$ by linear transform:

\begin{equation}
\begin{aligned}
    \mathbf{q}=&\mathbf{W}^q\mathbf{x}_i,\\
    \mathbf{k}=&\mathbf{W}^k\mathbf{x}_j.\\
\end{aligned}
\end{equation}

Then, an attention scoring function $\operatorname{att}:\mathbb{R}^D\times \mathbb{R}^D\rightarrow\mathbb{R}$ is used to calculate the attention scores, which represent the similarity or importance weight of node $j$'s feature on node $i$. Here, we select the scaled dot product \citep{vaswani2017attention} as the attention function. Followed by a normalization function, the attention scores are computed as:

\begin{equation}\label{GAT}
    \alpha_{ij}=\operatorname{Softmax}(\operatorname{att}(\mathbf{q},\mathbf{k}))=\operatorname{Softmax}(\frac{\mathbf{q}\cdot\mathbf{k}^\mathsf{T}}{\sqrt{d_k}}),
\end{equation}
where $d_k$ is a scaling constant.

Eq. \eqref{GAT} in effect computes the pair-wise similarity of each transformed speed vector and normalize it, making sure the sum is one.
In SPAM, we utilize a shared weight matrix $\mathbf{W}$ to convert the queries and keys to the same space, then apply a LeakyReLU activation function to add nonlinearity. And we use the product of $\mathcal{L}$-2 norm as the scaling constant. Formally, given the speed data $\mathbf{X}_s$ of all sensors, the SPAM $\mathbf{A}_{spa}$ as well as the attention weights matrix can be computed by:

\begin{equation}\label{graph_attention}
    a_{ij}^{spa}=\frac{\operatorname{exp}(\operatorname{LeakyReLU}((\mathbf{Wx_s}_i)^\mathsf{T}\cdot\mathbf{Wx_s}_j)/(\Vert\mathbf{Wx_s}_i\Vert\Vert\mathbf{Wx_s}_j\Vert))}{\sum_{j'}\operatorname{exp}(\operatorname{LeakyReLU}((\mathbf{Wx_s}_i)^\mathsf{T}\cdot\mathbf{Wx_s}_{j'})/(\Vert\mathbf{Wx_s}_i\Vert\Vert\mathbf{Wx_s}_{j'}\Vert))},
\end{equation}
where $\Vert\cdot\Vert$ is the $\mathcal{L}$-2 norm of vector. One can find that this attention matrix $\mathbf{A}_{spa}$ indicates the similarity between features $\mathbf{x_s}_i$ and $\mathbf{x_s}_j$, and it ranges from 0 to 1 with summation to 1, which can also be regarded as an adjacency matrix. \minew{Details about the intuition and rationale behind using Eq. \ref{graph_attention} are shown in the \ref{Appendix_B}.}

By developing SPAM, we provide a calibration-free way to infer the inter-dependence of volume from speed data automatically. In STCAGCN, we compute SPAM for each input speed batch and thereby constructing dynamic graphs over time. And the activation function enables SPAM to achieve nonlinear mapping from speed correlations to volume correlations. Additionally, different from predefined graphs that are highly localized, SPAM connects globally related nodes adaptively, even they are remote in the sensor graph.

As SPAM is obtained from the speed data of all sensors, for batch forms of input, the produced SPAMs change over time with size $\mathcal{A}_{spa}\in\mathbb{R}^{B\times N\times N}$. Combining SPAM with TDCN, the final spatial feature extractor at $l$-th ($l\geq1$) layer can be formulated as:
\begin{equation}\label{spatial_conv}
\mathcal{H}^{l+1}=\operatorname{a}\left(\sum_{k=0}^K\widetilde{\mathcal{A}}_f^k\otimes\mathcal{H}^l\times_4\mathbf{W}_{kf}+\widetilde{\mathcal{A}}_b^k\otimes\mathcal{H}^l\times_4\mathbf{W}_{kb}+\mathcal{A}_{spa}^k\otimes\mathcal{H}^l\times_4\mathbf{W}_{spa}\right).
\end{equation}

In Eq. \eqref{spatial_conv}, TDCN performs forward and backward diffusion convolution to capture localized correlations, while SPAM conducts single graph convolution to extract global correlations. Through stochastic gradient descent, $\mathbf{A}_{spa}$ is trained to filter most relevant features on the whole graph, thereby enlarging the receptive field of GCN. In this way, we inject the speed information to STCAGCN adaptively and can capture multiresolution spatial dependencies in a unified module. 
% The input feature at $l$-th sptial convolution layer is a third-order tensor $\mathcal{H}\in\mathbb{R}^{N\times C_{in}\times L}$, where $N$ is the node numbers, $C_{in}$ is the hidden dimension, $L$ is the time length, and 
Graph convolution is performed on each time slot of $\mathcal{H}[:,t,:,:]$, which can be parallelized across time. 

By this design, STCAGCN not only can correlate locally adjacent sensors, but also globally relevant sensors with similar traffic patterns. When dealing with underdetermined flow locations where local conservation is less effective, global information provides a useful complementary. When dealing with determined flows, local information plays a dominating role and proper features are aggregated by TDCN and feed into the next layer. 
Distinct from predicting volumes directly from speeds, SPAM does not assume a well-developed speed-volume scatter, or fundamental diagram, which is a model-free routine.

\subsection{Capturing time-asynchronous correlations by gated temporal convolution and temporal masked self-attention}\label{tcn_sec}

Traffic volumes are significantly affected by the formation and dissipation of traffic congestion. If traffic flows operate in free-flow speeds and the distance between upstream and downstream sensors is close enough that all vehicles can pass through within the aggregating time window, or in the case traffic is operating in an equilibrium state, it is expected that the inflows and outflows are the same at any time intervals and the volume profiles of them should be identical. 
While when congestion occurs, vehicles could show different react times and some vehicles can not reach downstream sensor within the time window.
This nonequilibrium phenomenon is strongly associated with the distance, traffic density and speed. In this case, we claim that the volume of target sensor at current time interval has time-asynchronous correlations with the volume of neighborhood sensors at previous time steps. Most recently, \cite{yang2023traffic} also notice this phenomenon in traffic forecasting task and suppose traffic propagation could affect other road segments with a flow propagation delay that depends on traffic speed and network topology. As a result, flow dependencies of some roads may be inactive during the current time window due to traffic propagation delay.

\begin{figure}[!htb]
\centering
\centering
\includegraphics[scale=0.6]{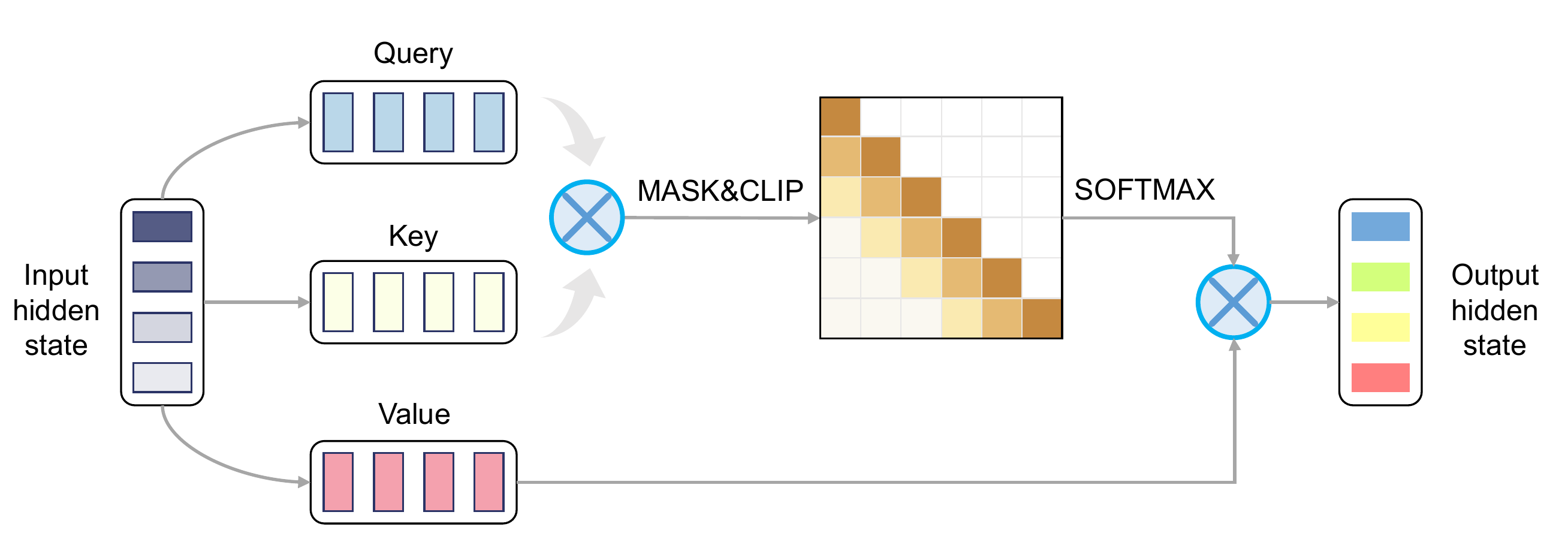}
\caption{Mechanism of temporal masked self-attention module. Each input temporal embedding is weighed by attention scores.}
\label{Tatt_arc}
\end{figure}

As discussed in \citep{yang2023traffic}, obtaining a reasonable time window to consider the impact of congestion delay is a non-trivial task. Besides, traffic state correlates with its previous states and such correlations vary with the traffic status at different times, e.g., during rush hours, the influence of congestion could exist a few hours. In order to capture the the propagation dynamics of traffic congestion and exploit time-asynchronous correlations adaptively to tackle the nonequilibrium phenomenon, we customize a masked and localized temporal self-attention module (Tatt), to dynamically obtain temporal embedding.
Attention mechanism has been successfully applied in sequence and time series data modeling \citep{zheng2020gman,hao2020temporal,wang2020forecast}. 
Given a temporal hidden state slice $\mathbf{Z}=\mathcal{H}[b,n,:,:]\in\mathbb{R}^{T\times C}$ produced by TDCN layer, we have:
\begin{equation}
    \mathbf{E}_{att}=\frac{\mathbf{Z}\mathbf{W}_t\cdot(\mathbf{Z}\mathbf{W}_t)^{\mathsf{T}}}{\sqrt{d_t}}
\end{equation}
where $\mathbf{E}_{att}$ is the unnormalized attention weight matrix, and a shared parameter $\mathbf{W}_t$ is used for projection of both key, value and query matrices. To avoid numerical overflow of $\operatorname{Softmax}$ function, we set the scale parameter as $\sqrt{d_t}=\Vert\mathbf{Z}\mathbf{W}_t\Vert_F^2$. 

After that, a time index masking matrix $\mathbf{M}_t$ is constructed to consider the casualty of time series, i.e., we take into account the impacts of previous time steps and avoid the future states affect the current state \citep{zheng2020gman}, which is given by:
\begin{equation}
    m_{ij} =\left\{\begin{array}{l}0,~\text{if } t_i\leq t_j, \\
    -\infty,~\text{otherwise}. \\    
    \end{array}\right. \\
\end{equation}
By adding $\mathbf{M}_t$, we can extract the lower triangular part of $\mathbf{E}_{att}$. Finally, the attention weighted hidden state is computed by applying a clip operation, normalization and weighted aggregation in sequence:
\begin{equation}\label{Tatt_form}
    \mathbf{Z}_{att} = \operatorname{SOFTMAX}\left(\operatorname{TOPk}(\mathbf{E}_{att}+\mathbf{M}_t)\right)\cdot(\mathbf{Z}\mathbf{W}_t),   
\end{equation}
in which $\operatorname{TOPk}(\cdot)$ is a attention clip operation that only the largest $k$ weights are kept, to ensure the model pay more attention on most relevant time slots and adapt for temporal features that have marginal influences on current time slot, e.g., free-flow intervals. Fig. \ref{Tatt_arc} illustrates the workflow of Tatt layer.
In vector form, Eq. \eqref{Tatt_form} can also be formulated as:
\begin{equation}
    \mathbf{z}_i=\sum_{j\in\mathcal{N}_{t_i}\cap\mathcal{N}_{k_i}}e_{ij}(\mathbf{W}_t\mathbf{z}_j),
\end{equation}
where $\mathcal{N}_{t_i}$ is the set consists of time points before $t_i$, and $\mathcal{N}_{k_i}$ denotes the set contains the largest $k$ values in $i$-th row of $\mathbf{E}_{att}$.
$\mathbf{z}_i$ is the aggregated representation of hidden state at time point $i$, considering potential influences of previous traffic states.

% Previous speed kriging works \citep{appleby2020kriging,wu2021inductive} have empirically demonstrated temporal dependence has minor effects on speed estimation accuracy and often neglect the propagation of traffic flow. However, case is quite different for traffic volume which is directly affected by the formation and dissipation of traffic congestion.

Furthermore, we also employ the gated temporal convolution network (TCN) after Tatt layer to ensure the consistent and shift-invariant traits of the estimaed traffic time series. 
Some recent works use dilated causal convolution \citep{wu2019graph,liang2022spatial}, recurrent neural network \citep{li2017diffusion} models to capture long-term temporal dependence for traffic forecasting. Consider the fact that we do not focus on modeling complex temporal dependencies in a long time range, because traffic congestion propagation between upstream and downstream detectors happens within a short time window. Therefore, TCN with localized receptive field is suitable for this purpose.

Besides, a gated linear unit (GLU) \citep{dauphin2017language} is used to screen features of previous time steps that can pass through TCN layer. For an input temporal embedding $\mathbf{Z}_{att}\in\mathbb{R}^{T\times C}$ produced by temporal attention layer, a gated 1-D temporal convolution with kernel size $k_t$ and residual connection is performed along the time dimension:
\begin{equation}\label{tcn_2d}
    \operatorname{GatedTCN}(\mathbf{Z})=(\bm{\theta}_1\star\mathbf{Z}+\mathbf{b}_1+\mathbf{Z})\odot\sigma(\bm{\theta}_2\star\mathbf{H}+\mathbf{b}_2)
\end{equation}

where $\star$ denotes the convolution operator, $\odot$ denotes the Hadmard product, $\sigma(\cdot)$ is the Sigmoid gate function which controls the ratio of temporal features flow to the following layer, $\bm{\theta}_1,\bm{\theta}_2$ and $\mathbf{b}_1,\mathbf{b}_2$ are learnable parameters. Note that we use a zero-padding technique to keep the input time length unchanged. For each node, TCN assumes data in each time point only correlates with previous time points within the length of convolution kernel $k_t$. Owing to this non-recursive computation way, Eq. \eqref{tcn} can also be parallelized across all nodes, which makes TCN more lightweight than recurrent neural network.

Both Eq. \eqref{Tatt_form} and \eqref{tcn_2d} can also be calculated in the form of tensors. Given the output of TDCN layer $\mathcal{H}\in\mathbb{R}^{B\times N\times T \times C}$, we first permute it to $\widetilde{\mathcal{H}}\in\mathbb{R}^{B\times T\times N \times C}$ and the following equations process the temporal features $\mathcal{T}$:
\begin{equation}
\mathcal{E}_{att} = \frac{\operatorname{UNFOLD}(\widetilde{\mathcal{H}}\times_4\mathbf{W}_t)\otimes\operatorname{UNFOLD}(\widetilde{\mathcal{H}}\times_4\mathbf{W}_t)^{\mathsf{T}}}{\sqrt{d_t}},
\end{equation}

\begin{equation}
    \widetilde{\mathcal{H}}_{att} = \operatorname{SOFTMAX}\left(\operatorname{TOPk}(\mathcal{E}_{att}+\mathcal{M}_t)\right)\cdot(\widetilde{\mathcal{H}}\times_4\mathbf{W}_t),   
\end{equation}

\begin{equation}\label{tcn}
    \mathcal{T}=(\Theta_1\star\widetilde{\mathcal{H}}_{att}+\mathcal{B}_1)\odot\sigma(\Theta_2\star\widetilde{\mathcal{H}}_{att}+\mathcal{B}_2)\in\mathbb{R}^{\widetilde{\mathcal{H}}\in\mathbb{R}^{B\times T\times N \times C_{out}}},
\end{equation}
where $\operatorname{UNFOLD}(\cdot): \mathbb{R}^{B\times T\times N \times C}\rightarrow\mathbb{R}^{B\times T\times NC}$ is a linear transform to change the order of a tensor.

In STCAGCN, we apply a single Tatt-TCN layer after the first TDCN layer without self-loops, so that the aggregated spatial features of neighborhood nodes are feed into Tatt-TCN layer and it extracts temporal features from most correlated previous time points for each node. Then the obtained temporal embedding containing the information of neighboring nodes at previous time slots are feed into next TDCN layer with self-loops. By this means, each sensor could have access to time-asynchronous observations with neighbor nodes in an adaptive manner. 

\subsection{Model implementations}
\subsubsection{Output layer with feature concatenation and residual connection}
To alleviate the over-smoothing problem of stacked GNNs \citep{chen2020measuring} and accelerate the back propagation of gradient, we add a residual connection \citep{he2016deep} at the end of each TDCN block, i.e.,
\begin{equation}
    \mathcal{H}^{l+1}=\mathcal{H}^{l}+\operatorname{TDCN}(\mathcal{H}^{l}).
\end{equation}

Besides, to fully utilize the  multi-scale spatiotemporal features produced by TDCN and Tatt-TCN, we further employ a feature concatenation layer after the last TDCN layer to integrate these features. Finally, a linear fusion layer is used to produce the output:
\begin{equation}
    \widehat{\mathcal{X}}_v=\operatorname{CONCAT}(\mathcal{H}^0,\mathcal{H}^1,\dots,\mathcal{H}^m)\times_4\mathbf{W}_o+\mathcal{B}_o,
\end{equation}
where $\mathbf{W}_o\in\mathbb{R}^{mC_{out}\times C_{out}}$ and $\mathcal{B}_o\in\mathbb{R}^{C_{out}}$ are parameters of the fusion layer, $m$ is the number of TDCN layers, and $\operatorname{CONCAT}(\cdot)$ denotes the data concatenation operation along the last dimension.

\subsubsection{Training strategy and loss function}\label{train}

As our volume estimation problem can be viewed as a special kriging problem, the objective is to reconstruct the manually masked sub-graphs and generalize to unknown sensor locations during testing. Therefore, the reconstruction loss is computed on the whole input volume data, including both observed and masked nodes, to benefit the generalization ability :
\begin{equation}
    \mathscr{L}_r = \Vert\mathcal{X}_v-\widehat{\mathcal{X}}_v\Vert_1,
\end{equation}
where $\Vert\cdot\Vert_1$ is the $\mathcal{L}_1$ norm.

We observe that the adjacency matrix in reality is sparse and a small number of connected edges play a dominant role, i.e., a node's volume value is related with a fraction of other nodes'. And the node values usually vary smoothly across adjacent nodes. To control the sparsity and smoothness of the learned graph, we apply a graph Laplacian (a.k.a., graph Dirichlet energy \citep{belkin2001laplacian}) regularization on the training graph using SPAM $\mathbf{A}_{spa}$.

Specifically, given the volume signals $\mathcal{X}_v$, the graph Laplacian is computed on all batch inputs as follows:
\begin{equation}\label{laplacian}
    \mathscr{L}_g=\mathscr{D}(\mathbf{A}_{spa},\mathcal{X}_v) = \frac{1}{BNT}\sum_b\sum_{j, j'}a_{jj'}^{spa}\Vert \mathbf{x_s}_j-\mathbf{x_s}_{j'}\Vert_2^2=\sum_b\operatorname{Tr}(\mathcal{X}_v[b,:,:,:]\mathbf{L}\mathcal{X}_v[b,:,:,:]^\mathsf{T}),
\end{equation}
where $\mathbf{L}=\operatorname{Diag}(\sum_{j}a_{ij}^{spa})-\mathbf{A}_{spa}$ is the graph Laplacian matrix, $\operatorname{Tr}(\cdot)$ computes the trace of a matrix, and $\Vert\cdot\Vert_2$ represents the Frobenius norm. It is observed that minimizing $\mathscr{D}(\mathbf{A}_{spa},\mathcal{X}_v)$ encourages connected nodes of the learned graph to share similar embedding, leading to a smooth and sparse graph.

Then, the total loss function controlled by a parameter $\lambda$ is:
\begin{equation}\label{loss}
    \mathscr{L} = \mathscr{L}_r + \lambda\mathscr{L}_g.
\end{equation}

Following \citep{wu2021inductive}, the training strategy is in fact a random sub-graphs masking and reconstructing process (as discussed in section \ref{estimation_kriging}). The input is a series of manually masked graph signals, and output is the complete graph signals. Locations in $\mathscr{V}^u$ are not exposed in the training stage and only used for testing.
By this setting, STCAGCN is trained to reconstruct incomplete graphs and generalize to unseen nodes (not in the constructed graph during training).
We provide the pseudo code of model training in Algorithm \ref{training}. Owing to this inductive training manner, the proposed model has the ability to perform real-time volume kriging given instant measurements, which is a clear advantage over matrix/tensor based methods \citep{lei2022bayesian,nie2022truncated}.

\begin{algorithm}[!htb]
\caption{Training sample generation and model update for STCAGCN training}\label{training}
\KwIn{Training volume data from $n_o$ observed locations during $T$ time steps $\mathbf{X}_v\in\mathbb{R}^{n_o\times T}$, speed data from $n_o$ observed locations during the same period $\mathbf{X}_s\in\mathbb{R}^{n_o\times T}$, adjacency matrix $\mathbf{A}_f, \mathbf{A}_b$, sequence length $L$, hidden dimension $D$, kernel size $k_t$, diffusion steps $K$, batch size $B$, number of batches $N_b$, number of masked nodes $n_m$, maximum iterations $I$.}
\KwOut{Trained model $\mathscr{F}^I$ and learned graph $\mathbf{A}_{spa}$.}
initialize the model as $\mathscr{F}^0 = \operatorname{STCAGCN}(L,D,k_t,K)$;  \\ 
\For{$i=1:I$}{
\For{$n=1:N_b$}{
\tcp{Split sample batches along time dimension}
randomly generate $B$ individual  time points $T_r = [t_{r1},t_{r2},\dots,t_{rB}]$ within $[0,T-h]$ to split samples for each batch; \\
\tcp{Generate input data tensors}
initialize $\mathcal{X}_v\in\mathbb{R}^{B\times n_o\times L},\mathcal{X}_s\in\mathbb{R}^{B\times n_o\times L}$ and $\mathcal{M}\in\mathbb{R}^{B\times n_o\times L}$ as zeros;\\
\For{$b=1:B$}{prepare training volume data with sequence length $L$: $\mathcal{X}_v[b,:,:]\leftarrow\mathbf{X}_s[:,T_r[b]:T_r[b]+L]$;\\
prepare training speed data with sequence length $L$: $\mathcal{X}_s[b,:,:]\leftarrow\mathbf{X}_v[:,T_r[b]:T_r[b]+L]$;\\
randomly sample $n_m$ masking index $M_r\leftarrow[m_{n_1},m_{n_2},\dots,m_{n_m}]$ within $[0,n_o]$ and set $\mathcal{M}[b,M_r,:]\leftarrow0$
}
construct masked input volume tensor as: $\mathcal{X}_{\operatorname{input}}\leftarrow\mathcal{M}*\mathcal{X}_v$;\\
\tcp{Model training via stochastic gradient descent}
forward computation: $\widehat{\mathcal{X}},\mathbf{A}_{spa} \leftarrow \operatorname{STCAGCN}(\mathcal{X}_{\operatorname{input}},\mathcal{X}_s,\mathbf{A}_f, \mathbf{A}_b)$;\\
compute loss function by Eq. \eqref{loss};\\
loss backward and model parameters updating; \\
}}
\end{algorithm}

\section{Case study}\label{experiments}
In this section, we evaluate the proposed STCAGCN model on a publicly available highway traffic volume dataset and provide comprehensive result analysis and discussions. All experiments are carried out on a Windows 10 platform with Intel Core i7-12700KF CPU and an NVIDIA GeForce RTX 3080 GPU. \textbf{The trained models and PyTorch implementations are publicly available in our GitHub repository:} \url{https://github.com/tongnie/GNN4Flow}.

\subsection{Data preparation and experiment setup}
\subsubsection{Highway traffic volume dataset}
To verify our model, we conduct case studies using 
PeMS-Bay traffic volume data collected from the Performance Measurement System of California \footnote{from \url{https://pems.dot.ca.gov/}}. 
We collect this traffic volume data from 325 loop sensors in the San Francisco bay area, ranging from January to March 2018 with a 5-min time interval. Note that due to failures of some volume sensors, original PeMS data suffers from a proportion of missing values and data imputation is already conducted for users. Data in the selected time span has a relatively high observation rate (about 80\%) that is desired to perform experiments in this paper. The speed data within the same period, network topology and road background information (including lane number and driving direction) are also prepared.

\subsubsection{Analysis of traffic flow phenomenon of PeMS data}\label{data_explore}
\begin{figure}[!htb]
\centering
\subfigure[Statistical results of WDSSI]{
\centering
\includegraphics[scale=0.52]{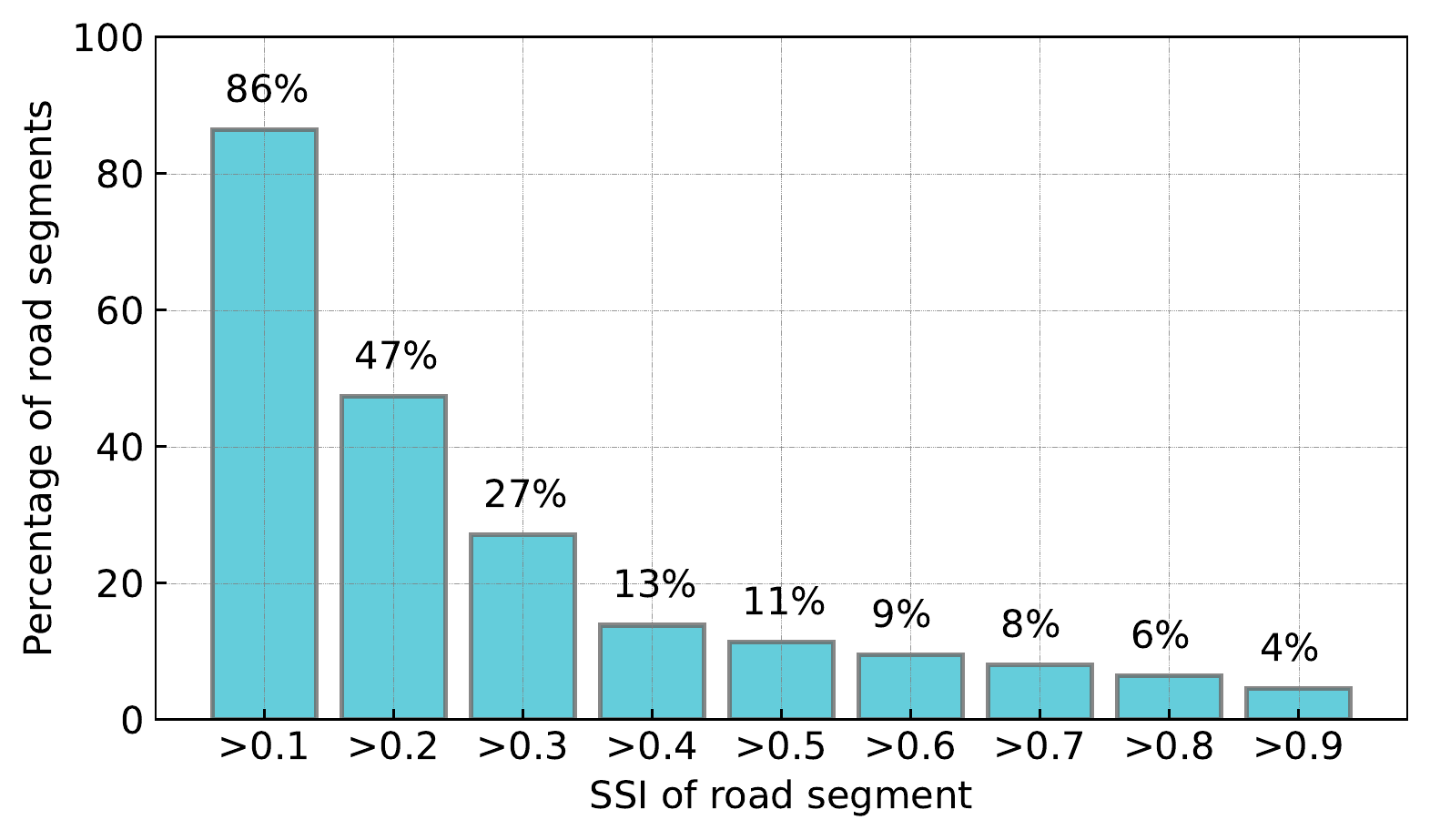}
}
\subfigure[Statistical results of TAI]{
\centering
\includegraphics[scale=0.52]{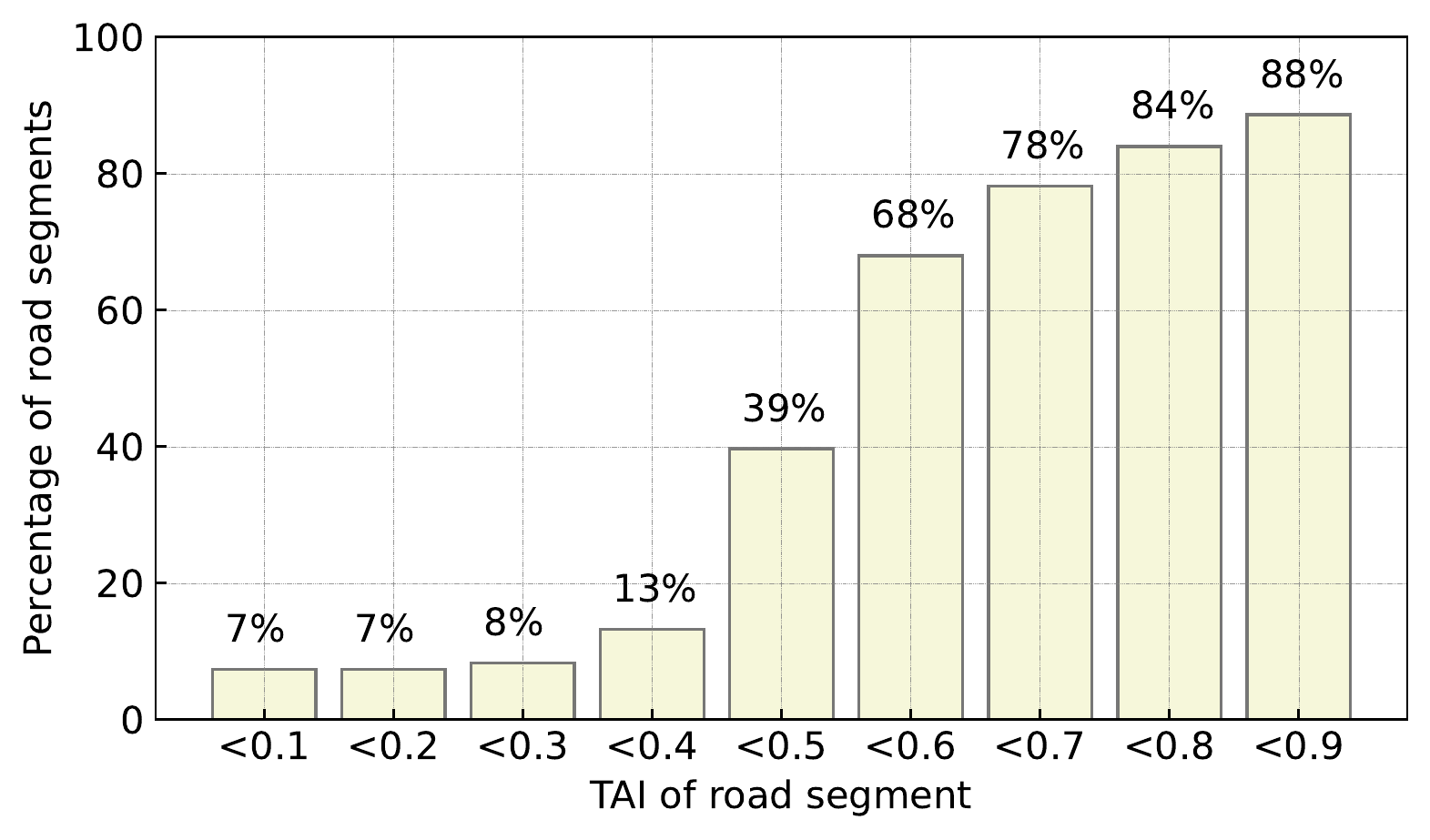}
}
\caption{Histogram of WDSSI and TAI of sensors in PeMS-Bay data set.}
\label{PeMS_SSI_TAI}
\end{figure}

Recall that we interpret the challenges of volume estimation as underdetermined and nonequilibrium problems, which are caused by undetected movements and traffic congestion delay respectively. To reveal these phenomena on PeMS-Bay dataset, we first calculate the WDSSI using Eqs. \eqref{SSI}. The percentage of road segments/sensors with different WDSSI values are shown in Fig. \ref{PeMS_SSI_TAI} (a).

As can be seen, about $30\%$ of sensors have WDSSI larger than $0.3$, which could cause high estimation error due to the unmeasured drive-in and drive-out between sensors \citep{zhang2020network}. We further compute the TAI of those road segments with WDSSI smaller than $0.3$ (can be approximated as closed road segments) using Eq. \eqref{TAI} to examine the influence of traffic propagation. Fig. \ref{PeMS_SSI_TAI} (b) reports that the TAI of nearly $40\%$ of sensors is less than $0.5$, and we can regard them as the results of being affected by nonequilibrium traffic flows.

We will show that these small proportions of sensors could account for the majority of estimation errors. In order to demonstrate the effectiveness of our model design, these two metrics are used to distinguish between two types of estimation errors in the following experiments.

\subsubsection{Baseline models and experiment setup}
We compare our STCAGCN model with several baseline models, including the benchmark KNN method, several state-of-the-art deep learning and matrix/tensor models, to justify our model designs. Brief descriptions of each method are displayed as follows.

\begin{itemize}
\item \textbf{KNN}: K-nearest neighborhoods method is the benchmark for spatial kriging and it estimates unobserved volumes by simply averaging neighborhoods' values. By comparing with KNN, we could show the significance of modeling time-asynchronous and non-local correlations. 
\item \textbf{ST-SSL \citep{meng2017city}}: Graph based semi-supervised learning model for city-wide traffic volume estimation. This is a pioneering work for estimating city-wide traffic volume using speed information as supplements, which formulates a semi-supervised learning model.
\item \textbf{TGMF-S \citep{zhang2020network}}: Temporal geometric matrix factorization with speed Pearson coefficients. This model uses graph Laplacian and one-order Toeplitz matrix as spatial and temporal regularization. Speed information is used to calculate Pearson coefficients and construct a static adjacency matrix.
\item \textbf{IGNNK \citep{wu2021inductive}}: Inductive graph neural networks for traffic speed kriging. By replacing the adjacency matrix with a speed-based adjacency matrix, this model can also perform volume kriging.
\item \textbf{STAR \citep{liang2022spatial}}: This model modifies IGNNK by adding dilated causal convolution layers for temporal features and external attention modules for spatial features.
\item \textbf{LETC \citep{nie2023correlating}}: Spatiotemporal Laplacian-enhanced low-rank tensor completion model. Both spatiotemporal correlations and low-rankness are modeled in a unified tensor nuclear norm minimization framework.
\end{itemize}

Main hyper-parameters in STCAGCN are specified as follows. For spatial module, the diffusion step is set to 1, hidden dimension is set to 128, and the number of TDCN layers is 5. For temporal module, we set the kernel size as 3, the sequence length as 24, and the hidden dimension is also 128. In the training stage, the input batch size is 32, and we choose the Adam optimizer with a fixed learning rate being 5e-4. In the loss function, $\lambda$ is set to 1e-4. We train STCAGCN for 300 epochs with early stopping to avoid over-fitting. To give a fair comparison, the predefined adjacent matrices in IGNNK, STAR and LETC are calculated by speed Pearson coefficients, which are the same as in TGMF-S.

We first compare our model with baselines under $50\%$ sensor coverage rate, i.e., 160 sensors are randomly selected as observed locations used for model training, and the remaining 165 sensors are used only for evaluation during testing. In this scenario, we set the randomly masked nodes in Algorithm \ref{training} as 80, which equals to $50\%$ of the observed nodes. This consistency between training masked rate and testing coverage rate is essential to lean an unbiased model (discussed in subsection \ref{sec_train_test_rate}). 

Three quantitative metrics including Mean Absolute Error (MAE), Root Mean Square Error (RMSE), Mean Absolute Percentage Error (MAPE), and Weighted Mean Absolute Percentage Error (WMAPE) in Eq. \eqref{metric} are used to evaluate the estimation results. As the range of traffic volume values varies widely, WMAPE is applied to alleviate the influence of extremes. 
% GEH is a commonly used indicator to measure the goodness of a traffic flow model, we use herein to demonstrate the actual applicability of different models.

% \begin{equation}\label{metric}
%     \begin{aligned}
%     &\text{MAE}= \frac{1}{NT}\sum_{i=1}^N\sum_{j=1}^T |x_{ij}-\hat{x}_{ij}|,\\
%     &\text{RMSE}= \sqrt{\frac{1}{NT}\sum_{i=1}^N\sum_{j=1}^T \left(x_{ij}-\hat{x}_{ij}\right)^2},\\
%     &\text{MAPE}= \frac{1}{NT}\sum_{i=1}^N \sum_{j=1}^T\vert\frac{x_{ij}-\hat{x}_{ij}}{x_{ij}}\vert,\\
%     &\text{WMAPE}= \frac{\sum_{i=1}^N \sum_{j=1}^T\vert x_{ij}-\hat{x}_{ij}\vert}{\sum_{i=1}^N \sum_{j=1}^Tx_{ij}},\\
%     &\text{GEH}_{ij}=\sqrt{\frac{2*(x_{ij}-\hat{x}_{ij})^2}{x_{ij}+\hat{x}_{ij}}},
%     \end{aligned}
% \end{equation}
% where $\text{GEH}_{ij}$ computes a point value, we use the average GEH over all time slots and locations as an indicator. 
\begin{equation}\label{metric}
    \begin{aligned}
    &\text{MAE}= \frac{1}{NT}\sum_{i=1}^N\sum_{j=1}^T |x_{ij}-\hat{x}_{ij}|,~
    \text{RMSE}= \sqrt{\frac{1}{NT}\sum_{i=1}^N\sum_{j=1}^T \left(x_{ij}-\hat{x}_{ij}\right)^2},\\
    &\text{MAPE}= \frac{1}{NT}\sum_{i=1}^N \sum_{j=1}^T\vert\frac{x_{ij}-\hat{x}_{ij}}{x_{ij}}\vert,~
    \text{WMAPE}= \frac{\sum_{i=1}^N \sum_{j=1}^T\vert x_{ij}-\hat{x}_{ij}\vert}{\sum_{i=1}^N \sum_{j=1}^Tx_{ij}}.
    \end{aligned}
\end{equation}

\subsection{Volume estimation results and model comparison}\label{results_on_50_missing}
\subsubsection{Estimation results}
\begin{figure}[!htb]
\centering
\subfigure[\minew{Sensors with underdetermined and nonequilibrium flows (only data in test are displayed).}]{
\centering
\includegraphics[scale=0.5]{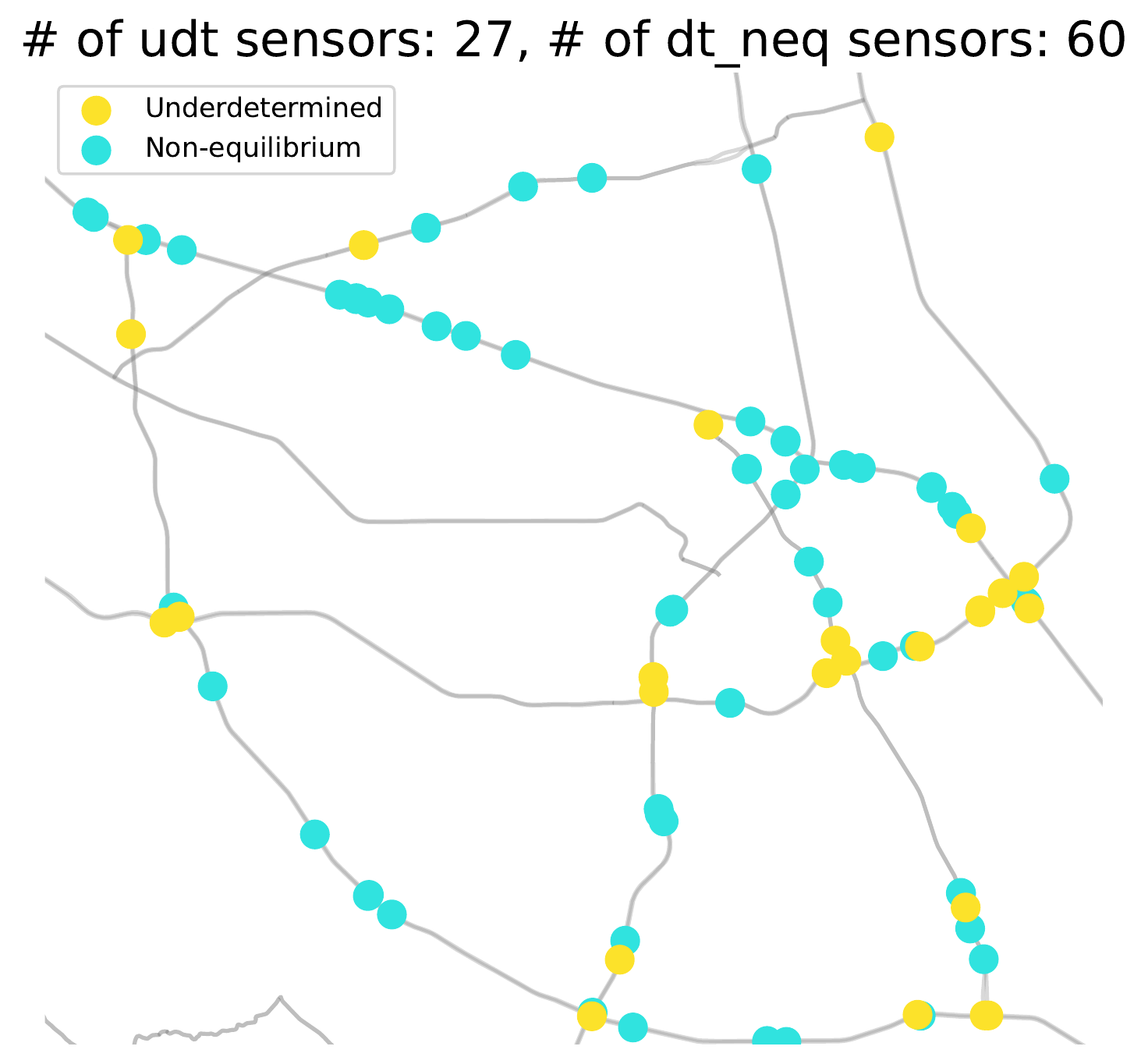}
}
\subfigure[Traffic speeds measured by loop sensors of PeMS-Bay area.]{
\centering
\includegraphics[scale=0.51]{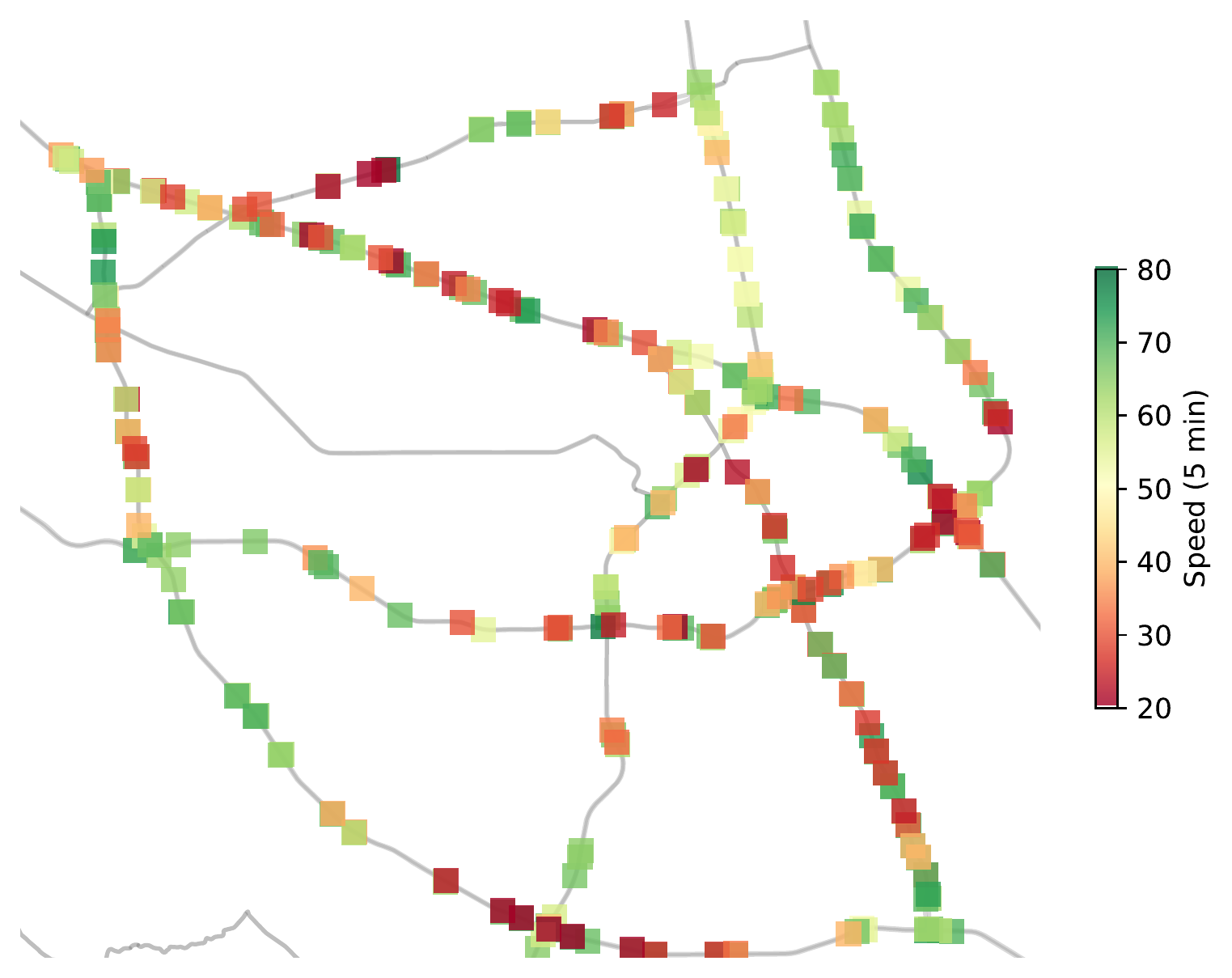}
}
\caption{Locations of two kinds of sensors and traffic states on PeMS highway network.}
\label{two_types_sensors}
\end{figure}

Tab. \ref{PeMS_50_results} reports the volume estimation performances of STCAGCN and other baselines under $50\%$ spatial coverage rate. As described in subsection \ref{data_explore}, we first split these results into sensor groups caused by different reasons. Specifically, three types of road segments can be categorized: 
1) \textit{MAPE\_udt/WMAPE\_udt}: errors of sensors with WDSSI higher than 0.4, which can be viewed as underdetermined parts and account for about $17\%$ of the test sets; 
2) \textit{MAPE\_dt\_eq/WMAPE\_dt\_eq}: errors of sensors with WDSSI lower than 0.4 but TAI larger than 0.5, which are those stationary segments with minor influence of propagation delay and account for about $46\%$ of the test sets; 
3) \textit{MAPE\_dt\_neq/WMAPE\_dt\_neq}: errors of sensors with lower WDSSI and smaller TAI, which are space-stationary but time-asynchronous parts and account for about $37\%$ of the test sets. Their respective positions are shown in Fig. \ref{two_types_sensors}. 

\begin{table}[!htb]
  \centering
  \caption{Volume estimation results under $50\%$ coverage rate of different methods.}
    \begin{tabular}{l|>{\columncolor{orange!20!white}}ccccccc}
    \toprule
          & \multicolumn{1}{l}{STCAGCN} & \multicolumn{1}{l}{KNN} & \multicolumn{1}{l}{TGMF-S} & \multicolumn{1}{l}{ST-SSL} & \multicolumn{1}{l}{IGNNK} & \multicolumn{1}{l}{STAR} & \multicolumn{1}{l}{LETC} \\
    \midrule
    MAE   & \textbf{43.83} & 53.83 & 49.28 & 52.78 & 49.43 & 50.09 & 51.88 \\
    RMSE  & \textbf{65.99} & 77.73 & 74.28 & 82.10 & 72.77 & 74.89 & 77.51 \\
    MAPE  & \textbf{30.13\%} & 46.51\% & 39.05\% & 40.69\% & 38.25\% & 38.59\% & 44.41\% \\
    WMAPE & \textbf{19.59\%} & 24.17\% & 22.12\% & 23.71\% & 22.09\% & 22.38\% & 23.38\% \\
    \midrule
    MAPE\_udt & \textbf{58.85\%} & 113.98\% & 99.47\% & 95.96\% & 72.15\% & 71.20\% & 101.56\% \\
    WMAPE\_udt & \textbf{30.90\%} & 56.67\% & 48.55\% & 51.65\% & 34.56\% & 36.17\% & 50.66\% \\
    \midrule
    MAPE\_dt\_eq & \textbf{22.41\%} & 27.76\% & 23.89\% & 25.95\% & 24.78\% & 25.97\% & 24.36\% \\
    WMAPE\_dt\_eq & \textbf{16.37\%} & 18.90\% & 17.02\% & 17.76\% & 18.04\% & 18.35\% & 17.53\% \\
    \midrule
    MAPE\_dt\_neq & \textbf{26.74\%} & 39.27\% & 31.79\% & 33.13\% & 39.58\% & 39.46\% & 42.98\% \\
    WMAPE\_dt\_neq & \textbf{19.80\%} & 22.23\% & 20.81\% & 25.07\% & 22.97\% & 22.79\% & 29.83\% \\
    \bottomrule
    \multicolumn{5}{l}{\scriptsize{Best results are bold marked.}}
    \end{tabular}%
  \label{PeMS_50_results}%
\end{table}%

Observing the distributions of two types of sensors and speeds, an interesting finding is that underdetermined flows generally occur at the interchange of freeways, where vehicles frequently access to other expressways via on/off ramps. Besides, as expected, nonequilibrium phenomenon widely exists in congested road segments. These observations justify the use of two metric in section \ref{sec_SSI_TAI} to quantify the two special challenges in this work.

By comparing three types of errors longitudinally, we can see that type 1 error is the largest, followed by type 3 error. This result clearly verify our assumption that a small number of underdetermined or nonequilibrium road segments dominate most of the estimation errors, thereby confirming the necessity of customizing models targeting these two phenomena. As enough accurate information can be borrowed from neighborhoods directly, it is natural that type 2 error is lowest.

By comparing different models horizontally, several noteworthy observations are stated as follows:
\begin{enumerate}[(1)]
    \item STCAGCN outperforms all baselines consistently with significant improvements. Compared with the second best model IGNNK, our model is improved by more than 8 percentage points with respect to MAPE.
    \item Due to the design of SPAM, STCAGCN reduces MAPE\_udt/WMAPE\_udt dramatically. With the adaptive injection of speed information, STCAGCN can borrow information from globally related sensors on the whole network to fix the spatial underdetermined problem.
    \item All models achieve desired results on MAPE\_dt\_eq/WMAPE\_dt\_eq because the volumes of this portion of sensors can be effectively recovered by neighborhood values. Notice that STCAGCN can also reduce this part of errors since the TDCN module can filter the most appropriate neighbor sensors according to the directed nature of traffic flow.
    \item By considering time-asynchronous correlations caused by propagation delay, STCAGCN can increase accuracy of nonequilibrium sensors significantly. 
\end{enumerate}

To conclude, our model achieves state-of-the-art in two aspects: STCAGCN performs better than other baselines on the simplest tasks, while at the same time can achieve satisfactory performances on tricky tasks that other models are less effective.

\begin{figure}[!htb]
\centering
\includegraphics[scale=0.6]{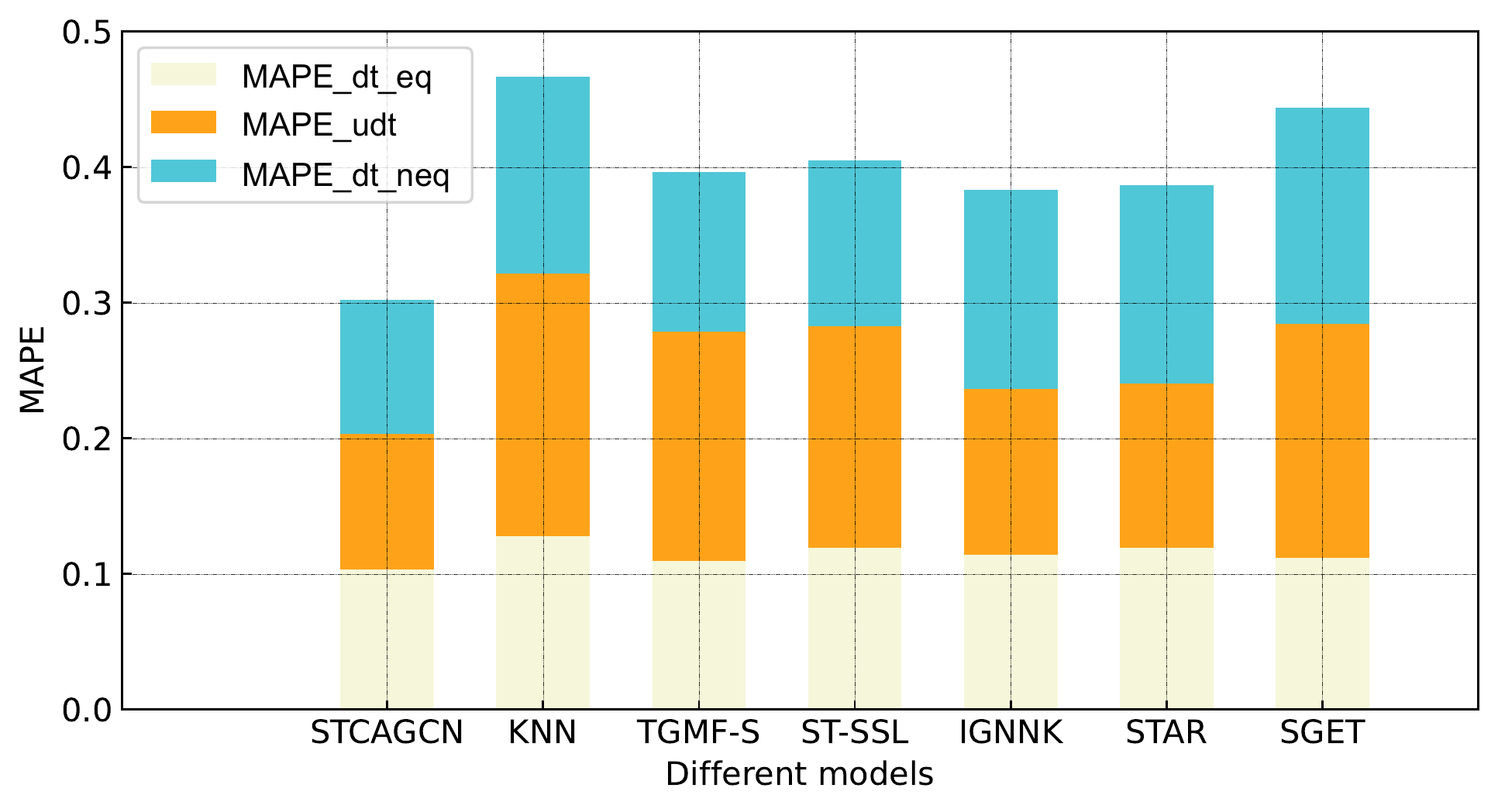}
\caption{Proportion of different types of errors for each model.}
\label{stacked_bar}
\end{figure}

To display the proportion of each type of error, the stacked bar chart is given in Fig \ref{stacked_bar}. Each error value is weighted by their percentage, i.e., $17\%$, $46\%$ and $37\%$ for MAPE\_udt, MAPE\_dt\_eq, and MAPE\_dt\_neq respectively, and the sum of them equals to the overall MAPE in Tab. \ref{PeMS_50_results}. From this figure, we can see that underdetermined and nonequilibrium errors account for more than half of the MAPE errors, which emphasizes the necessity of optimizing these parts. Moreover, it can be intuitively observed that STCAGCN reduces both underdetermined and nonequilibrium errors significantly, compared with other baselines.

\begin{figure}[!htb]
\centering
\subfigure[Ground truth values]{
\centering
\includegraphics[scale=0.5]{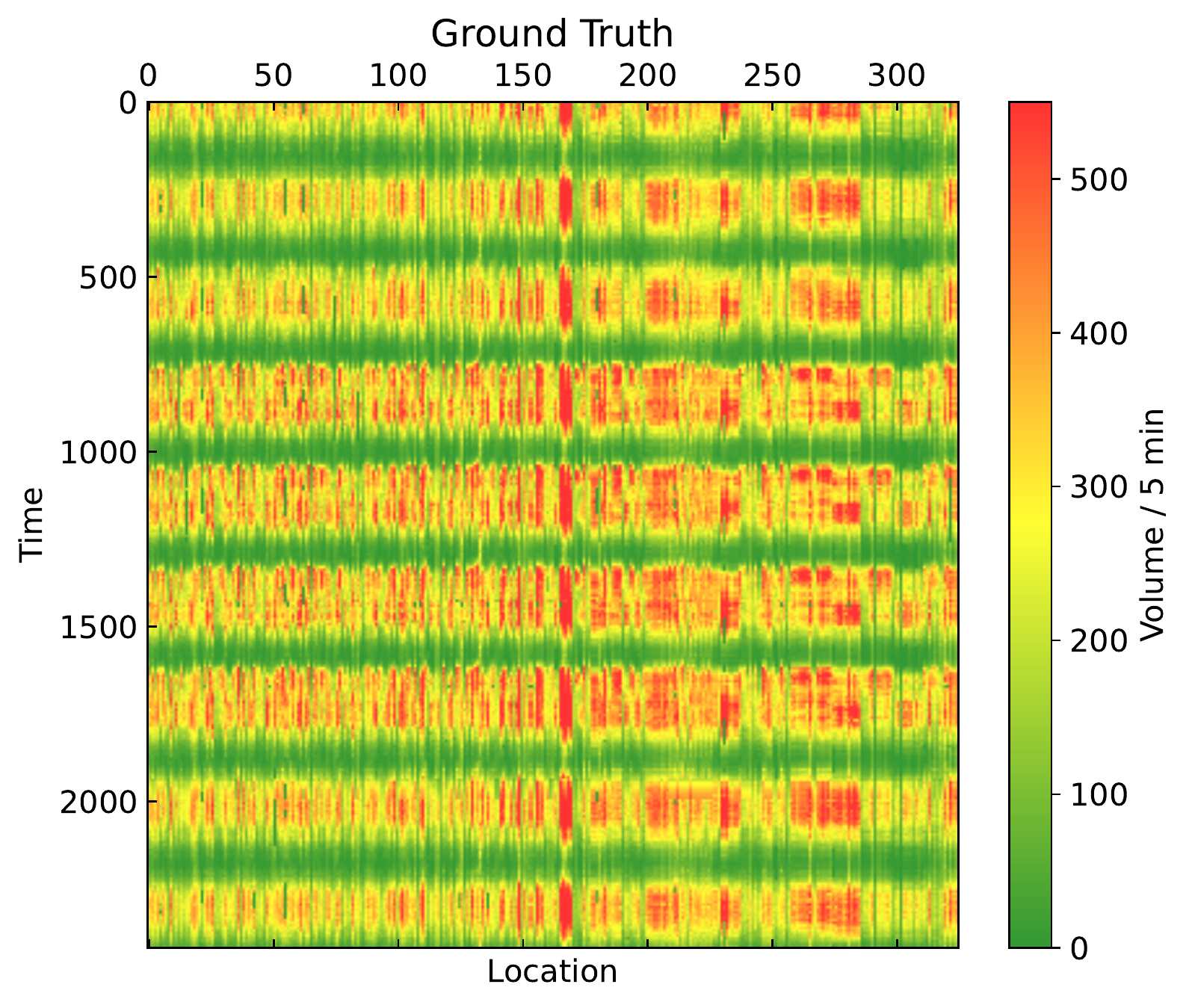}
}
\subfigure[STCAGCN estimated values]{
\centering
\includegraphics[scale=0.5]{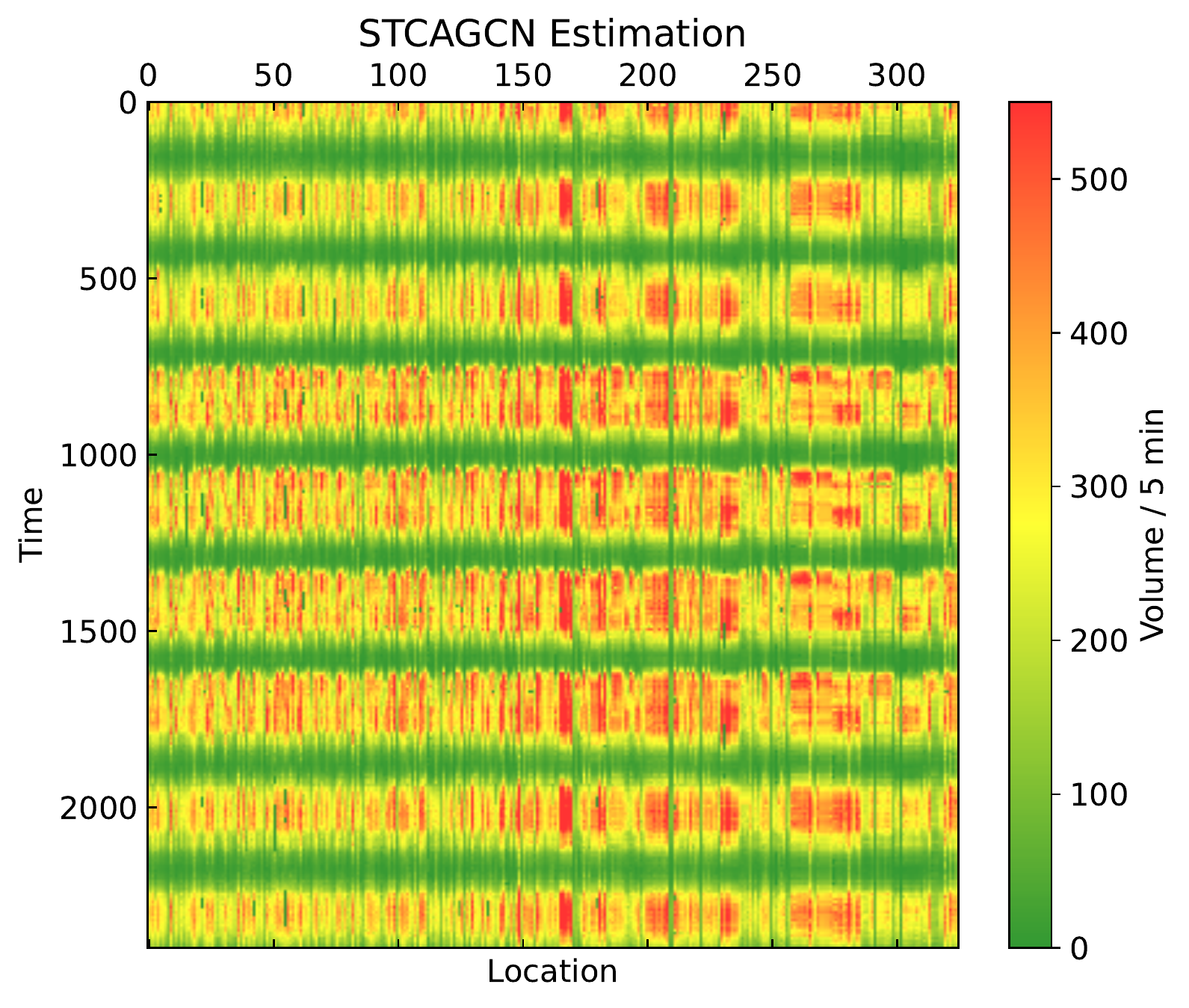}
}
\caption{Ground truth traffic volumes versus STCAGCN estimated traffic volumes of PeMS data.}
\label{volume_GT_HAT}
\end{figure}

We further visualize the ground truth volume values and STCAGCN estimations using the testing data in Fig. \ref{volume_GT_HAT}. As we can see, traffic volumes in this district are relatively high and many road segments have volume values higher than 400 within 5 min. Therefore, traffic congestion could have obvious implications for the propagation of traffic flow. STCAGCN can produce traffic volume estimations quite close to the true values even $50\%$ of the observations are unavailable, which intuitively confirms the effectiveness of our model.

\subsubsection{Model performances under lower coverage rates}
\begin{figure}[!htb]
\centering
\includegraphics[scale=0.45]{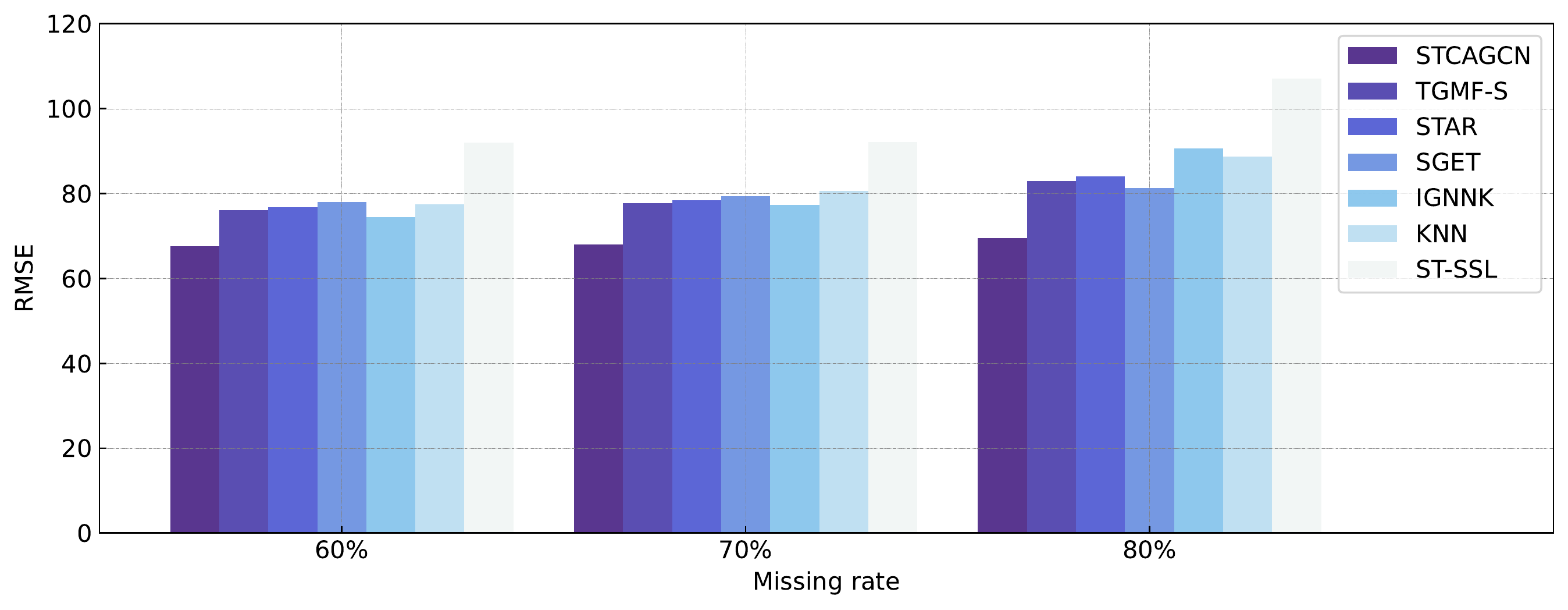}
\caption{Model performances under lower coverage rate.}
\label{rmse_compare}
\end{figure}

In order to examine the model effectiveness under lower sensor coverage rate, we set the missing rate to range from $60\%,70\%$ to $80\%$, to compare different models. Tab. \ref{PeMS_60_80_results} gives the evaluation metrics in total, and the RMSE results are also shown in Fig. \ref{rmse_compare}. \minew{We also provide the spatial distributions of both observed and unobserved sensors (including the locations of two special types of sensors) under various missing rates on the PeMS network in \ref{Appendix_A}.}

\begin{table}[!htb]
  \centering
  \caption{Model comparison under lower coverage rates.}
    \begin{tabular}{c|l|>{\columncolor{orange!20!white}}ccccccc}
    \toprule
    \multicolumn{1}{r}{} &       & \multicolumn{1}{l}{STCAGCN} & \multicolumn{1}{l}{KNN} & \multicolumn{1}{l}{TGMF-S} & \multicolumn{1}{l}{ST-SSL} & \multicolumn{1}{l}{IGNNK} & \multicolumn{1}{l}{STAR} & \multicolumn{1}{l}{LETC} \\
    \midrule
    \multirow{4}[2]{*}{60\%} & MAE   & \textbf{45.99} & 52.22 & 50.61 & 57.72 & 52.66 & 51.94 & 52.56 \\
          & RMSE  & \textbf{67.57} & 77.44 & 76.12 & 92.03 & 74.46 & 76.78 & 77.94 \\
          & MAPE  & \textbf{32.83\%} & 43.95\% & 39.00\% & 39.08\% & 48.47\% & 38.23\% & 45.14\% \\
          & WMAPE & \textbf{20.50\%} & 23.47\% & 22.74\% & 25.95\% & 23.55\% & 23.23\% & 23.70\% \\
    \midrule
    \multirow{4}[2]{*}{70\%} & MAE   & \textbf{46.37} & 55.24 & 51.79 & 58.51 & 55.65 & 53.71 & 54.05 \\
          & RMSE  & \textbf{68.02} & 80.58 & 77.80 & 92.12 & 77.37 & 78.37 & 79.40 \\
          & MAPE  & \textbf{31.18\%} & 44.85\% & 38.60\% & 38.16\% & 50.39\% & 47.62\% & 46.73\% \\
          & WMAPE & \textbf{20.74\%} & 24.74\% & 23.20\% & 26.22\% & 24.81\% & 23.94\% & 24.29\% \\
    \midrule
    \multirow{4}[2]{*}{80\%} & MAE   & \textbf{48.51} & 63.91 & 56.02 & 70.92 & 66.64 & 59.23 & 56.77 \\
          & RMSE  & \textbf{69.55} & 88.67 & 83.02 & 107.07 & 90.63 & 84.09 & 81.26 \\
          & MAPE  & \textbf{36.38\%} & 54.12\% & 40.60\% & 44.85\% & 72.16\% & 65.42\% & 55.82\% \\
          & WMAPE & \textbf{21.61\%} & 28.61\% & 25.08\% & 31.76\% & 29.69\% & 26.39\% & 25.50\% \\
    \bottomrule
    \multicolumn{5}{l}{\scriptsize{Best results are bold marked.}}
    \end{tabular}%
  \label{PeMS_60_80_results}%
\end{table}%

With lower coverage rate, less redundancy of the sensor network can be used so that the estimation performances of all models degrade. In spite of this, STCAGCN is more robust than competing models with minimal increase in errors. Under $80\%$ missing rate ($20\%$ coverage rate), the proposed model is the only one with a MAPE error of less than $40\%$. These findings further confirm the superiority of the proposed model.

\subsection{Model analysis}
\subsubsection{Ablation studies}
To inspect the significance of each component of the proposed model, in this part we conduct ablation studies to compare the performances of model variations under $50\%$ coverage rate. Totally, we examine the following three variations of the original STCAGCN model by removing or replacing specific modules:

\begin{itemize}
    \item \textbf{STCAGCN w/o SPAM}: We remove the speed pattern adaptive adjacency matrix construction module in STCAGCN. A static speed Pearson coefficients matrix is used instead.
    \item \textbf{STCAGCN w/o Tatt-TCN}: STCAGCN without temporal attention and convolution layer. In this case, input time series are treated as individual samples and only spatial features are extracted.
    \item \textbf{STCAGCN w/o TDCN}: We replace the direction-specific diffusion convolution with a symmetric graph spectral convolution operator in \citep{yu2017spatio}.
    \item \minew{\textbf{STCAGCN w/o PredA}: We remove the predefined adjacency matrices $\mathbf{A}_f, \mathbf{A}_b$ and only rely on the SPAM to perform volume kriging.}
\end{itemize}

\begin{table}[!htb]
  \centering
  \caption{\minew{Ablation studies.}}
    \begin{tabular}{l|>{\columncolor{orange!20!white}}ccccc}
    \toprule
          & \multicolumn{1}{l}{STCAGCN}& \multicolumn{1}{l}{w/o Tatt-TCN} & \multicolumn{1}{l}{w/o SPAM} & \multicolumn{1}{l}{w/o TDCN} & \multicolumn{1}{l}{w/o PredA}\\
    \midrule
    MAE   & \textbf{43.83} & 46.21 & 51.56 & 46.77 &52.60 \\
    RMSE  & \textbf{65.99}& 67.75 & 76.36 & 69.30 &75.99\\
    MAPE  &\textbf{30.13}\% & 34.00\% & 49.00\% & 35.35\% &39.03\%\\
    WMAPE & \textbf{19.59}\%& 20.52\% & 23.04\% & 20.90\% &23.96\%\\
    \midrule
    MAPE\_udt &\textbf{58.85}\% & 66.34\% & 87.51\% & 68.98\% &69.17\%\\
    WMAPE\_udt &\textbf{30.90}\% & 33.04\% & 38.23\% & 33.22\% &34.18\%\\
    \midrule
    MAPE\_dt\_eq &\textbf{22.41}\% & 25.76\% & 29.20\% & 27.09\% &32.72\%\\
    WMAPE\_dt\_eq &\textbf{16.37}\% & 17.47\% & 18.22\% & 18.08\% &22.52\%\\
    \midrule
    MAPE\_dt\_neq &\textbf{26.74}\% & 30.21\% & 56.03\% & 30.43\% &33.27\%\\
    WMAPE\_dt\_neq &\textbf{19.80}\% & 20.39\% & 23.98\% & 20.25\% &23.15\%\\
    \bottomrule
    \end{tabular}%
  \label{ablation}%
\end{table}%

Results of different variations are shown in Tab. \ref{ablation}. 
% By comparing results of the integral STCAGCN in Tab. \ref{PeMS_50_results},
\minew{By comparing results of the full STCAGCN model in the first column,}
it is observed that the performances of all variations are degraded, indicating that each component contributes the overall improvements of STCAGCN model remarkably. 

After removing Tatt-TCN layer, the errors at time-asynchronous locations have an obvious increase, showing that time-asynchronous correlations are indispensable factors of traffic volume estimation in congested area. The ablation study on TDCN manifests that the direction-specific, topology-related characteristics of traffic flow should be considered and incorporated into the computation process of graph convolution and the increments of MAPE\_udt and MAPE\_dt\_eq verify this finding.
Moreover, it is notable that without SPAM, the accuracy decreases sharply on all three types of errors. This indicates that a static speed adjacency matrix is incapable to fully reflect the dynamic, time-varying and non-linear speed-volume relationships. Therefore, defining a speed adjacency matrix subjectively could lead to a biased model with inferior prediction performance. \minew{It is worth commenting that the performance also degrades significantly without the predefined adjacency matrix, especially on the determined and equilibrium locations. This may be because when the road network proximity constraint is missing and only the global correlation is activated, the SPAM may be difficult to optimize due to the large parameter space and fall into local optima.}

\subsubsection{Sensitivity studies on hyper-parameters}
To investigate the affects of hyper-parameters on estimation accuracy, we conduct model sensitivity studies on three main hyper-parameters of STCAGCN, including the diffusion step $K$, number of TCN layers, and hidden dimension. Hidden dimension is the most common parameter in deep learning model and we set the hidden dimension of TCN, TDCN and output layer to the same value. Diffusion step determines the order of neighbors that GCN can aggregate information, which can be viewed as the spatial feaure receptive field. In addition, in our STCAGCN we apply a single TCN layer after the first TDCN layer to capture temporal correlations in a localized window. The more the TCN layers, the longer temporal dependence is extracted. 

\begin{figure}[!htb]
\centering
\subfigure[Influence of hidden dimensions]{
\centering
\includegraphics[scale=0.47]{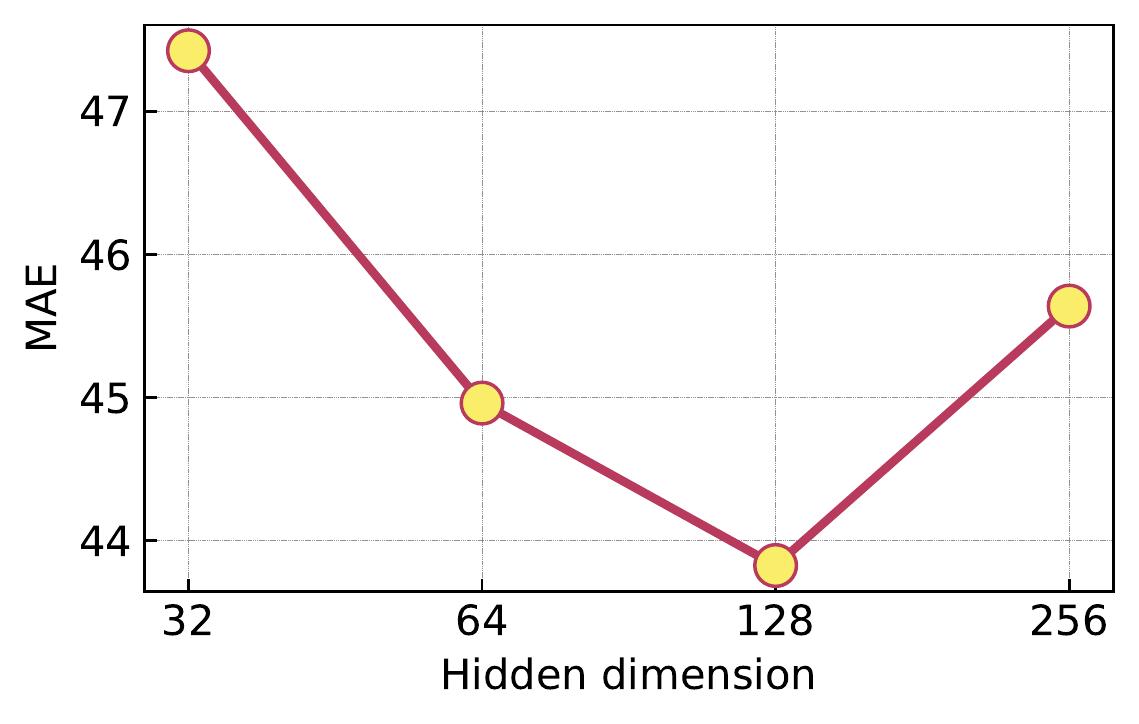}
}
\subfigure[Influence of diffusion steps]{
\centering
\includegraphics[scale=0.47]{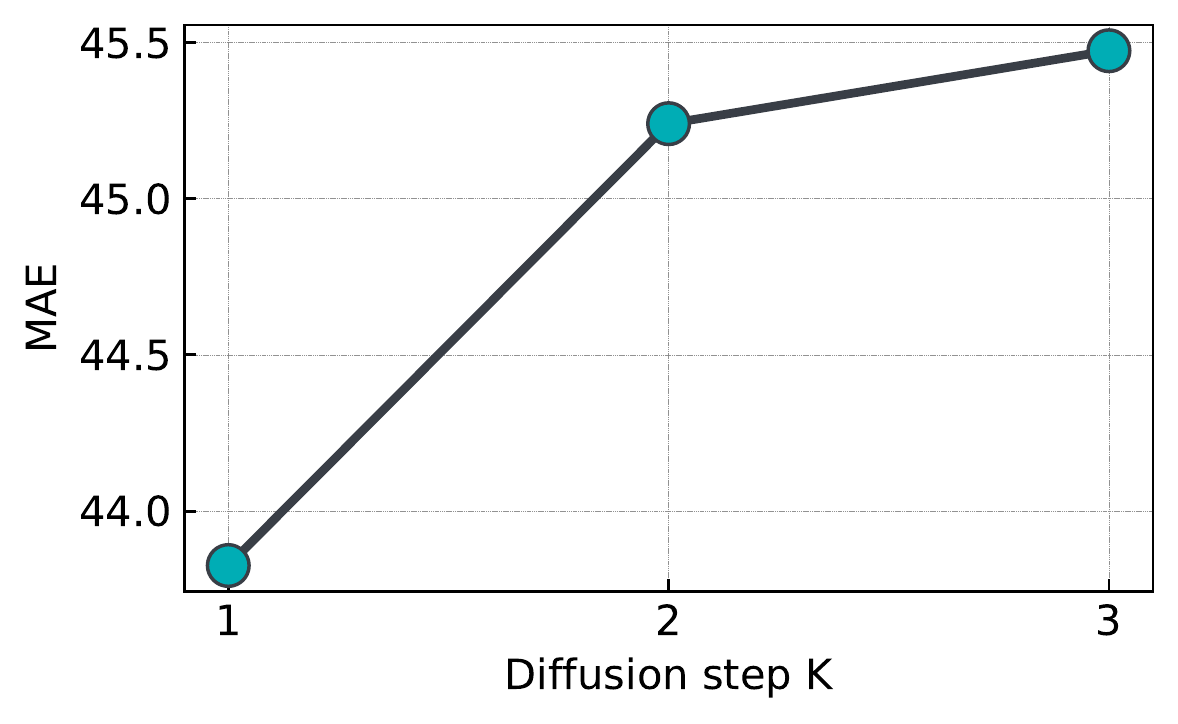}
}
\subfigure[Influence of number of TCN layers]{
\centering
\includegraphics[scale=0.47]{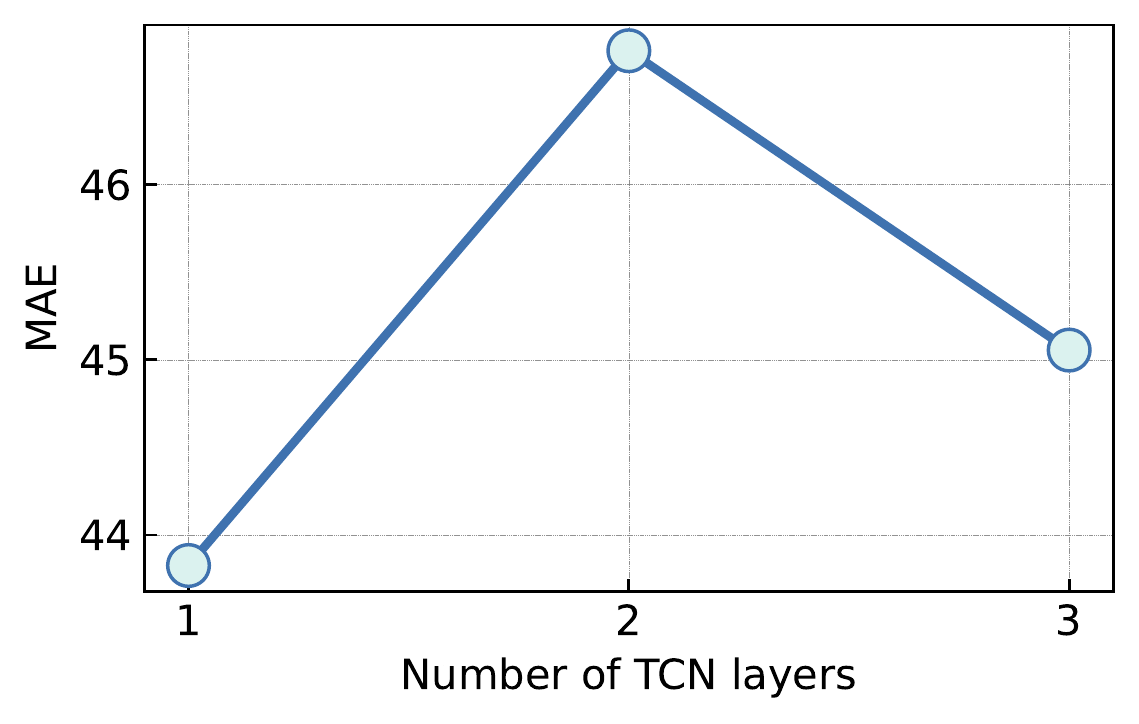}
}
\caption{STCAGCN performances under different hyper-parameter settings.}
\label{sensitivity_results}
\end{figure}

Fig. \ref{sensitivity_results} shows the estimation MAE under $50\%$ missing rate with different settings. First, it can be seen that with the increment of hidden dimension, the estimation error first drops sharply but then increases. A larger hidden dimension enables the GNN model to have more parameters to learn, thereby enhancing the fitting ability. While at the same time, the risk of over-fitting and additional learning complexity may undermine the capacity. Second, with the increase of $K$, the MAE grows gradually. Enlarging $K$ can incorporates irrelevant information from higher-order neighbors and cause over-smoothed node representations.
Third, Fig. \ref{sensitivity_results} (c) shows that stacking more TCN layers has no benefits for estimation results. A rational explanation is that a series of stacked TCN layers are used to capture long-term temporal dependence, while as stated in section \ref{tcn_sec}, the influence of propagation delay happens within a short time period.

\subsubsection{Influence of different sample rates during model training and testing}\label{sec_train_test_rate}

\begin{figure}[!htb]
\centering
\subfigure[Model trained with $50\%$ masking rate tests on different missing rates scenarios]{
\centering
\includegraphics[scale=0.49]{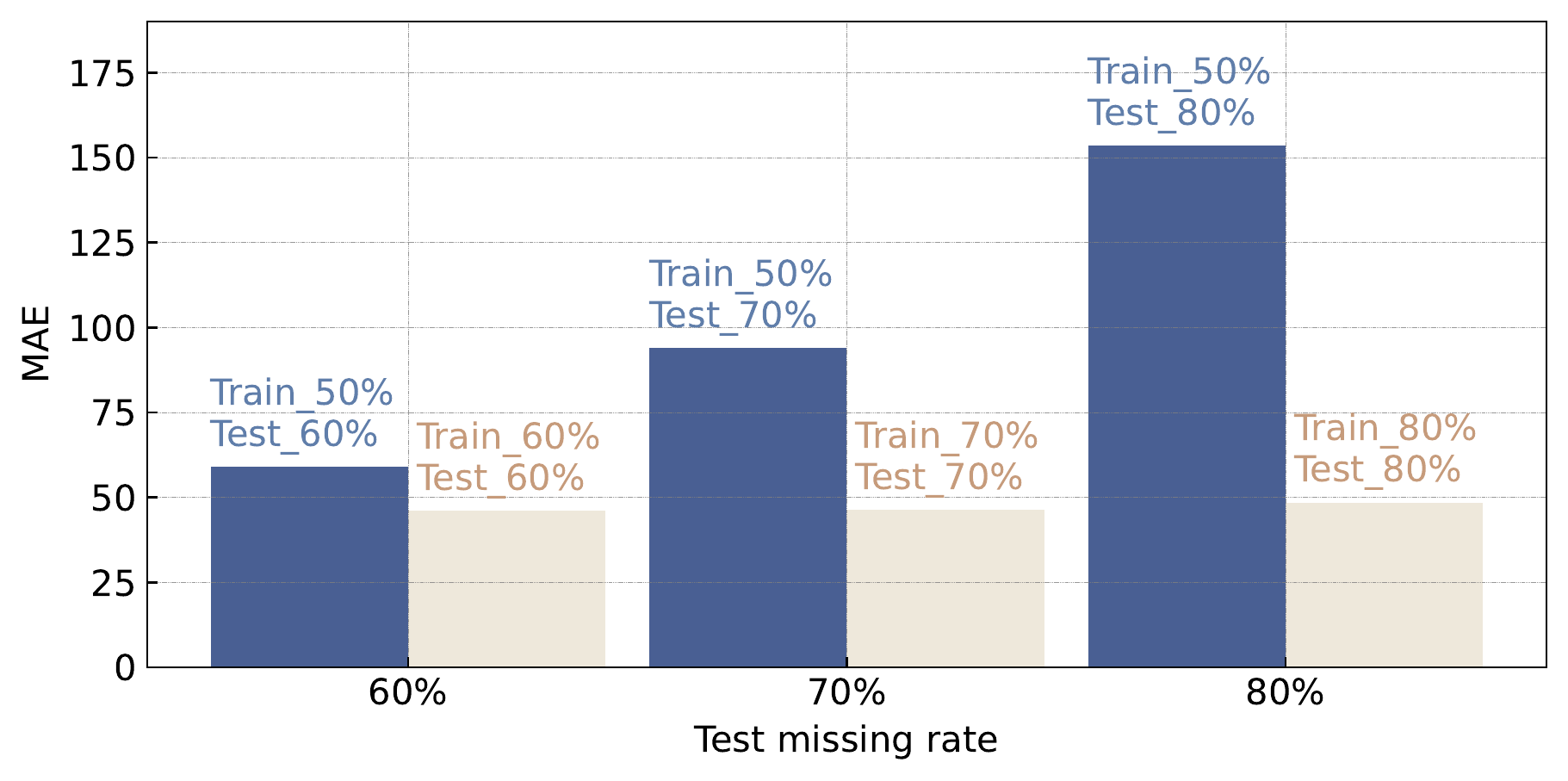}
}
\subfigure[Models trained with different masking rates test on $50\%$ missing rate scenario]{
\centering
\includegraphics[scale=0.48]{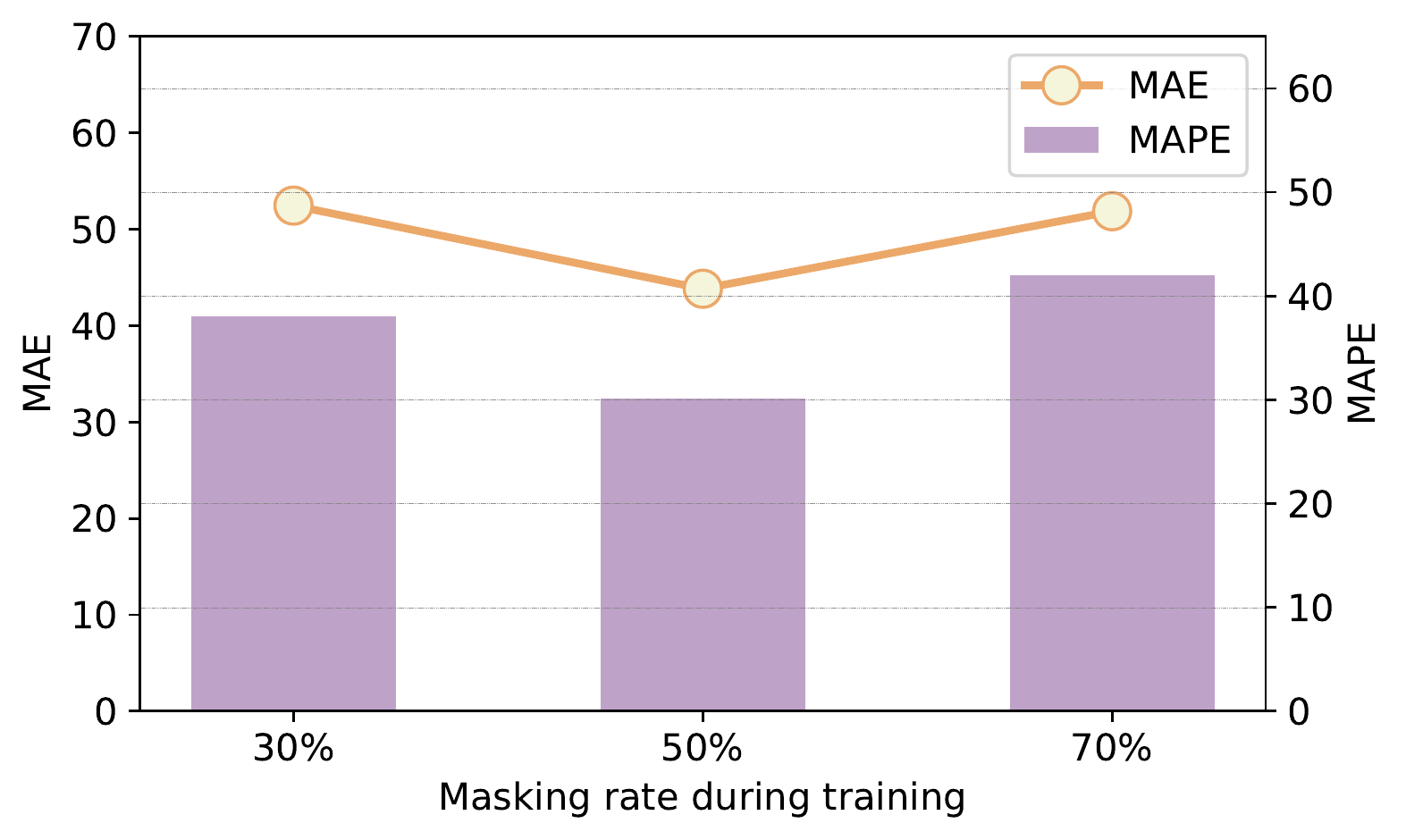}
}
\caption{Influence of different masking and missing rate during training and testing.}
\label{train_test_rate}
\end{figure}

Masking rate is an important hyper-parameter during model training. This parameter controls what proportions of sub-graphs need to be reconstructed during training. We notice that recent works about traffic data imputation or kriging \citep{wu2021inductive,wu2021spatial,liang2022spatial} usually treat it as a general hyper-parameter and ignore its significance. In contrast, we find that it is essential to keep the training masking rate the same as the testing missing rate, in order to train a model with better generalization ability and make the training stage more stable with less fluctuation. 
 
To give evidence of this proposition, we inspect the performances of models trained with different number of masked nodes and tested on different missing rates scenarios. Fig. \ref{train_test_rate} (a) displays the errors of the model trained at $50\%$ masking rate but tested at different missing rates, and the model trained at the same masking rate as the tested missing rate. All these models with different masking rate have obviously worse performances than models with the same masking rate as missing rate. Fig. \ref{train_test_rate} (b) shows the model trained at various masking rates while tested at $50\%$ missing rate and model trained at $50\%$ has the best result.

Both of the two figures reveal the importance of making the masking rate close to the missing rate of the whole sensor graph, thus keeping the distributions of masked sub-graph and actually observed graph as close as possible. We suppose that this setting could alleviate the covariate shift problem of deep learning models \citep{ioffe2015batch}.

\subsubsection{Interpretation of learned graphs}
To demonstrate that SPAM truly leans meaningful patterns, we visualize the learned SPAM at different moment in Fig. \ref{SPAM} and more examples are given in \ref{Appendix_A}. Please note that we remove the self-loop (diagonal) for better visualization and the values of SPAM are normalized to one.
From these figures, we can observe that only a small proportion of nodes have relative high values, which means the correlation graphs of traffic volume is sparse in reality. In this sense, speed information is not always helpful for volume estimation and filtering valid speed data to construct speed-volume relationships in an adaptive way is thus important.

\begin{figure}[!htb]
\centering
\includegraphics[scale=0.38]{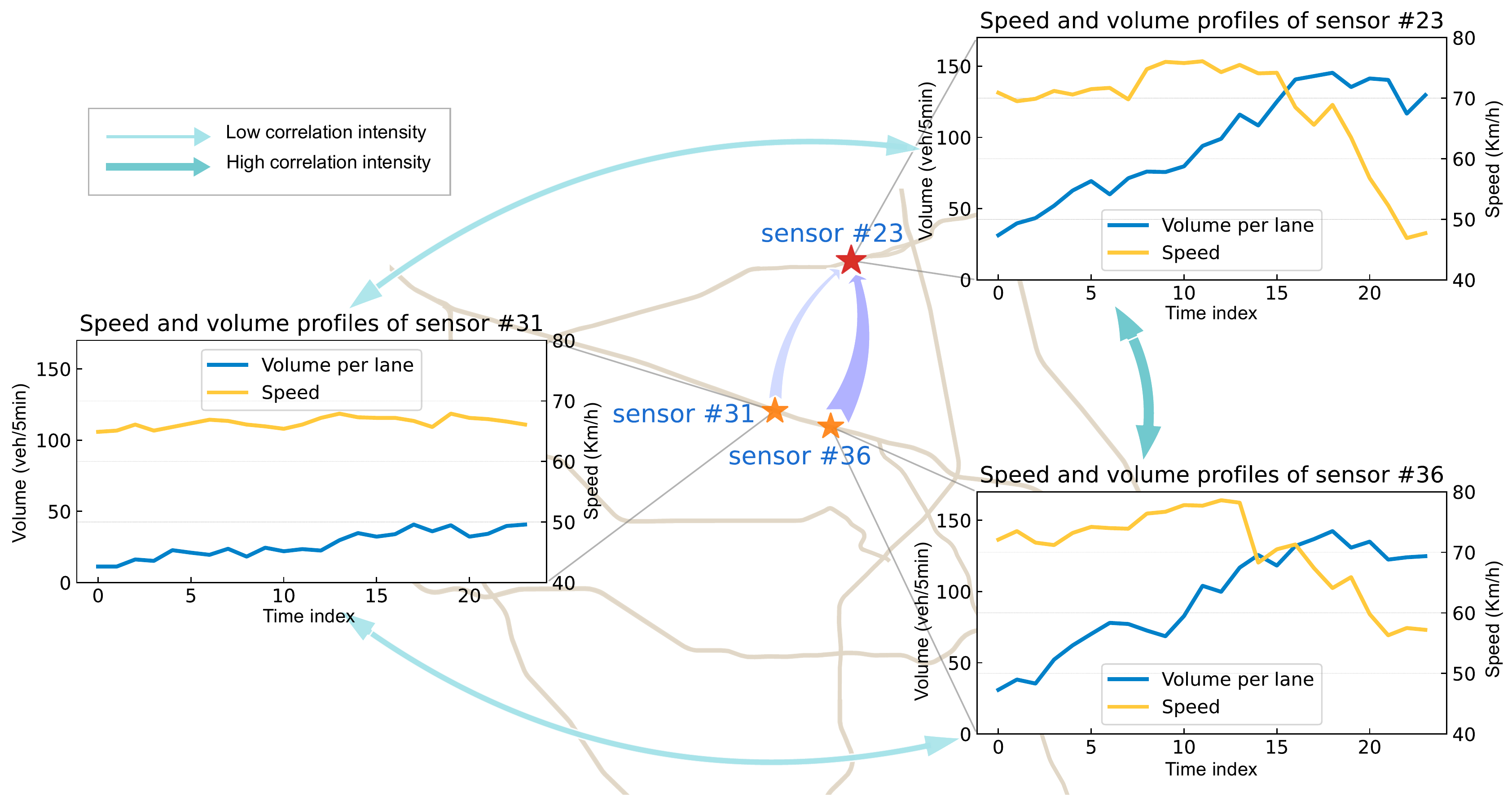}
\caption{Illustration of the non-local correlations learned by SPAM. In this figure, wider arrows with darker color denotes higher intensity of correlation.}
\label{SPAM_on_graph}
\end{figure}

We further take sensors \#23, \#31 and \#36 at $8:00-10:00$ as an example and their locations, speed, and volume (per lane) profiles are visualized on the road network in Fig. \ref{SPAM_on_graph}. 
From Fig. \ref{SPAM} (a), the attention score (correlation) between \#23 and \#36 is much higher than that between \#23 and \#31. At the same time, we can observe similar speed-volume patterns of sensors \#23 and \#36 in Fig. \ref{SPAM_on_graph}, even though they are not proximal on the road network. By seizing the speed similarity, the volume values of \#23 are highly informative for \#36, and vice versa.
Moreover, despite sensors \#31 and \#36 are pretty close in distance, their volume series show very different patterns. This inconsistency phenomenon prompts us to seek for non-local correlations that can overcome the deficiency of typical graph convolution.

Therefore, these observations suggest that the SPAM successfully captures non-local correlations between sensors by dynamically learning the speed-volume relationships. Compared with static speed adjacent used in previous works, our SPAM is agnostic to the form of speed-volume function and robust to misleading information, which is much more flexible and reliable.

\begin{figure}[!htb]
\centering
\subfigure[SPAM at $8:00-10:00$]{
\centering
\includegraphics[scale=0.26]{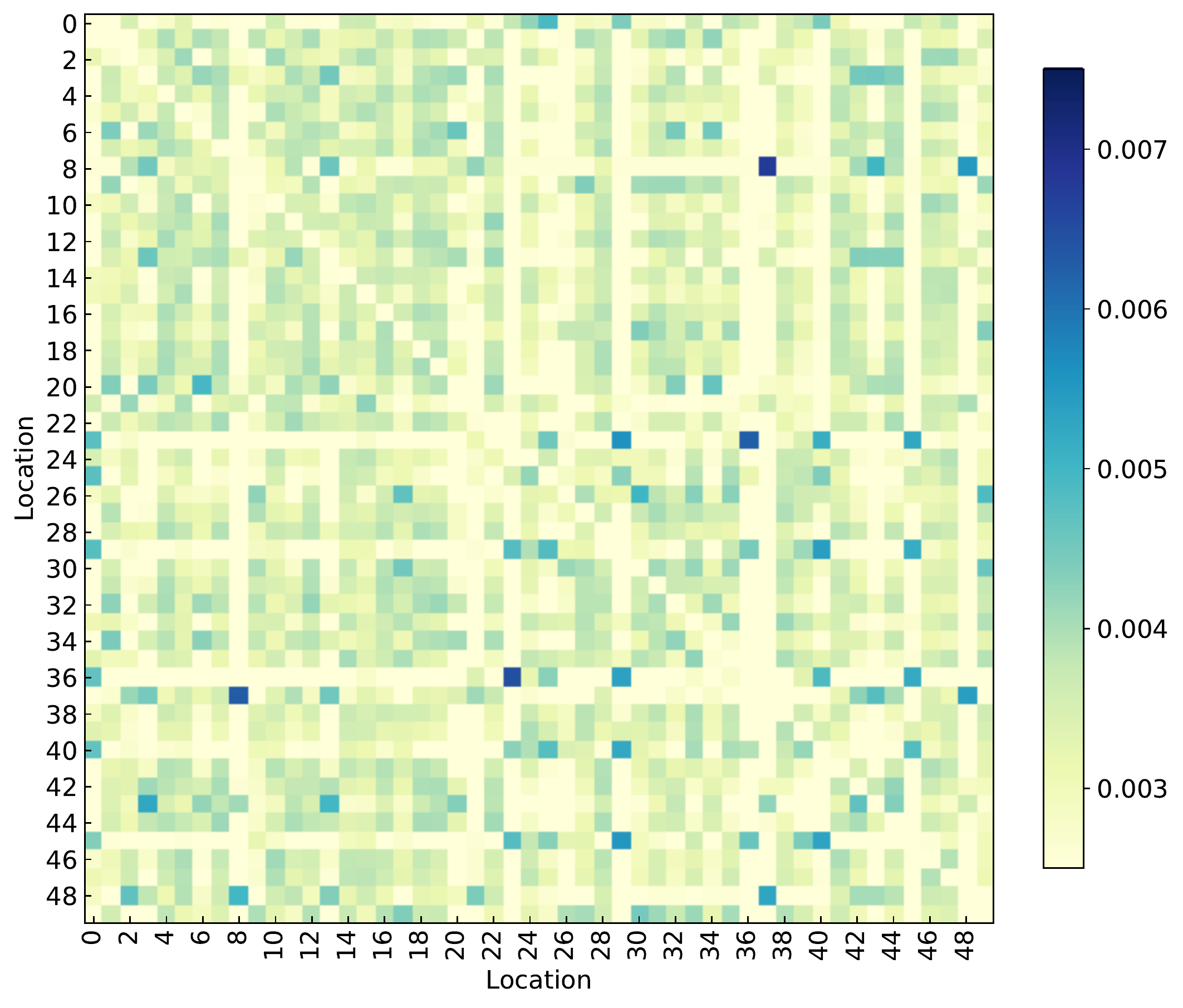}
}
\subfigure[SPAM at $10:00-12:00$]{
\centering
\includegraphics[scale=0.26]{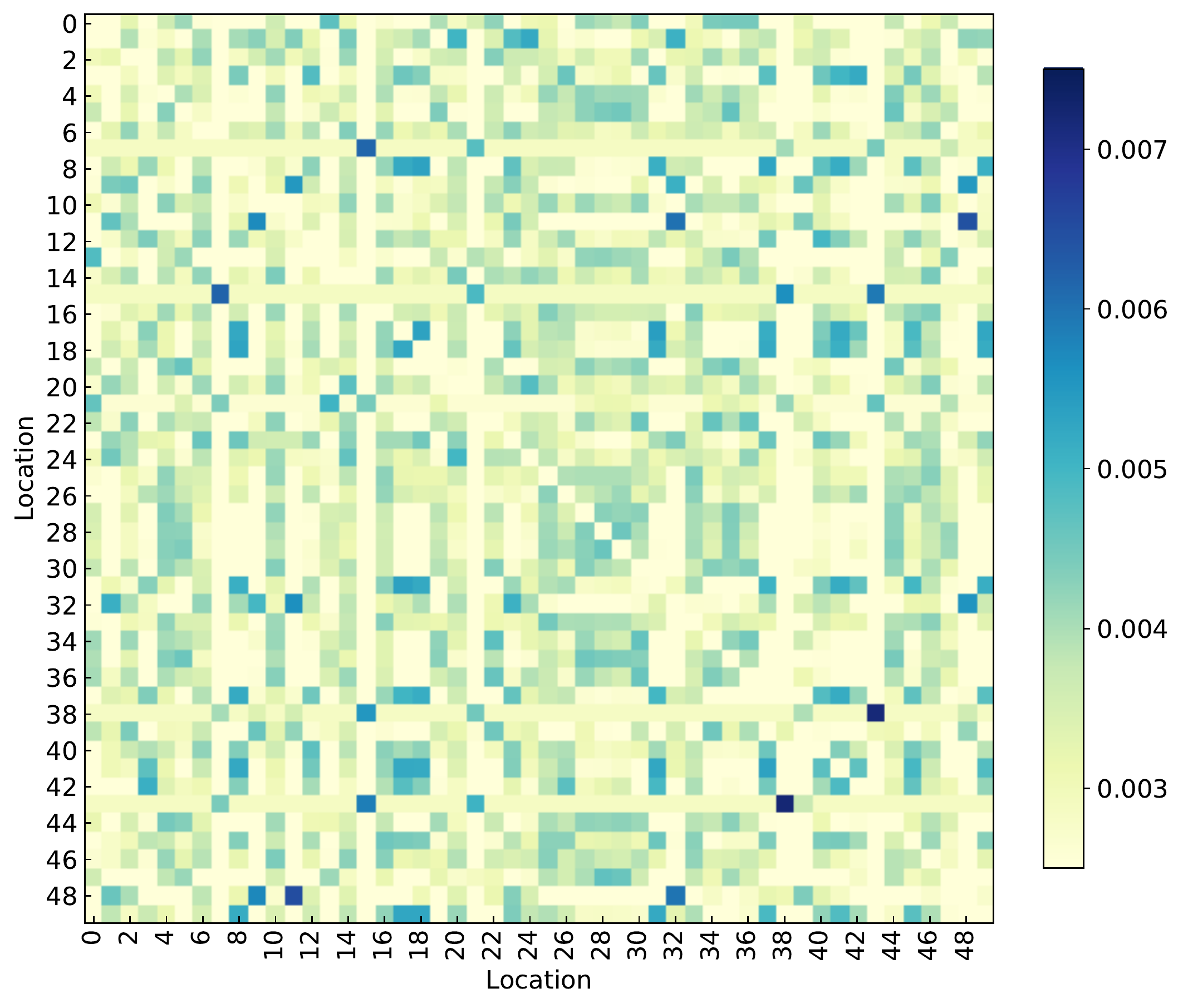}
}
\subfigure[SPAM at $14:00-16:00$]{
\centering
\includegraphics[scale=0.25]{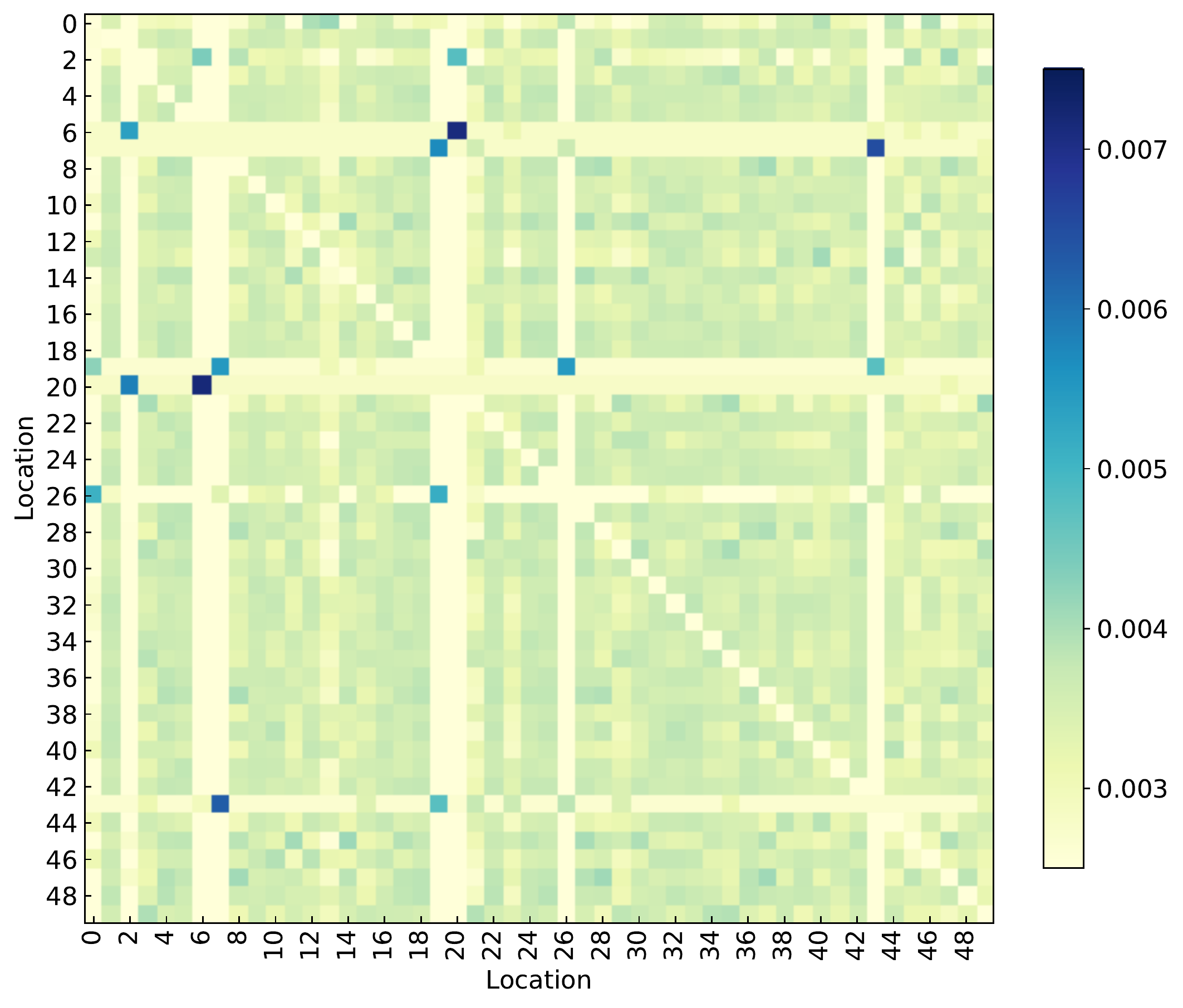}
}
\subfigure[SPAM at $16:00-18:00$]{
\centering
\includegraphics[scale=0.25]{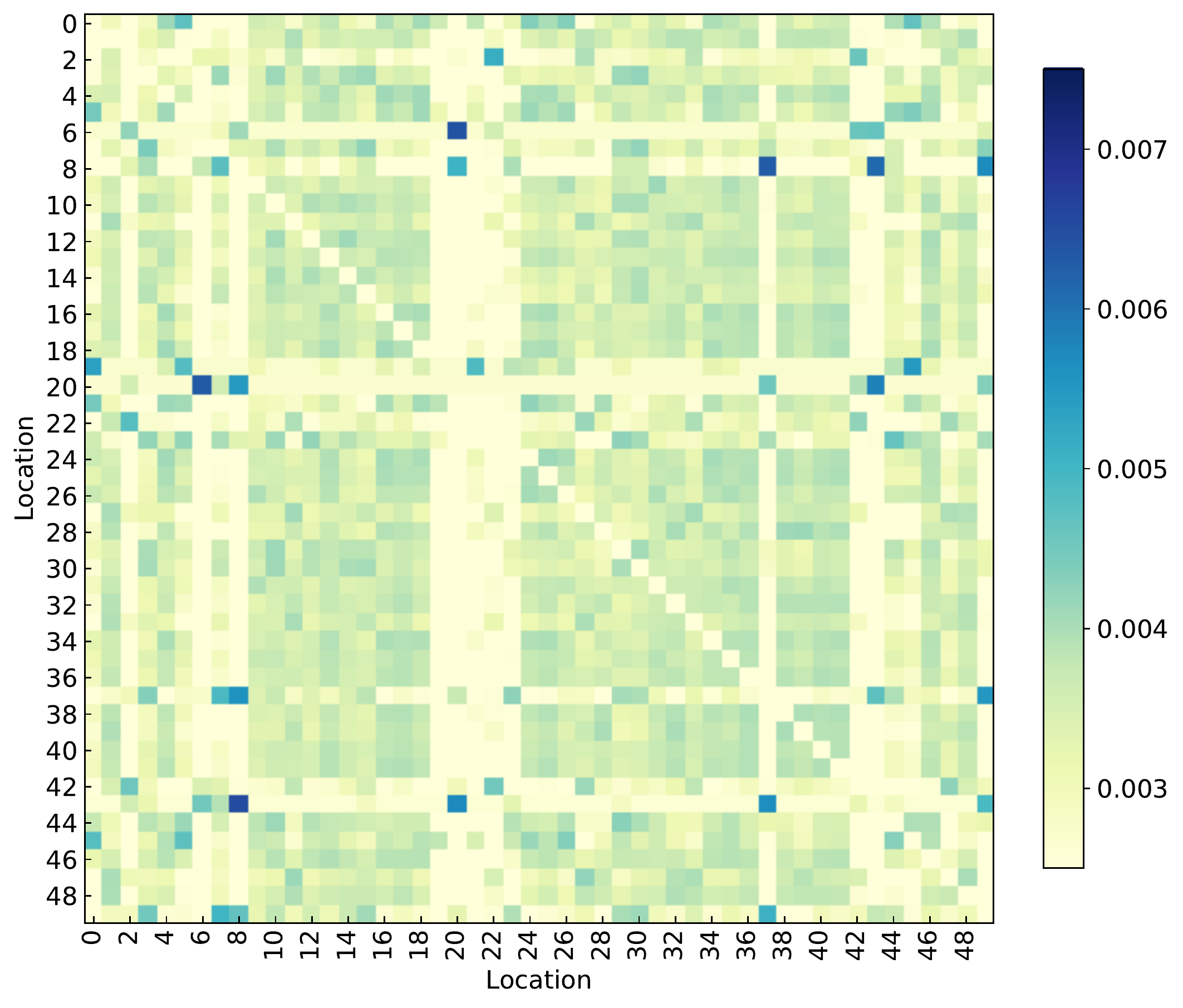}
}
\caption{SPAM at different time point on test set. Note that only the first 50 sensors are displayed for better viewing.}
\label{SPAM}
\end{figure}

\section{Conclusion and future directions}\label{conclusions}
Acquiring fine-grained spatiotemporal traffic volume data is of great challenge due to the limited deployment of traffic detectors. The underdetermined and nonequilibrium phenomena further complicate the requirements of utilizing available observations. To tackle these issues, in this paper we elaborate a correlation-adaptive graph convolution network named STCAGCN that features both non-local spatial dependence and time-asynchronous correlation for highway traffic volume estimation under sparse sensor coverage. To consider the impacts of road geometry and propagation direction, we develop a task-specific diffusion convolution module to model the semantic similarities between sensors. To leverage the dynamic, nonlinear and time-varying speed-volume relationships for enhancing the estimation of underdetermined flows, we propose a graph attention based speed pattern-adaptive learning module to encode the speed information as a complement. To deal with nonequilibrium flows caused by congestion propagation, we develop a temporal masked self-attention model combined with a localized temporal convolution layer to exploit time-asynchronous correlations adaptively. Detailed case studies and model analysis are conducted on a real-world highway traffic volume dataset, and several conclusions can be drawn as follows:
\begin{enumerate}[(1)]
    \item Results compared with other baseline models including state-of-the-art deep learning and tensor learning methods on different observation conditions, illustrate obvious superiority of our STCAGCN model on both overall accuracy and critical scenarios.
    \item Model ablation studies further justify the rationality of model designs. Especially, SPAM captures the global correlations of volume to provide complementary side information for underdetermined flows. And Tatt-TCN can seize the time-shifted correlations under varying impacts of congestion delay.
    \item As our model achieves graph reconstruction based on randomly masked sub-graphs, random sampling plays a key role in model generalization capability. Studies on sampling rate indicate the necessity of keeping the masking rate in training stage and missing rate in testing stage the same.
    \item It is well known that there exists complicated mapping relationships between traffic speed and volume. Our model captures this subtle connection well and interpretations of learned spatial correlation graphs reveal the globally shared volume patterns by extracting from the underlying speed-volume relationships.
\end{enumerate}

In addition, there are still some directions require future efforts.
First, how to achieve joint estimation of incomplete traffic speed and volume is worth trying. A possible solution is to impute the incomplete speed data at the same time, with speed reconstruction errors included in the loss function. Second, as collecting link volume data of urban road network is much more difficult, further efforts are needed to conduct case studies on urban networks.

% performing path flow estimation for urban road networks is still an open question. 

% In this case, this problem should be viewed as spatial regression rather than kriging, and more factors such as signal timing and road class should be taken into account.

\appendix
\section*{Appendix}

\section{Interpretations on WDSSI}
\label{Appendix_C}
\begin{figure}[!htb]
\centering
\includegraphics[scale=0.46]{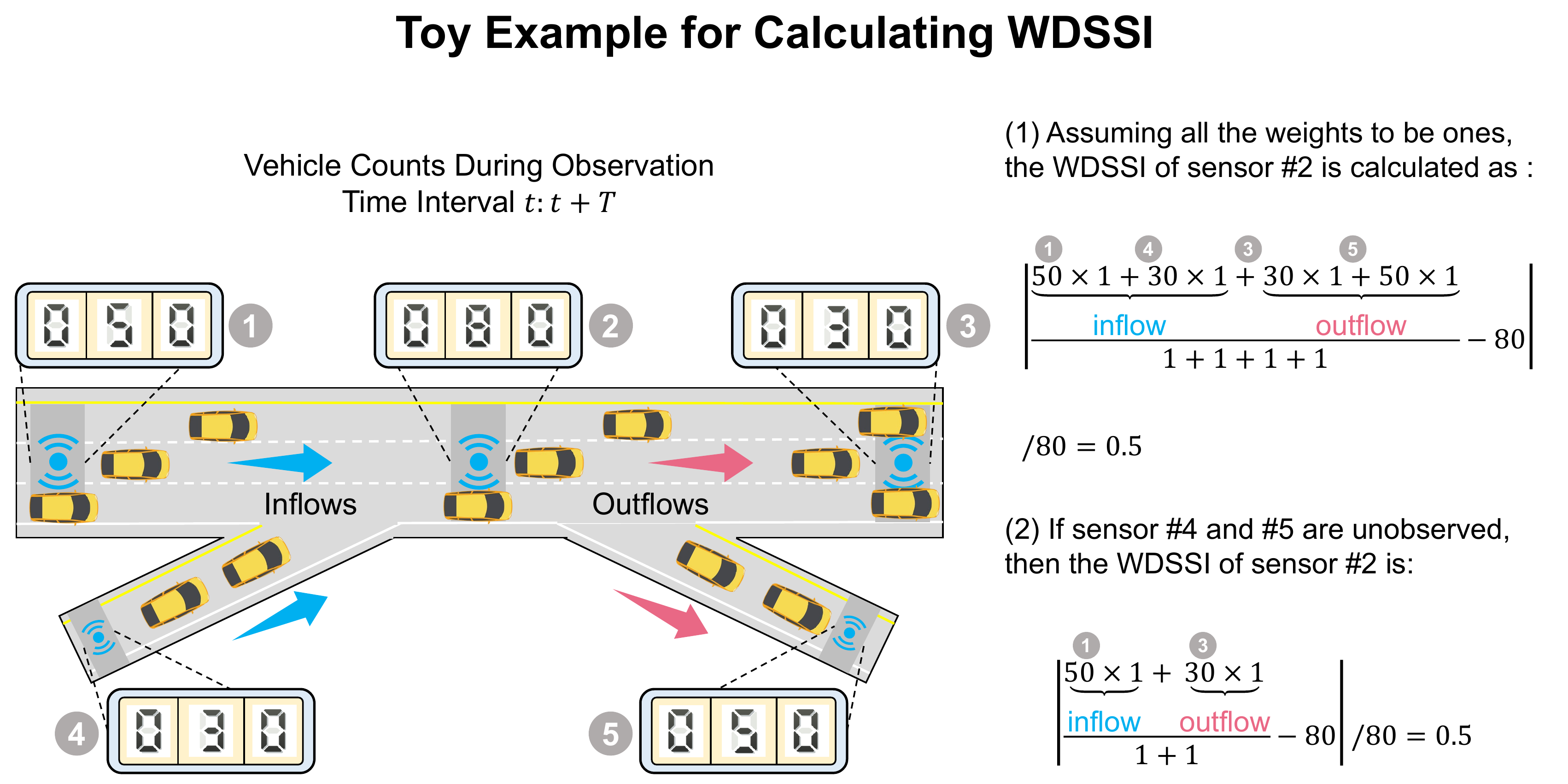}
\caption{\minew{Toy example of WDSSI calculation of on/off ramp regions.}} 
\label{WDSSI_example}
\end{figure}

\minew{WDSSI is inspired by the definition of graph Laplacian regularization in machine learning research \citep{kondor2002diffusion}. This metric measures the smoothness of the subgraph and describes how the sensor's reading is close to the weighted average of sensors in the surrounding area. As the distance-based adjacency graph includes surrounding sensors within a certain range, it can exploit high-order neighborhood structures. In addition, to reflect the long-term pattern, we obtain the WDSSI of a sensor by the average over the whole observation period. For highway trunk and corridor, this value should be close to zero. For on/off ramp segment, it can be a large value. Fig. \ref{WDSSI_example} provides a toy example to show these cases.
}

\minew{Both of the two cases result in a relatively large WDSSI value, and the latter corresponds to the locations with underdetermined flows, which are more challenging in volume estimation.}

\section{Theoretical justifications on the speed-volume adaptive correlation}
\label{Appendix_B}
\minew{
We provide more interpretations and discussions on Eq. \ref{graph_attention}. Recall that the graph attention technique in Eq. \ref{graph_attention} is developed to capture the nonlinear and global traffic volume correlations by introducing traffic speed data. We can derive this formulation from the perspective of speed-volume relationships in traffic flow theory. 

Empirically, the relationship between flow rate and spot speed obeys a nonlinear function \citep{geroliminis2011properties}. Let $\mathbf{s}_i,\mathbf{v}_i\in\mathbb{R}^T$ denote the speed and volume observations of sensor $i$, the volume is indicated as a nonlinear function of speed:
\begin{equation}
    \mathbf{v}_i=\mathcal{F}(\mathbf{s}_i),
\end{equation}
\noindent where the mapping $\mathcal{F}(\cdot)$ can either be node-specific or shared across all locations. Then the correlation of volume between a sensor pair is:
\begin{equation}\label{corr}
    \texttt{corr}(\mathbf{v}_i,\mathbf{v}_j)=\texttt{corr}(\mathcal{F}(\mathbf{s}_i),\mathcal{F}(\mathbf{s}_j)).
\end{equation}

Considering that we can use a multilayer perceptron (MLP) to approximate any nonlinear function, after omitting the bias of MLP and using the cosine similarity function as the correlation function $\texttt{corr}(\cdot)$, we can reformulate Eq. \ref{corr} as follows:
\begin{equation}\label{corr_appendix}
\texttt{corr}(\mathbf{v}_i,\mathbf{v}_j)=\frac{\sigma(\mathbf{Ws}_i)^\mathsf{T}\cdot\sigma(\mathbf{Ws}_j)}{\Vert\sigma(\mathbf{Ws}_i)\Vert\Vert\sigma(\mathbf{Ws}_j\Vert}.
\end{equation}

Finally, by using the \texttt{LeakyReLU} nonlinearity and the \texttt{SoftMax} function to ensure the normalization, Eq. \ref{corr_appendix} yields exactly the graph attention computation in Eq. \ref{graph_attention}. Therefore, we theoretically prove that SPAM can be exploited to capture the nonlinear and global volume correlations with available speed information.
}

\section{Supplementary figures}\label{Appendix_A}
We give more examples about the learned SPAM in the following figure \ref{SPAM_sup}.
\begin{figure}[!htbp]
\centering
\subfigure[SPAM of test sample \#19]{
\centering
\includegraphics[scale=0.18]{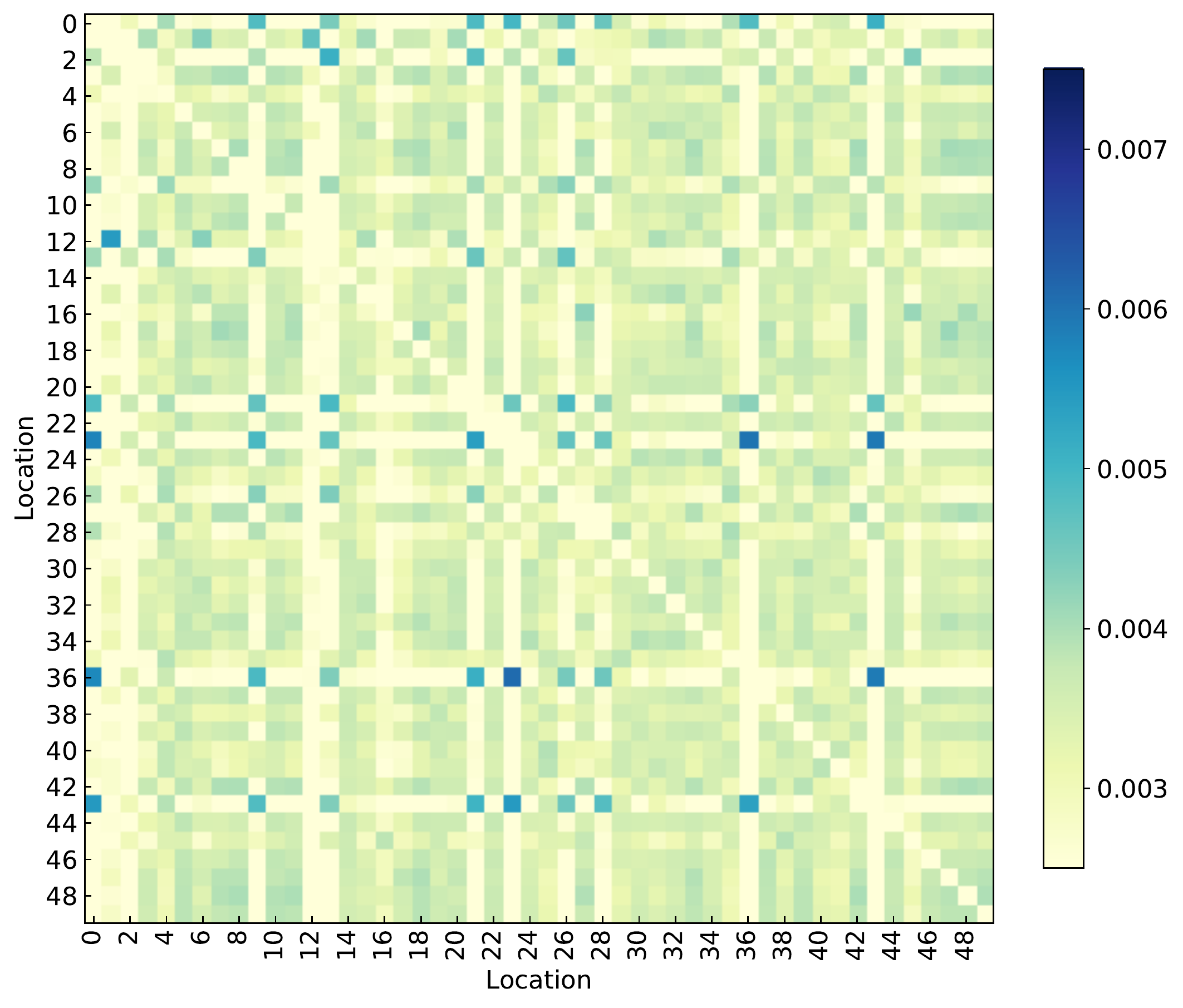}
}
\subfigure[SPAM of test sample \#21]{
\centering
\includegraphics[scale=0.18]{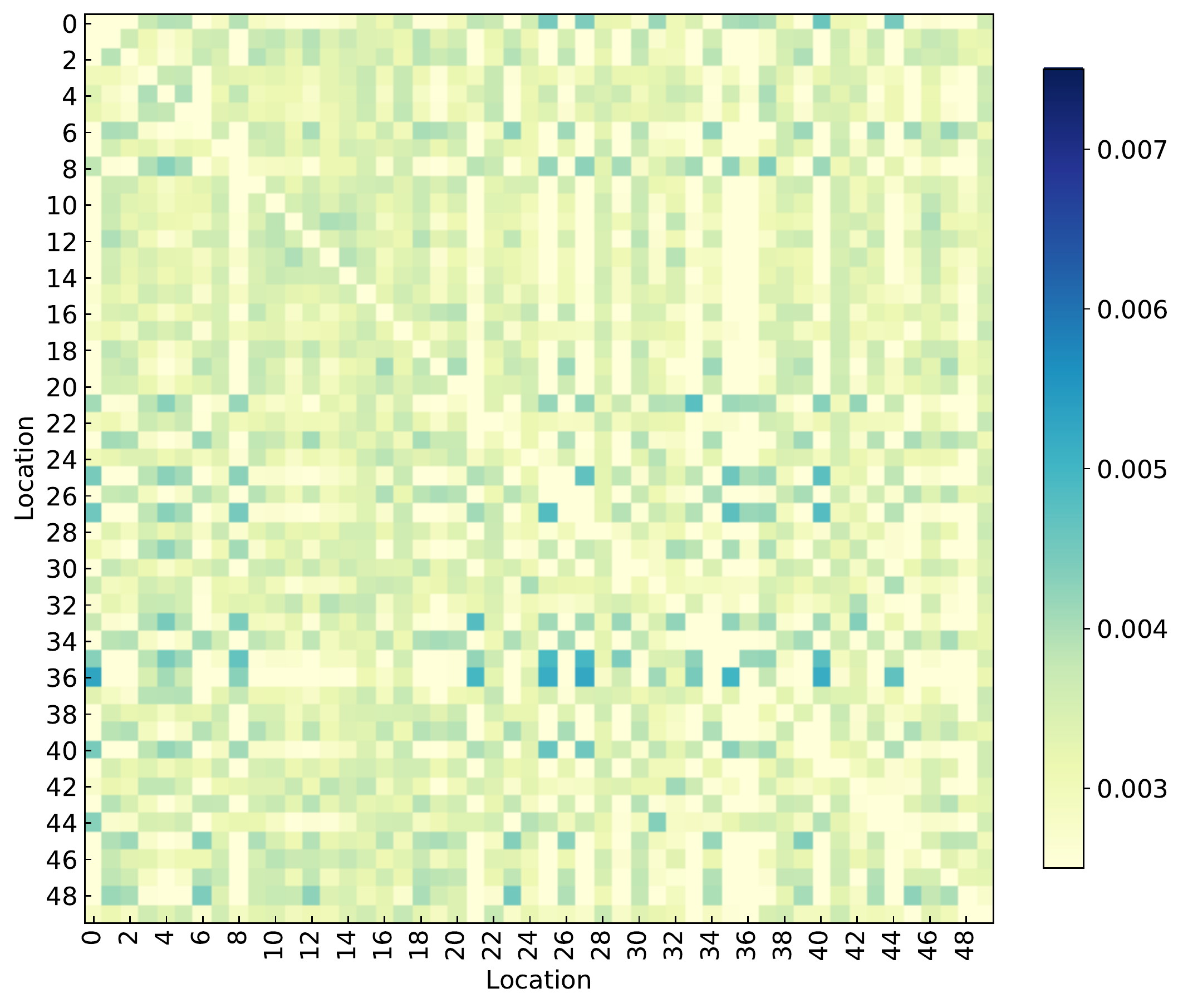}
}
\subfigure[SPAM of test sample \#24]{
\centering
\includegraphics[scale=0.18]{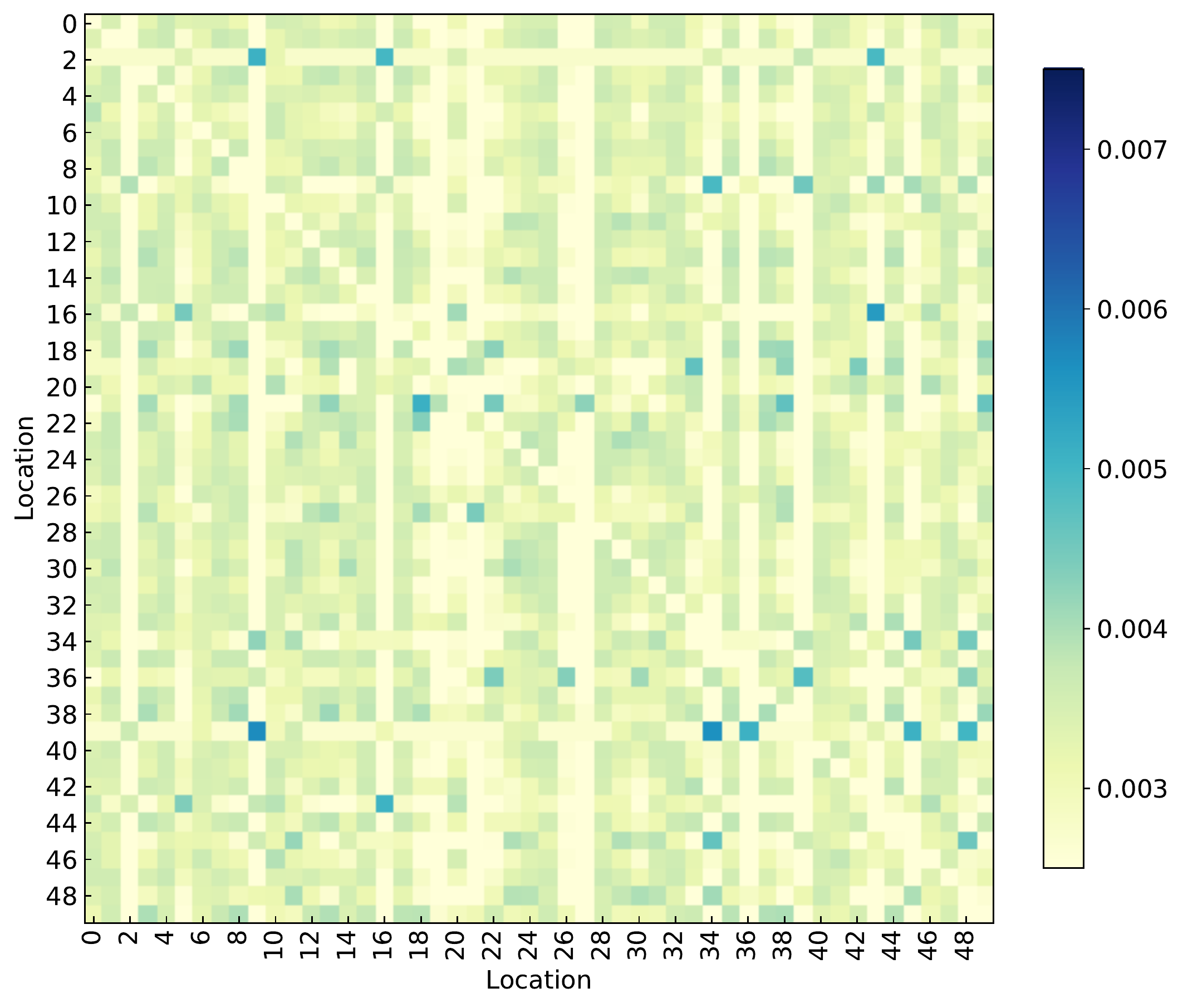}
}
\subfigure[SPAM of test sample \#25]{
\centering
\includegraphics[scale=0.18]{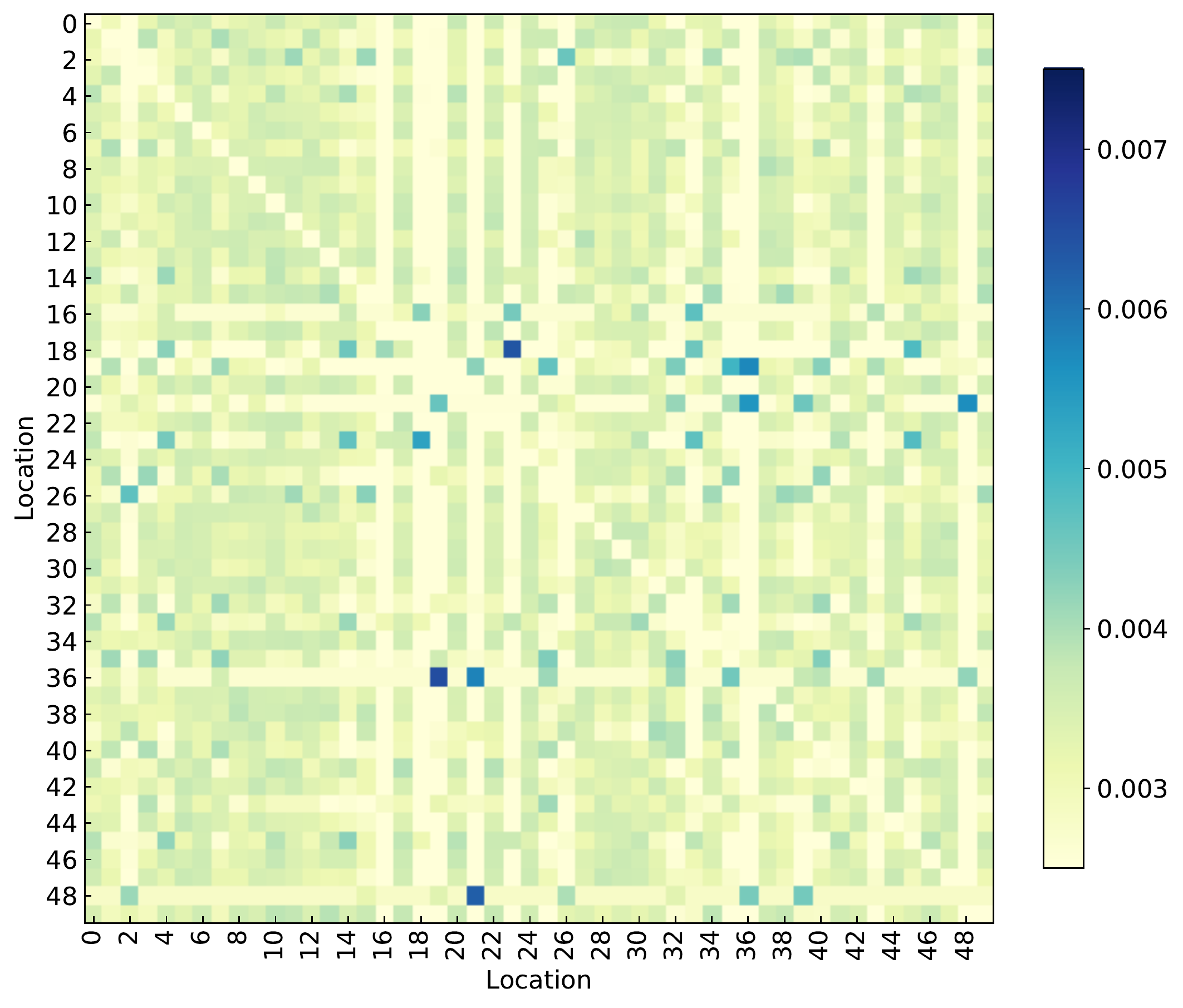}
}
\subfigure[SPAM of test sample \#31]{
\centering
\includegraphics[scale=0.18]{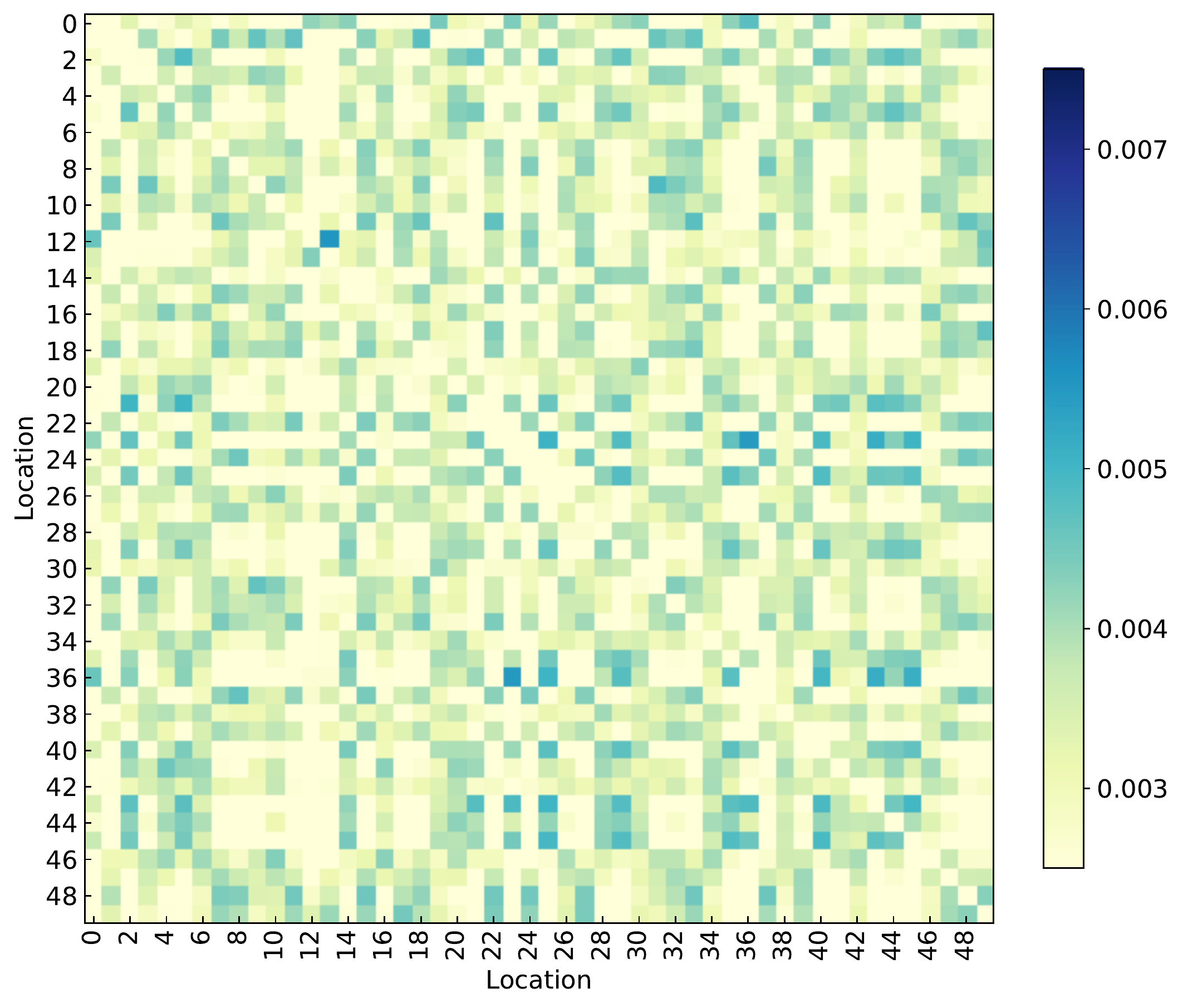}
}
\subfigure[SPAM of test sample \#32]{
\centering
\includegraphics[scale=0.18]{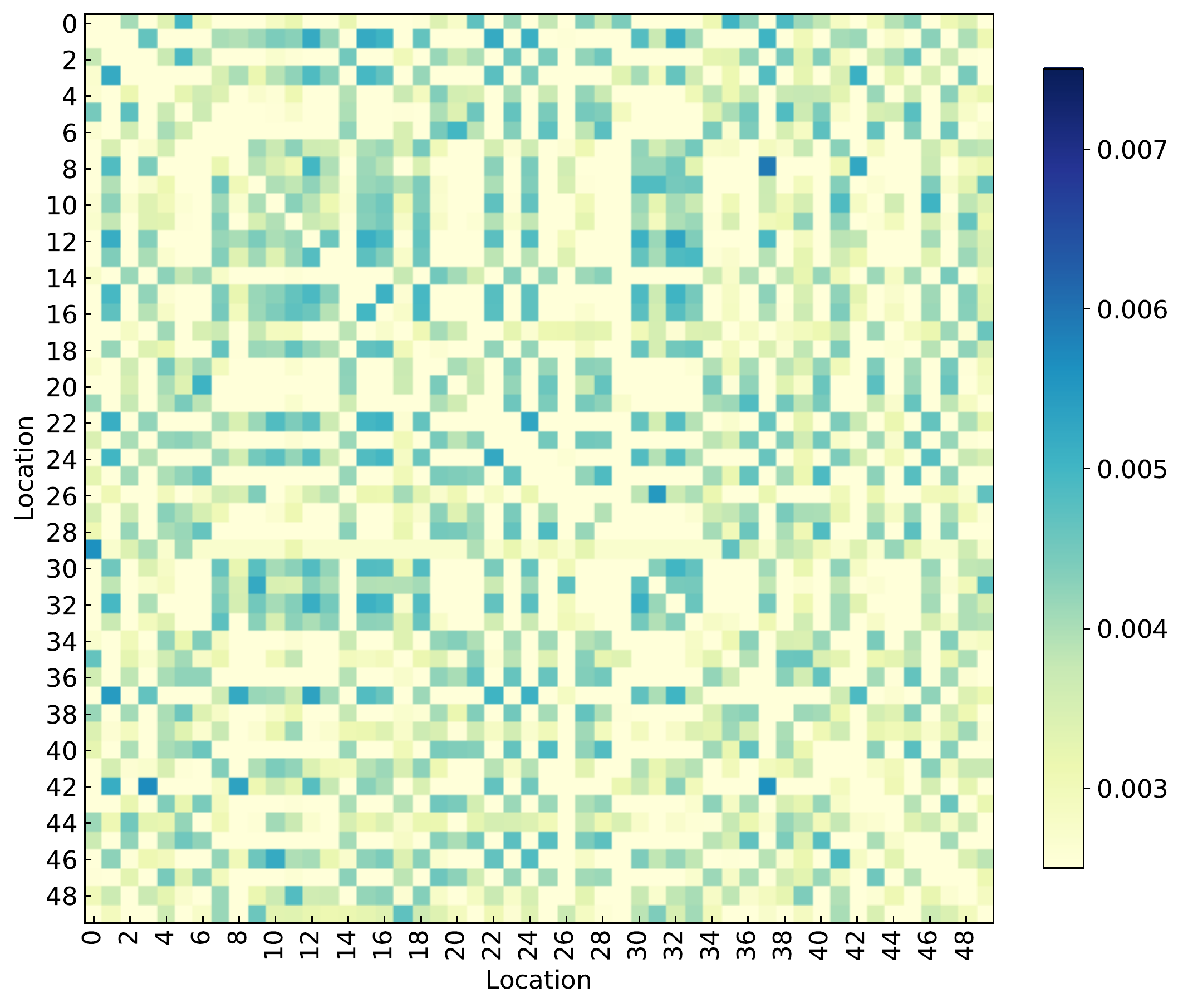}
}
\subfigure[SPAM of test sample \#35]{
\centering
\includegraphics[scale=0.18]{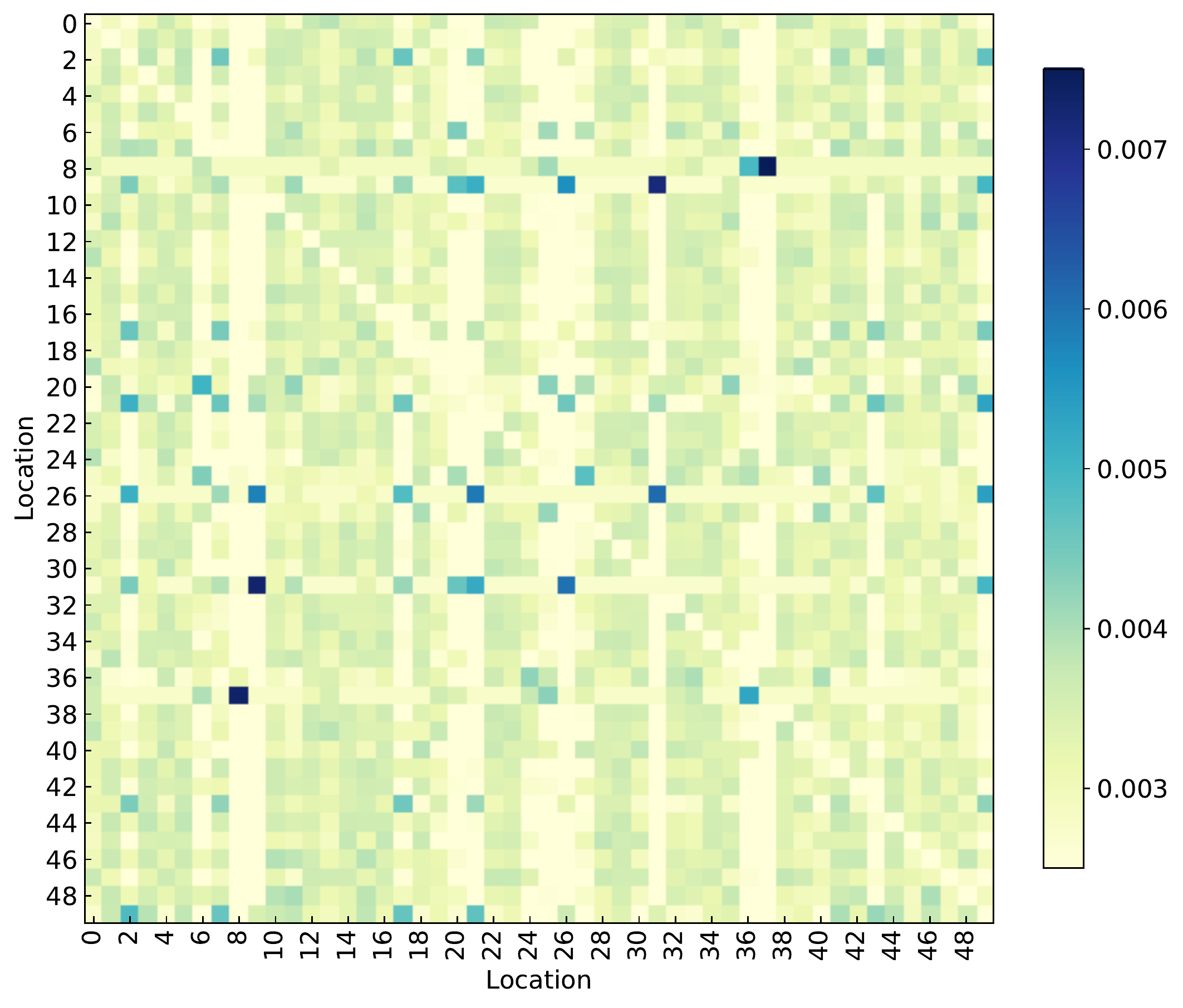}
}
\subfigure[SPAM of test sample \#39]{
\centering
\includegraphics[scale=0.18]{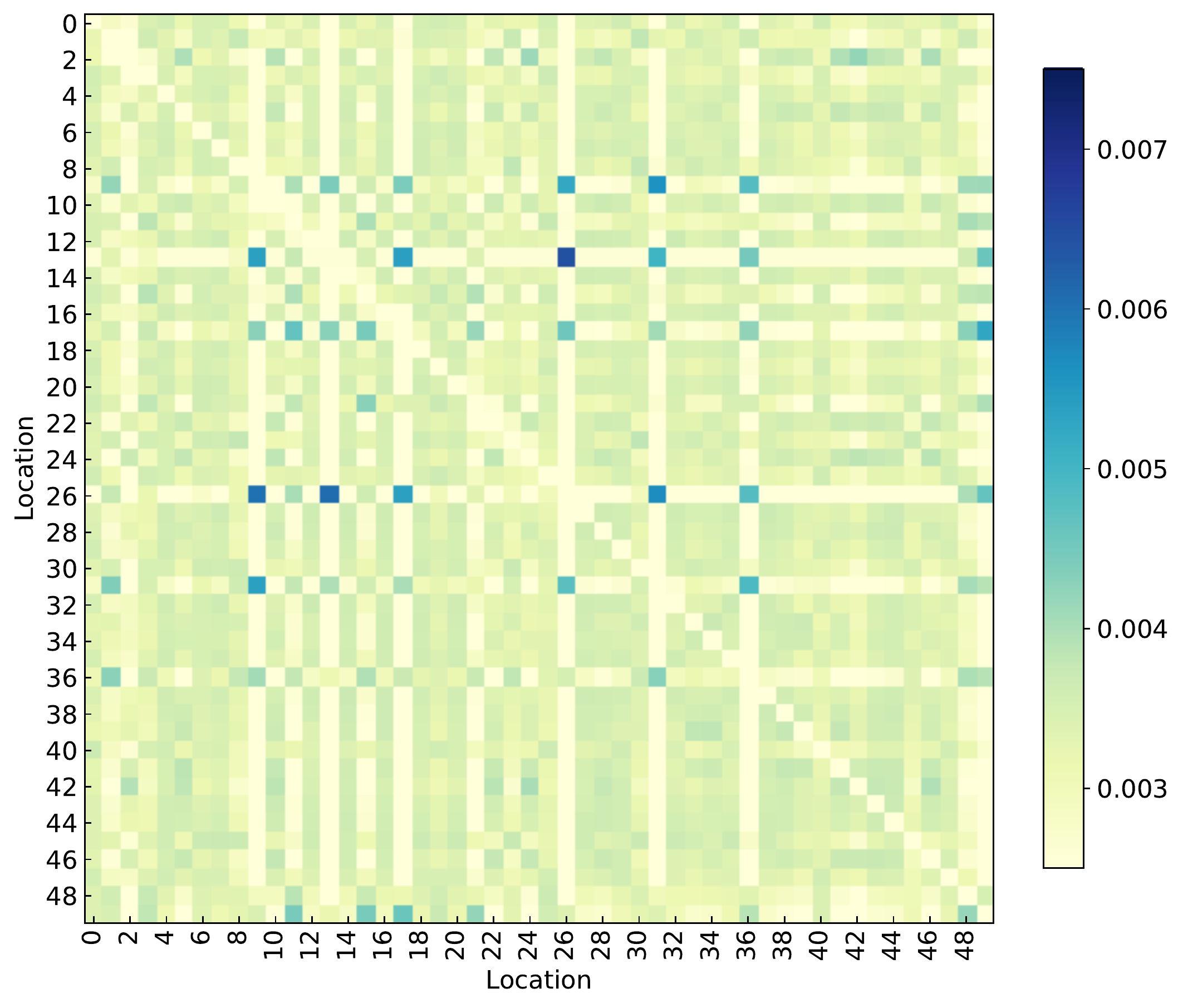}
}
\caption{More examples of the learned SPAM on test samples.}
\label{SPAM_sup}
\end{figure}

\begin{figure}[!htb]
\centering
\subfigure[50\% missing rate]{
\centering
\includegraphics[scale=0.4]{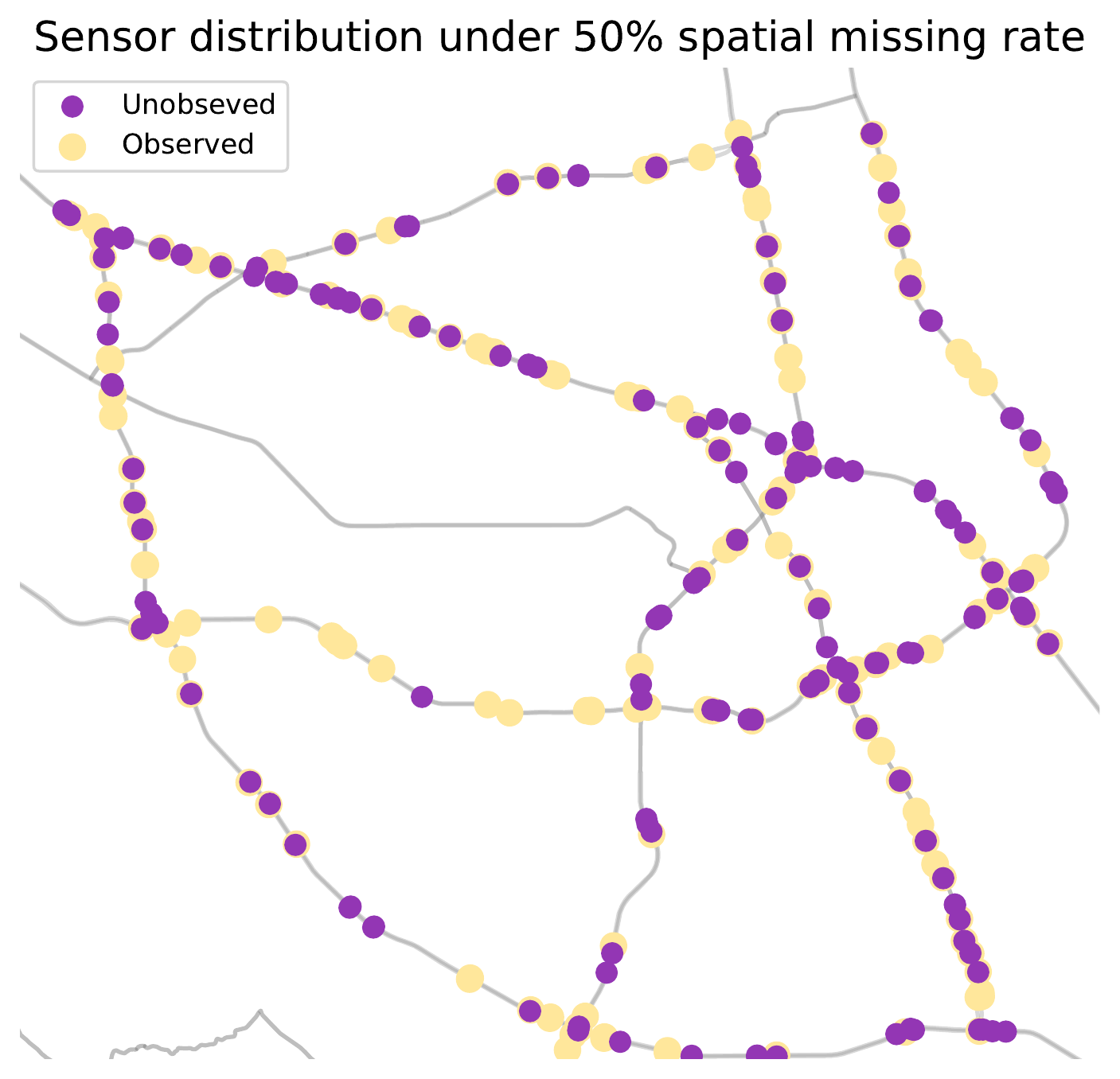}
}
\subfigure[60\% missing rate]{
\centering
\includegraphics[scale=0.4]{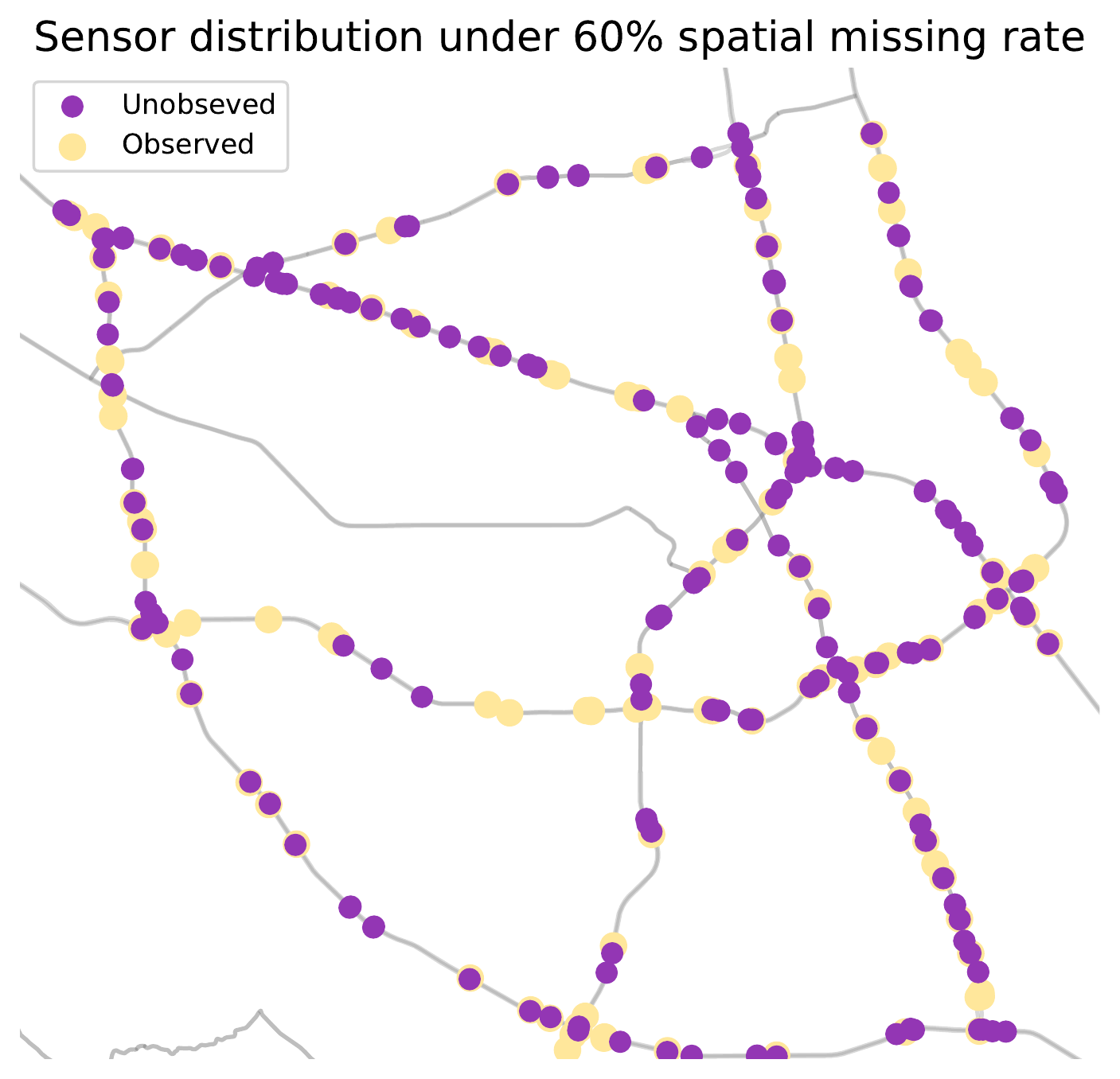}
}
\subfigure[70\% missing rate]{
\centering
\includegraphics[scale=0.4]{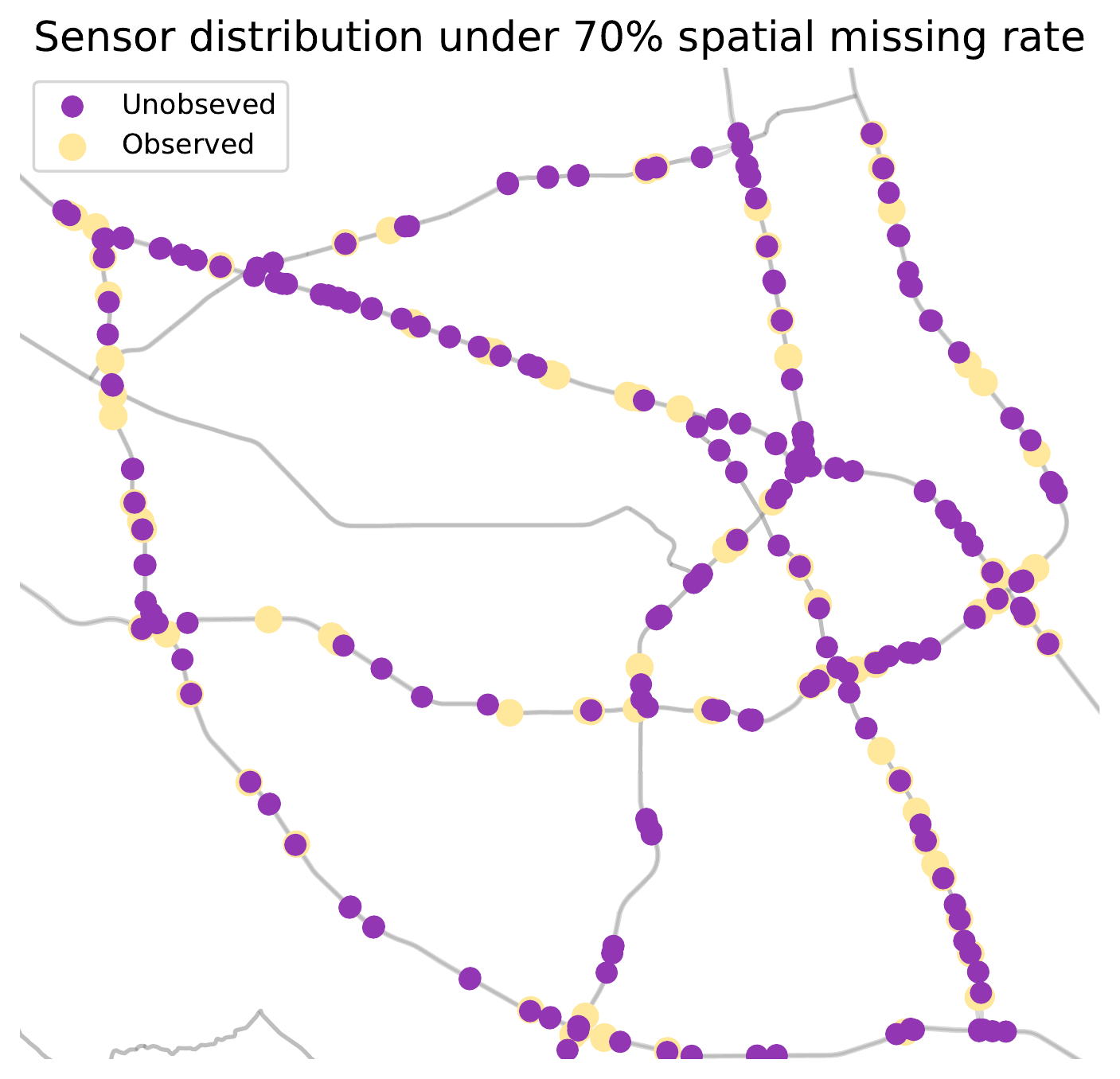}
}
\subfigure[80\% missing rate]{
\centering
\includegraphics[scale=0.4]{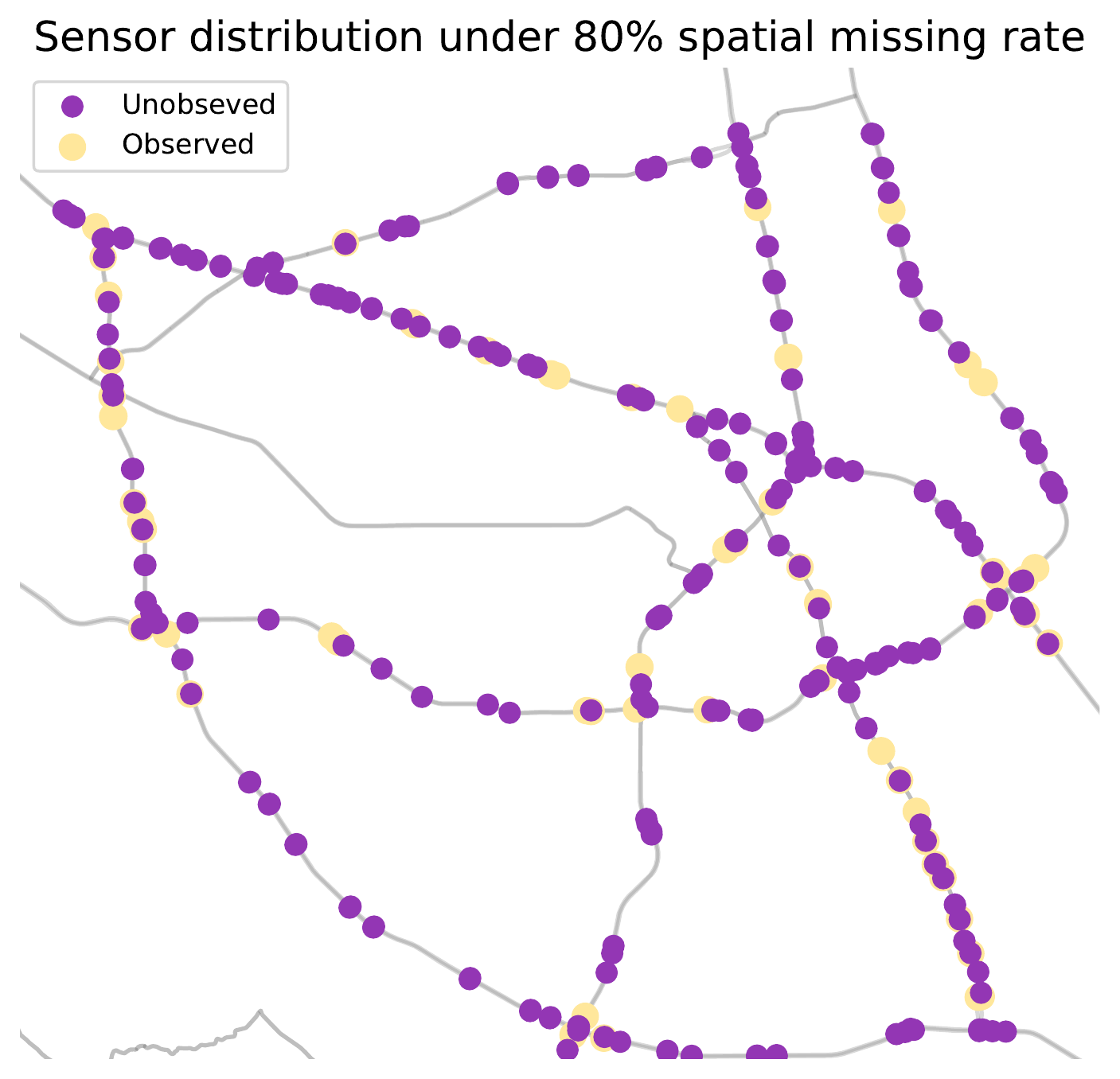}
}
\caption{\minew{Sensor distribution under different missing rates.}}
\label{distribution_sup}
\end{figure}

\minew{The spatial distributions of traffic sensors under different coverage (missing) rates are displayed in Fig. \ref{distribution_sup}. The unobserved sensors are randomly selected from the whole sensor set with a uniform distribution, which corresponding to a ``slice missing'' case in the literature \citep{yokota2018missing,nie2023correlating}.}

\begin{figure}[!htb]
\centering
\subfigure[60\% missing rate]{
\centering
\includegraphics[scale=0.33]{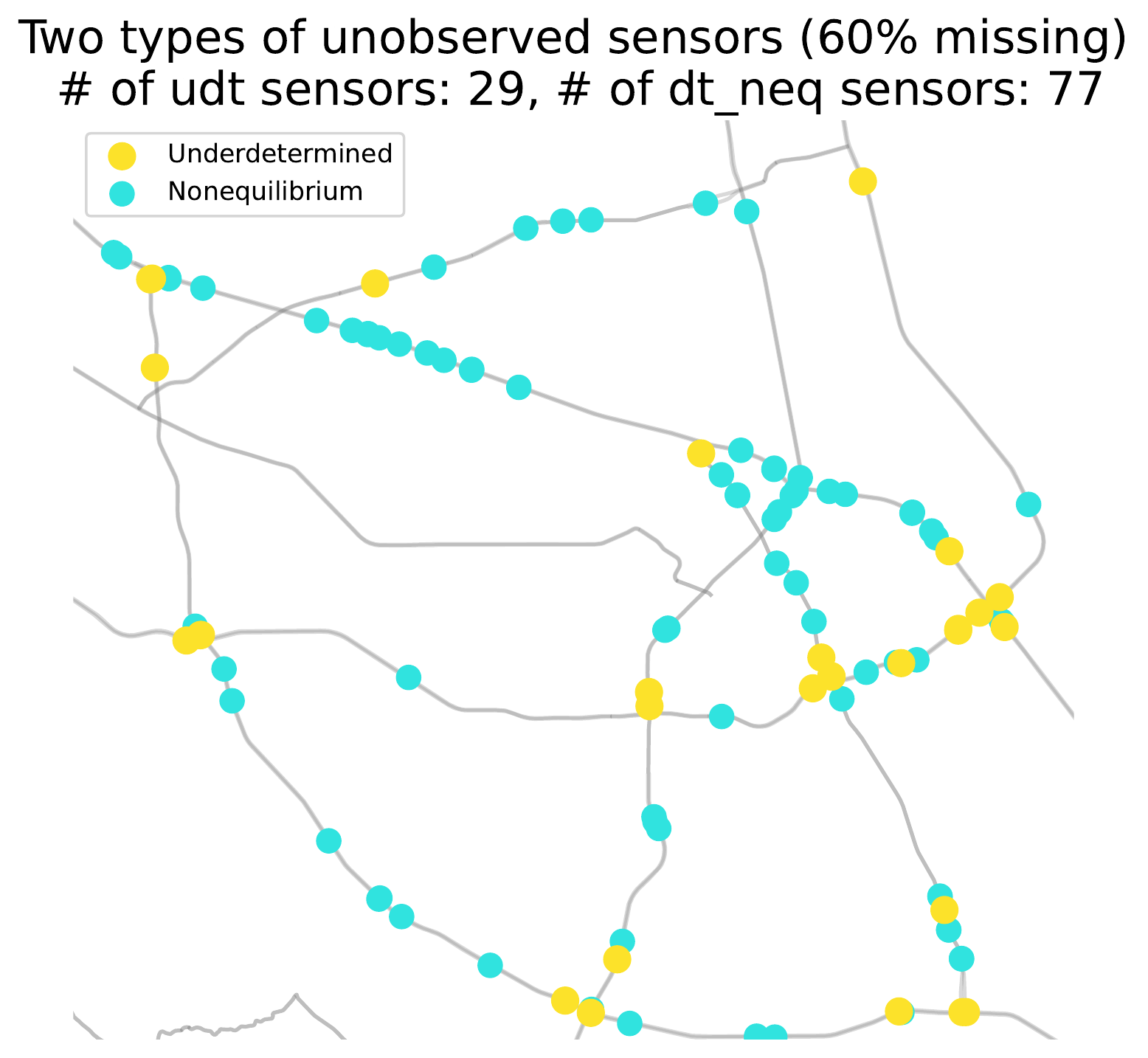}
}
\subfigure[70\% missing rate]{
\centering
\includegraphics[scale=0.33]{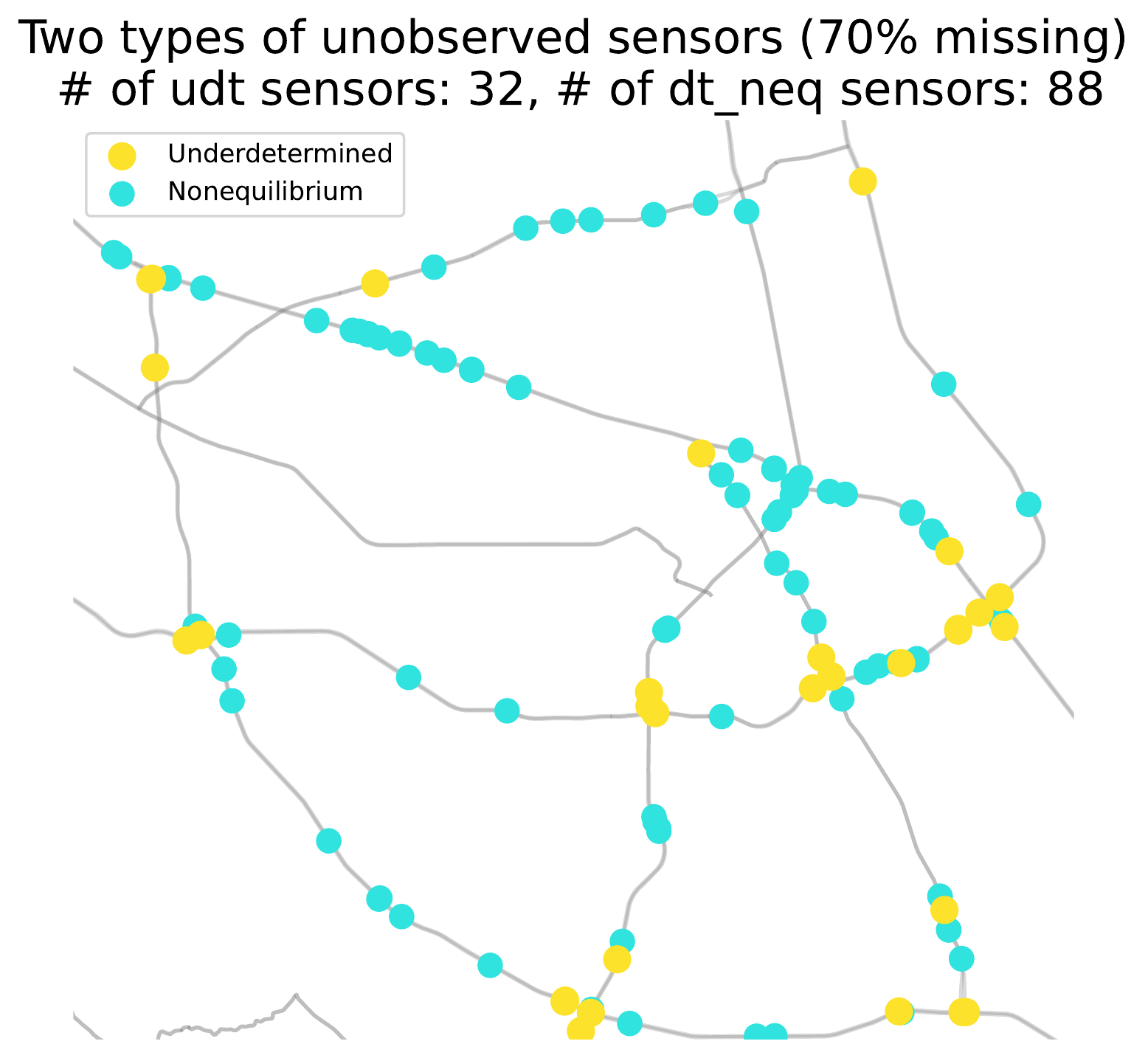}
}
\subfigure[80\% missing rate]{
\centering
\includegraphics[scale=0.33]{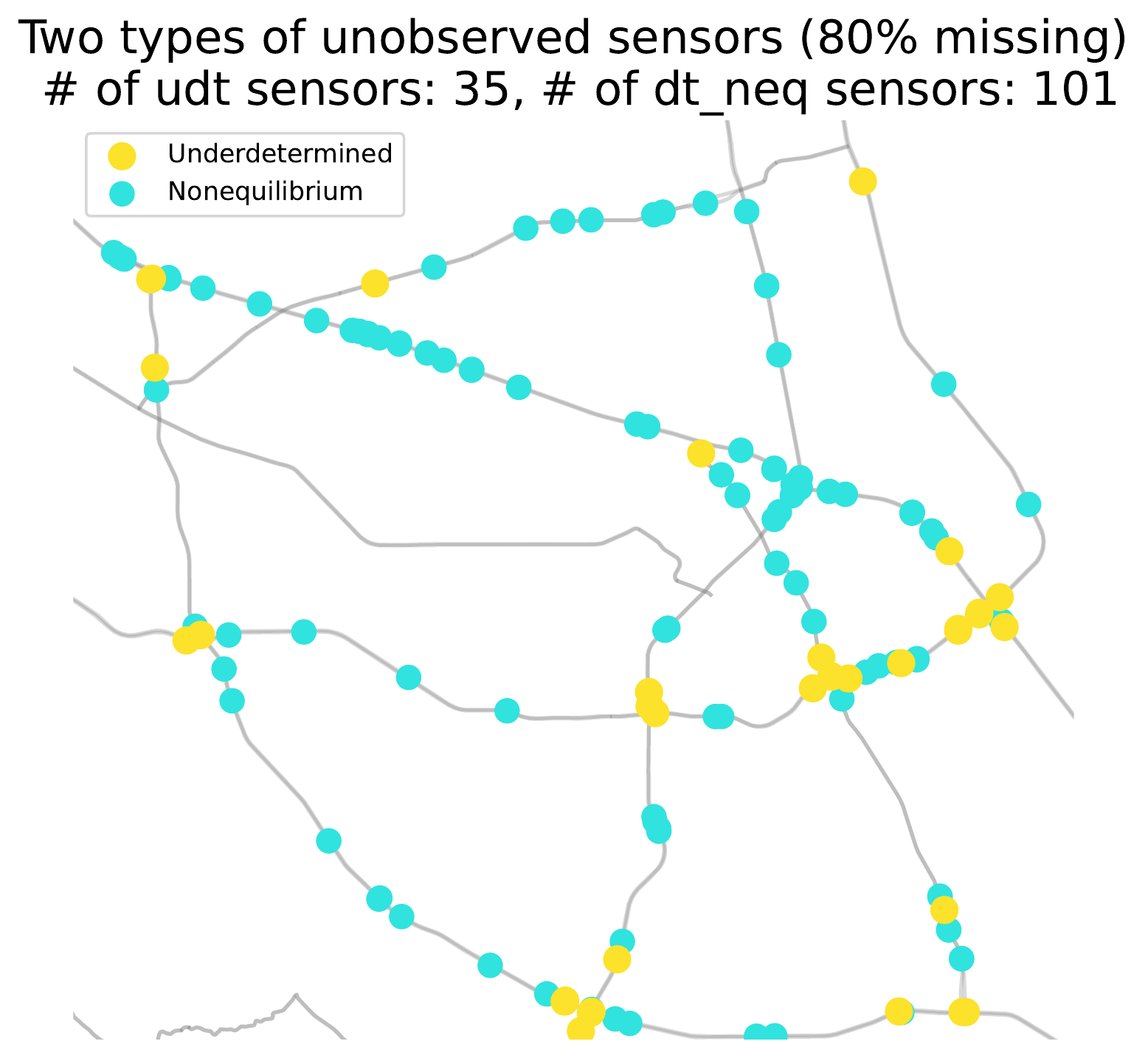}
}
\caption{\minew{Distribution of underdetermined and nonequilibrium flows under different missing rates. The total count for each type is indicated in the title. Note that the ``dt\_eq'' type is omitted here for a clearer visualization.}} 
\label{two_types_distribution_sup}
\end{figure}
\minew{We further divide the unobserved sensors into three categories according to the descriptions in section \ref{results_on_50_missing} and display the locations and numbers of underdetermined and nonequilibrium flows in Fig. \ref{two_types_distribution_sup}.}

\nolinenumbers
\newpage
% \bigskip
\section*{Acknowledgement}
This research was sponsored by the National Natural Science Foundation of China (52125208), the Science and Technology Commission of Shanghai Municipality (No. 22dz1203200), the National Natural Science Foundation of China Youth Science Foundation Program (52302413), and National Key Research and Development Program of China (2022YFB2602100).

% , the Shanghai Municipal Science and Technology Major Project (No. 2021SHZDZX0100)

% \newpage
\footnotesize
\bibliographystyle{elsarticle-harv}
\bibliography{main}

% \begin{thebibliography}{33}
% \expandafter\ifx\csname natexlab\endcsname\relax\def\natexlab#1{#1}\fi
% \expandafter\ifx\csname url\endcsname\relax
%   \def\url#1{\texttt{#1}}\fi
% \expandafter\ifx\csname urlprefix\endcsname\relax\def\urlprefix{URL }\fi

% \bibitem[{Asif et~al.(2016)Asif, Mitrovic, Dauwels, and
%   Jaillet}]{asif2016matrix}
% Asif, M.~T., Mitrovic, N., Dauwels, J., Jaillet, P., 2016. Matrix and tensor
%   based methods for missing data estimation in large traffic networks. IEEE
%   Transactions on intelligent transportation systems 17~(7), 1816--1825.
% \bibitem[{Bolte et~al.(2014)Bolte, Sabach, and Teboulle}]{bolte2014proximal}
% Bolte, J., Sabach, S., Teboulle, M., 2014. Proximal alternating linearized
%   minimization for nonconvex and nonsmooth problems. Mathematical Programming
%   146~(1), 459--494.
% \end{thebibliography}

\end{document}